\documentclass[10pt,twocolumn,letterpaper]{article}

\usepackage[pagenumbers]{cvpr} % To force page numbers, e.g. for an arXiv version

\definecolor{cvprblue}{rgb}{0.21,0.49,0.74}
\usepackage[pagebackref,breaklinks,colorlinks,allcolors=cvprblue]{hyperref}
\usepackage{multirow}
\usepackage{arydshln}
\usepackage[accsupp]{axessibility}  % Improves PDF readability for those with disabilities.

\def\paperID{*****} % *** Enter the Paper ID here
\def\confName{CVPR}
\def\confYear{2025}

\title{DeNVeR: Deformable Neural Vessel Representations for Unsupervised Video Vessel Segmentation}

\author{
Chun-Hung Wu$^1$
\quad
Shih-Hong Chen$^1$
\quad
Chih-Yao Hu$^2$
\quad
Hsin-Yu Wu$^1$
\quad
Kai-Hsin Chen$^1$
\\
Yu-You Chen$^1$
\quad
Chih-Hai Su$^1$
\quad
Chih-Kuo Lee$^2$
\quad
Yu-Lun Liu$^{1}$\vspace{0.75em}
\\
\centerline{$^1$National Yang Ming Chiao Tung University \quad $^2$National Taiwan University}
}

\begin{document}
\maketitle
\begin{abstract}

This paper presents \textbf{De}formable \textbf{N}eural \textbf{Ve}ssel \textbf{R}epresentations (DeNVeR), an unsupervised approach for vessel segmentation in X-ray angiography videos without annotated ground truth. DeNVeR utilizes optical flow and layer separation techniques, enhancing segmentation accuracy and adaptability through test-time training. Key contributions include a novel layer separation bootstrapping technique, a parallel vessel motion loss, and the integration of Eulerian motion fields for modeling complex vessel dynamics. A significant component of this research is the introduction of the XACV dataset, the first X-ray angiography coronary video dataset with high-quality, manually labeled segmentation ground truth. Extensive evaluations on both XACV and CADICA datasets demonstrate that DeNVeR outperforms current state-of-the-art methods in vessel segmentation accuracy and generalization capability while maintaining temporal coherency. 
Please see our project page at \href{https://kirito878.github.io/DeNVeR/}{kirito878.github.io/DeNVeR}.
\end{abstract}    
\section{Introduction}
\label{sec:intro}

Coronary arteries (CAs) are essential for delivering oxygen-rich blood to the heart muscle~\cite{dodge1992lumen}. To assess coronary artery circulation and diagnose disease, cardiologists traditionally use hemodynamic measures like fractional flow reserve (FFR) to determine stenosis severity~\cite{gotberg2017instantaneous}. Since pressure wire-based techniques are invasive and risky~\cite{stables2022routine}, X-ray coronary angiography (XCA) has emerged as a preferred method, using contrast agents to capture vessel structure while minimizing radiation exposure.

Current approaches to vessel segmentation face several key challenges: (1) Annotation costs: Manual labeling of frame-by-frame vessel deformation is prohibitively expensive, particularly in medical settings where expert knowledge is required and hundreds of video frames need precise annotation. (2) Domain gaps: Models trained on one type of X-ray imaging system or protocol often fail to generalize to others, limiting clinical applicability across different institutions. (3) Temporal coherence: Frame-by-frame methods struggle with consistency, leading to flickering and discontinuities in vessel segmentation across video frames. This is especially critical in XCA, where cardiac motion and blood flow create complex temporal dynamics. (4) Physical constraints: Results often violate basic structural properties of blood vessels, such as connectivity and smooth deformation patterns, leading to physiologically implausible segmentations.

\begin{figure}[t]
\centering
\resizebox{1.\columnwidth}{!} 
{
\includegraphics[width=1.0\columnwidth]{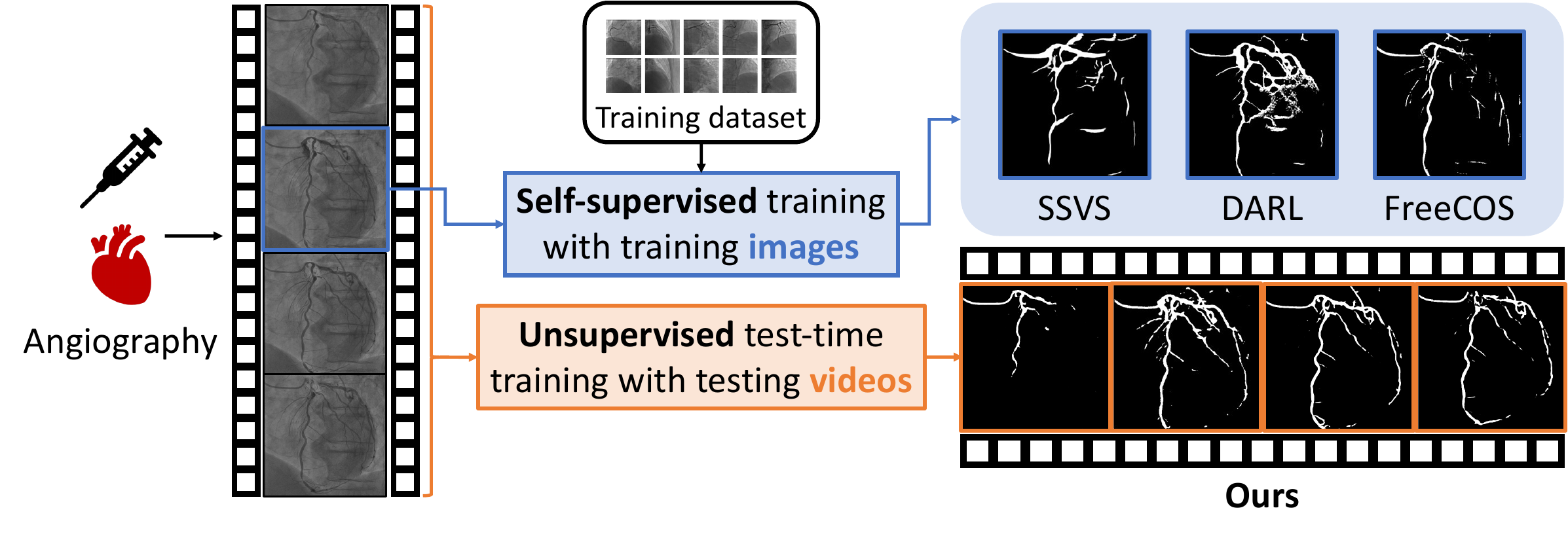}
}
\vspace{-4mm}
\caption{
\textbf{Vessel segmentation method comparison.} Unlike SSVS~\cite{ma2021self}, DARL~\cite{kim2022diffusion}, and FreeCOS~\cite{shi2023freecos}, which require extensive training data, which limits their ability to generalize to new data, our method uses \emph{unsupervised test-time training} on \emph{testing videos}. This approach achieves superior accuracy with finer, more consistent vessel contours, demonstrating robust generalization with minimal data.
}
\label{fig:motivation}
\end{figure}

These challenges are exacerbated by XCA's inherent limitations~\cite{toth2014evolving}: low signal-to-noise ratios, minimal radiation contrast~\cite{felfelian2016vessel}, and interference from surrounding structures~\cite{maglaveras2001artery}. Existing automatic segmentation algorithms require extensive professional supervision and time-consuming annotation~\cite{iyer2023multi}. Single-image approaches discard critical temporal information and show poor adaptability across imaging systems. Additionally, involuntary organ motions and overlapping structures increase ghosting artifacts~\cite{liu2020dynamic, lin2005extraction}, significantly limiting the practical application of current methods (\cref{fig:motivation}).

To address these challenges, we introduce DeNVeR (Deformable Neural Vessel Representations), an unsupervised approach for segmenting cardiac vessels in X-ray videos. Our method combines traditional Hessian-based filters for initial vessel masks with a novel layered separation process that decomposes foreground vessels from the background. We enhance segmentation through test-time optimization, incorporating Eulerian motion field representations~\cite{holynski2021animating} and a parallel vessel motion loss. This approach ensures detailed, temporally consistent segmentation while adapting to cardiac movements and vessel flow patterns. Our experimental results demonstrate significant improvements over state-of-the-art models in vessel region prediction.
The main contributions are:
\begin{itemize}
\item DeNVeR uses unsupervised learning on X-ray video data, leveraging the full temporal information of the videos and eliminating the need for annotated training datasets.
\item Using optical flow and a unique layer separation strategy, DeNVeR enhances segmentation accuracy and adjusts during test time, improving adaptability and ensuring consistent results across cardiac conditions.
\item We collect the first X-ray angiography coronary video dataset (XACV) with high-quality, manually labeled segmentation ground truth, serving as a new standard for training and evaluating video vessel segmentation models, making full use of video temporal information.
\end{itemize}

\section{Related Work}
\label{sec:related}

\noindent {\bf Traditional segmentation methods.}
Traditional object segmentation~\citep{khan2020hybrid,memari2019retinal} requires heuristic human design rules or filters. Several methods are proposed, one of which designs the Hessian-based filter~\citep{frangi1998multiscale} to enhance vessel filtering. Khan \textit{et al.}~\citep{khan2020hybrid} design retinal image denoising and enhancement of B-COSFIRE filters to perform segmentation. Memari \textit{et al.}~\citep{memari2019retinal} use contrast-limited adaptive histogram equalization and designed filters to achieve the task. Another line of work proposes optimally oriented flux (OOF)~\citep{wang2020higher,law2008three}, which performs better in adjacent curvilinear object segmentation. These human strategy design methods do not need any training, which gives them the advantage of fast segmentation of new data. However, these methods are often confined to certain datasets and loss of generalize ability.

\vspace{3pt}  \noindent {\bf Supervised and self-supervised segmentation.}
In the domain of the supervised segmentation method \citep{soomro2019impact}, Esfahani \textit{et al.} \citep{nasr2016vessel} design a Top-Hat transformation and Convolutional Neural Networks (CNNs) for segmentation. Khowaja \textit{et al.} \citep{khowaja2019framework} applied bidirectional histogram equalization. Another work \citep{yang2018automatic,revaud2016deepmatching} utilizes image masking to reduce artifacts, which relies on paired mask datasets. The most popular vessel segmentation backbone recently is U-Net \citep{ronneberger2015unet}. These methods \citep{fan2019accurate,soomro2019impact,yang2019deep} also require extensive and time-consuming human annotation, further limiting the application. Consequently, self-supervised learning methods are designed to elevate performance with large-scale unsupervised data. Some self-supervised learning research focuses on image painting \citep{pathak2016context}, image colorization \citep{larsson2017colorization}, and others \citep{doersch2015unsupervised,noroozi2016unsupervised,ledig2017photo,ren2018cross,misra2016shuffle,xu2019self,benaim2020speednet,doersch2015unsupervised,pathak2016context,gidaris2018unsupervised,misra2020self,ma2021self,xie2021detco,bar2022detreg,park2020contrastive,wu2021contrastive,wang2022contrastmask,alonso2021semi,zhong2021pixel}. Ma \textit{et al.} \citep{ma2021self} and Kim \textit{et al.} \citep{kim2022diffusion} proposed vessel segmentation methods with adversarial learning. Unlike these supervised and self-supervised methods requiring extensive annotations, our DeNVeR uses an unsupervised approach, training directly on test videos. It leverages optical flow and layer separation, enhancing accuracy and adaptability through test-time training. DeNVeR also utilizes temporal information, producing more coherent results than single-frame methods.

\vspace{3pt}  \noindent {\bf Unsupervised segmentation methods.}
Unsupervised segmentation methods fall into two categories: clustering-based and adversarial. Clustering-based approaches~\citep{ji2019invariant,li2021contrastive,do2021clustering} like Invariant Information Clustering (IIC) by Xu et al. \citep{ji2019invariant} inputs into clusters but struggle with curvilinear objects. Adversarial methods~\citep{chen2019unsupervised,abdal2021labels4free}, exemplified by Redo~\citep{abdal2021labels4free}, generate object masks by guiding generators with inputs to redraw objects in new colors.

\vspace{3pt}  \noindent {\bf Video segmentation methods.}
Coronary artery segmentation based on sequential images such as SVS-Net~\citep{hao2020sequential} uses an encoder-decoder deep network architecture that utilizes multiple contextual frames of 2D and sequential images to segment 2D vessel masks. However, these supervised methods often suffer from domain gaps between datasets and cannot generalize well.
Our work advances video decomposition into layers, originally proposed by Wang \& Adelson~\citep{wang1994representing} in the 1990s, by incorporating neural network techniques~\citep{liu2020learning,liu2021learning}. Unlike traditional methods~\citep{black1991robust,jojic2001learning,ost2021neural,shi1998motion,brox2010object}, we operate unsupervised, learning deformable canonical layers to model vessel motion more effectively. 
% Additionally, unlike unsupervised video segmentation methods~\citep{yang2019unsupervised,ye2022deformable}, we address motion segmentation by associating pixels with Eulerian motion~\citep{holynski2021animating} clusters and extend our approach to unsupervised video vessel segmentation. 
Additionally, while previous research, such as ~\cite{yang2019unsupervised} and~\cite{ye2022deformable}, focused on general unsupervised video segmentation, we extend these concepts to the specific domain of vessel segmentation. We address motion segmentation by associating pixels with Eulerian motion~\citep{holynski2021animating} clusters, adapting and extending this approach to unsupervised video vessel segmentation.
Our method, DeNVeR, separates X-ray video into canonical foreground and background, per-frame masks, and dynamic transformations for realistic vessel motion representation optimized through specific loss functions.
\section{Method}
\label{sec:method}

\begin{figure*}[t]
\centering
% \small
% \setlength{\tabcolsep}{1pt}
% \renewcommand{\arraystretch}{1}
\resizebox{1.0\textwidth}{!} 
{
\includegraphics[width=1.0\textwidth]{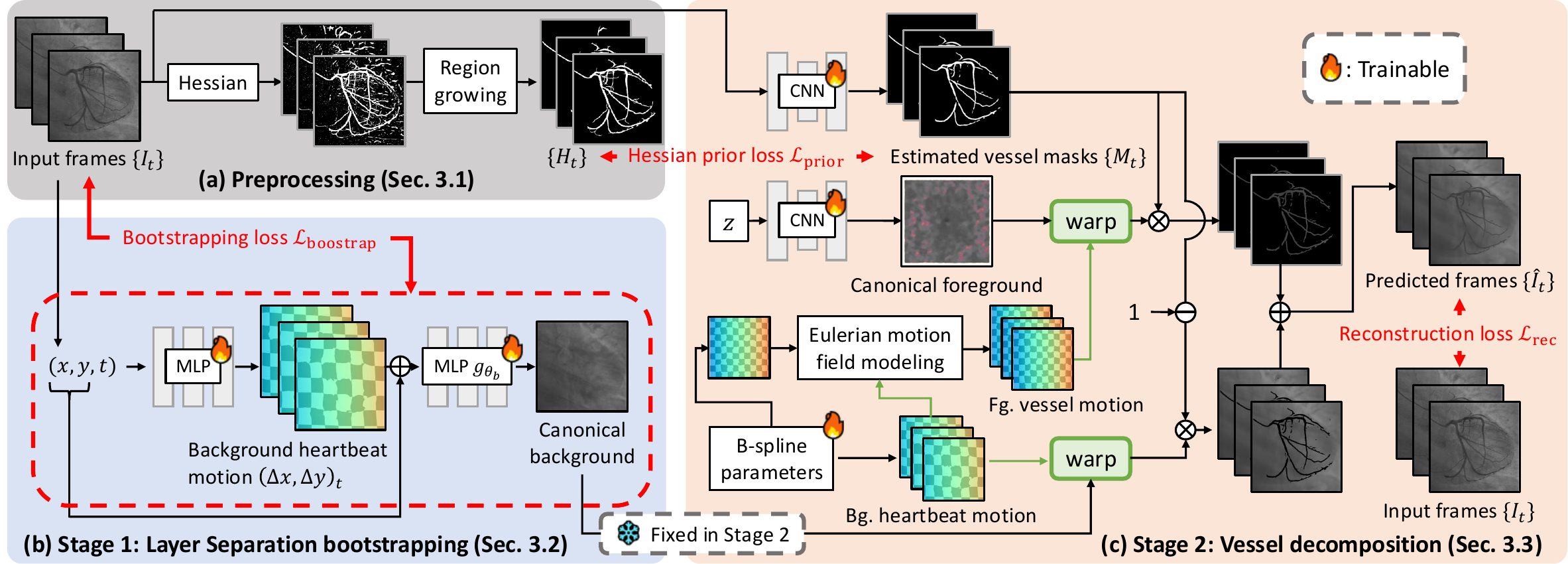}
}
% \caption{\textbf{The DeNVeR pipeline operates unsupervised during test time for vessel extraction from X-ray videos.} In the preprocessing phase (a), we apply a Hessian-based technique complemented by region growing for frame-specific vessel segmentation. Subsequently, in (b) Stage 1, Multi-Layer Perceptrons (MLPs) are employed to model both the background deformation fields and a canonical background image, establishing a baseline devoid of vessel structures via a bootstrapping loss criterion. Finally, in (c) Stage 2, the canonical background is held constant while we refine the canonical foreground vessel image, per-frame vessel masks, and their respective motions. This involves the utilization of B-spline parameters to capture vessel and background movement, followed by a warping process that merges the foreground and background layers to reconstruct the frames. The reconstruction loss is minimized to ensure fidelity to the original input frames. This entire pipeline is trained directly on test videos without the need for ground truth segmentation masks.}
\vspace{-4mm}
\caption{\textbf{Pipeline for unsupervised vessel segmentation from X-ray videos.}
(a) Preprocessing: Hessian-based technique with region growing for initial segmentation.
(b) Stage 1: MLPs model background deformation and canonical image using bootstrapping loss.
(c) Stage 2: Refine foreground vessel image, masks, and motions using B-spline parameters and warping.
Reconstruction loss ensures fidelity to input frames. The pipeline trains directly on test videos without ground truth masks.
}
% \vspace{-2mm}
\label{fig:pipeline}
\end{figure*}

% % Inspired by Deformable Sprites \citep{ye2022deformable}, 
% We develop a novel unsupervised algorithm for Cardiac Vessel Segmentation in X-ray video based on optical flow. Given a cardiac X-ray video, it can perform test-time optimization by computing reconstruction without any training data. Initially, we employ traditional blood vessel segmentation methods to extract vascular regions in a very coarse manner (Section~\ref{sec:preprocess}). 
% Since each input frame is processed independently, the temporal consistency of the output video is severely compromised, leading to fragmentation.
% To address those issues, we employ implicit neural representation to learn layer separation for segmenting blood vessels (foreground) from X-ray heart images (background) and estimate the corresponding background flow with lower degrees of freedom (Section~\ref{sec:Background bootstrapping}). 
% Subsequently, we fix the background transformation to optimize the contrast agent flow in the foreground. Here, we introduce our test-time training workflow (Section~\ref{sec:TTT}). Finally, we elaborate on the loss and regularization terms we use (Section~\ref{sec:loss and Regularization}).
We propose an unsupervised algorithm for Cardiac Vessel Segmentation in X-ray videos using optical flow and test-time optimization. Our approach involves coarse vascular region extraction (\cref{sec:preprocess}), layer separation using implicit neural representation (\cref{sec:Background bootstrapping}), background flow estimation and foreground optimization (\cref{sec:Background bootstrapping}), and application of specific loss and regularization terms (\cref{sec:loss and Regularization}). This method addresses temporal consistency issues and enables segmentation without training data.

\subsection{Preprocessing}  \label{sec:preprocess}
Segmenting vascular regions from X-ray images through unsupervised methods is a highly challenging task. To facilitate our subsequent work, we employ a Hessian-based filter~\cite{frangi1998multiscale} to generate a set of binary masks that crudely represent blood vessel regions. Specifically, our approach comprises the following two steps:\\
(1) Apply a Hessian-based filter to the entire sequence. The output pixel intensities range from 0 to 255, with higher numerical values indicating a stronger presence of tubular structural features.\\
(2) Based on the outputs from step (1), calculate the overall intensities of the entire image and set an appropriate threshold to convert them into binary images. We use Otsu's method~\cite{otsu1975threshold} to automatically determine the threshold. The purpose of selecting the threshold is to ensure that images with higher intensities correspond to larger vascular areas. Finally, we employ a region-growing post-processing technique to eliminate noise.\\
% \yulunliu{ preprocessing : Raft }
Additionally, we use a pre-trained optical flow model, RAFT~\cite{teed2020raft}, to generate the initial optical flow between consecutive frames. We illustrate the preprocessing steps in \cref{fig:pipeline} (a).

\subsection{Layer Separation Bootstrapping}  
\label{sec:Background bootstrapping}
While utilizing the Hessian-based filter allows us to quickly acquire a set of rough masks, the periodic heartbeat introduces temporal inconsistency, resulting in variations in the position of vascular regions over time. In addressing this challenge, we implement a solution by separating the input frames into foreground and background. Inspired by NIR~\cite{nam2022neural}, we embrace the approach of employing MLPs to learn implicit neural representations of images (\cref{fig:pipeline} (b)). Our canonical foreground and background images leverage neural representations analogous to canonical radiance field approaches recently used in 3D scene reconstruction~\cite{liu2023robust,su2024boostmvsnerfs} and video editing~\cite{chen2024narcan}. The primary objective of each MLP is to minimize the following losses:
\begin{equation}
\begin{gathered}
\mathcal{L}_{\text{recons}} = \sum_{x, y, t}\|\hat{{I}}(x,y,t) - {I}(x,y,t)\|_2^2, \\
\mathcal{L}_{\text{smooth}} = \sum_{x, y, t}\|{J}_{g_{\theta_{\scriptscriptstyle{b}}}(x,y,t)}\|_1, \mathcal{L}_{\text{limit}} = \sum_{x, y, t}\|{g_{\theta_{\scriptscriptstyle{f}}}(x,y,t)}\|_1, \\
\mathcal{L}_{\text{boostrap}} = \mathcal{L}_{\text{recons}} + \lambda_{\text{smooth}}\mathcal{L}_{\text{smooth}} + \lambda_{\text{limit}}\mathcal{L}_{\text{limit}},
\end{gathered}
\end{equation}
% \begin{equation}
% \mathcal{L}_{\text{smooth}} = \sum_{x, y, t}\|{J}_{g_{\theta_{\scriptscriptstyle{b}}}(x,y,t)}\|_1,
% \end{equation}
% \begin{equation}
% \mathcal{L}_{\text{limit}} = \sum_{x, y, t}\|{g_{\theta_{\scriptscriptstyle{f}}}(x,y,t)}\|_1,
% \end{equation}
% \begin{equation}
% \mathcal{L}_{\text{boostrap}} = \mathcal{L}_{\text{recons}} + \lambda_{\text{smooth}}\mathcal{L}_{\text{smooth}} + \lambda_{\text{limit}}\mathcal{L}_{\text{limit}},
% \end{equation}
where ${I}$ and $\hat{{I}}$ denote the original ground truth (i.e., input RGB frames) and the output of the first MLP, respectively. To ensure flow smoothness, we introduce a penalty term for the MLP computing the background, denoted as \(g_{\theta_{\scriptscriptstyle{b}}}\). Here, \({J}_{g_{\theta_{\scriptscriptstyle{b}}}}(x,y,t)\) represents a Jacobian matrix comprising gradients of \(g_{\theta_{\scriptscriptstyle{b}}}\).
Finally, since the stationary background should occupy the vast majority of the scene, we introduce an additional penalty term for $g_{\theta_{\scriptscriptstyle{f}}}$ which learns to represent the scene beyond the background. $\lambda_{\text{smooth}}$ and $\lambda_{\text{limit}}$ are weight hyperparameters.
While \(g_{\theta_{\scriptscriptstyle{f}}}\) and \(g_{\theta_{\scriptscriptstyle{b}}}\) estimate foreground and background, they lack the spatial resolution needed for precise segmentation. Our full pipeline below refines these estimates.

\subsection{Test-Time Training for Vessel Decomposition}   \label{sec:TTT}
Our approach introduces \emph{test-time training} as a key feature of our unsupervised segmentation method, DeNVeR. To ensure temporal coherence in the predicted segmentation masks, we adopt dense deformation fields, similar to approaches that stabilize video frames using dense warp estimation~\cite{liu2021hybrid}. Using the test video's inherent structure and patterns, the model refines its parameters, allowing it to tailor its learning to each video's unique characteristics.

After obtaining the Hessian-based approach as prior and bootstrapped static background, we focus on utilizing a pre-trained optical flow estimator, RAFT~\cite{teed2020raft}, to further separate the vessel and background layer and obtain vessel segmentation masks.
As shown in \cref{fig:pipeline} (c), for each image, we use a CNN model to predict masks for the foreground and background. Both foreground and background have their canonical images. It's important to note that these images do not correspond to a specific cardiac phase. Instead, it is learned to represent the overall vessel structure across the cardiac cycle. To simplify the problem, we first compute the canonical image for the background in Stage 1 (\cref{sec:Background bootstrapping}) and keep it fixed. Then, in Stage 2, we optimize the canonical foreground using a CNN model with a fixed latent code $z$ \cite{ulyanov2018deep}. The latent code $z$ is randomly initialized and fixed during optimization. Its purpose is to provide a consistent input for generating the canonical foreground across different optimizations. The CNN is trained during the test-time optimization process for each video. Next, we use motion flow to reconstruct respective images from the canonical images.
To enhance the coherence of each frame's mask, we calculate the flow warp loss, requiring both foreground and background motion fields. Thus, we utilize the spatial and temporal B-spline to model the entire motion trajectory.

\begin{figure}[t]
\centering
\resizebox{1.\columnwidth}{!} 
{
\includegraphics[width=1.0\columnwidth]{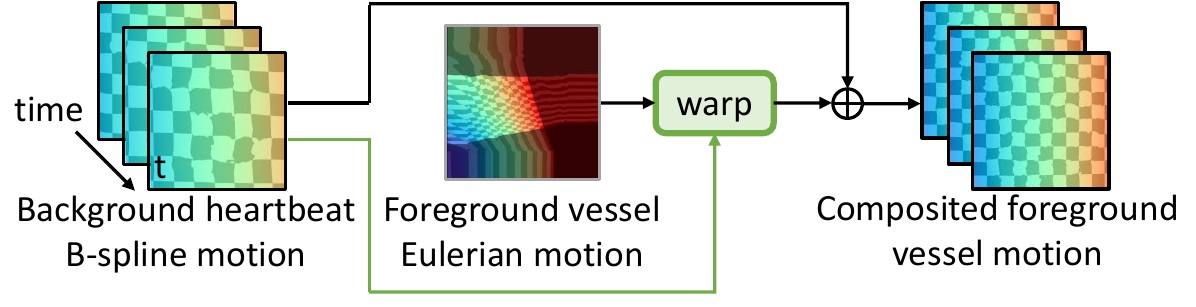}
}
\vspace{-4mm}
\caption{
\textbf{Eulerian motion field modeling.} Background heartbeat uses a low-degree B-spline; foreground vessel flow uses a stationary Eulerian field. Final vessel flow combines warped Eulerian motion with background flow, capturing both factors observed in X-ray videos.
}
\label{fig:eulerian}
\end{figure}

\vspace{3pt}  \noindent {\bf Background motion fields.}
In the case of cardiac X-ray imaging, the background usually includes the heart and ribs, which don't experience significant displacement. Therefore, we use a B-spline with lower degrees of freedom to estimate the motion flow.
% \end{itemize}

\vspace{3pt}  \noindent {\bf Foreground motion fields.}
% \begin{itemize}
    % \item \textbf{Foreground motion fields } 
As for the foreground, we observe that the contrast agent flows out from the catheter. Therefore, we consider the Eulerian motion field, as shown in \cref{fig:eulerian}, to be a more reasonable specification of blood flow behavior compared to the traditional motion field.
% \end{itemize}

\subsection{Losses and Regularizations} \label{sec:loss and Regularization}
\noindent {\bf Hessian prior loss.}
After obtaining the initial mask from preprocess (\cref{sec:preprocess}), we utilize a CNN model for the initial segmentation task. In this step, we aim for the model's predicted mask to closely resemble the mask generated by traditional algorithms. To achieve this, we employ a Hessian prior loss:
\begin{equation}
    \mathcal{L}_\text{prior} = \sum_{x} H_t(x) \cdot M_t(x) + \alpha \cdot (1-H_t(x)) \cdot (1 - M_t(x)),
\end{equation}
where $H_t$ represents the mask of frame t generated by the preprocessing part, $M_t$ denotes the background mask of frame t predicted by the mask model, and $\alpha$ represents the foreground weight. Note that masks generated by this per-frame operation are not temporally continuous, which means that there can be sudden changes in mask predictions between adjacent frames, and our method aims to ensure smoother transitions and consistency in vessel structure across consecutive frames. Therefore, we will optimize continuity through subsequent methods.

\begin{figure}[t]
\centering
\small
\setlength{\tabcolsep}{1pt}
\renewcommand{\arraystretch}{1}
\resizebox{1.\columnwidth}{!} 
{
\begin{tabular}{ccc}
\includegraphics[width=0.33\columnwidth]{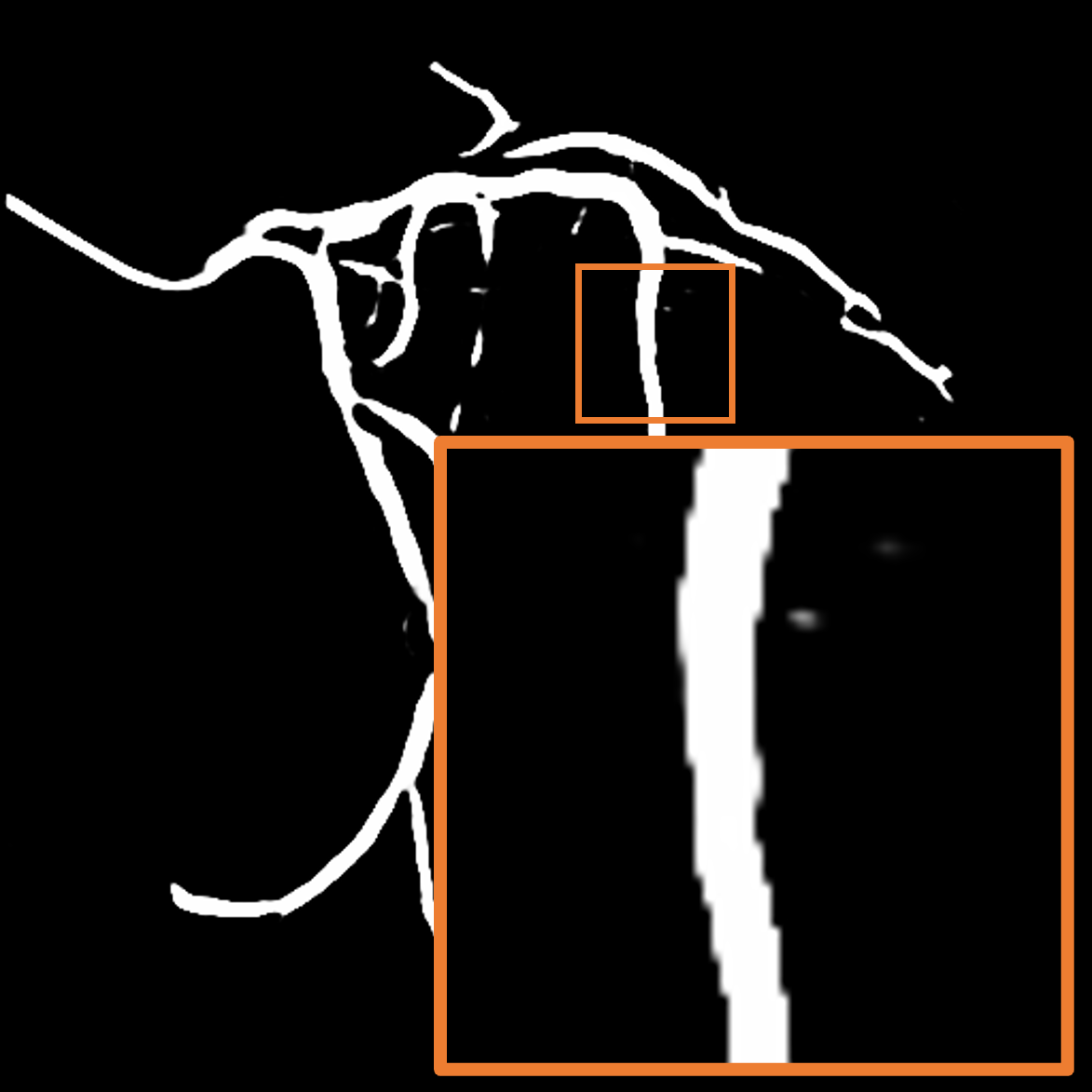} & 
\includegraphics[width=0.33\columnwidth]{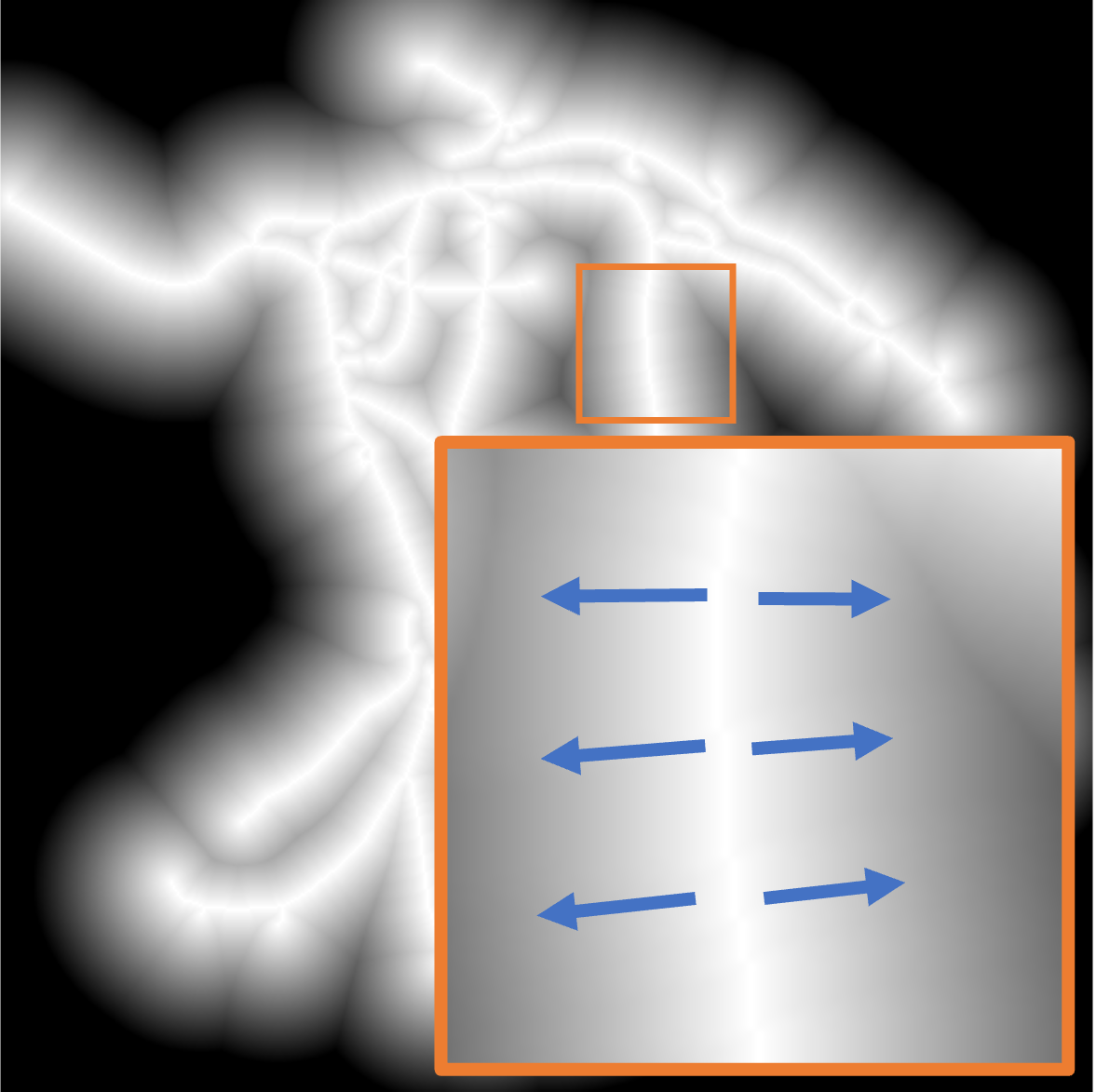} & 
\includegraphics[width=0.33\columnwidth]{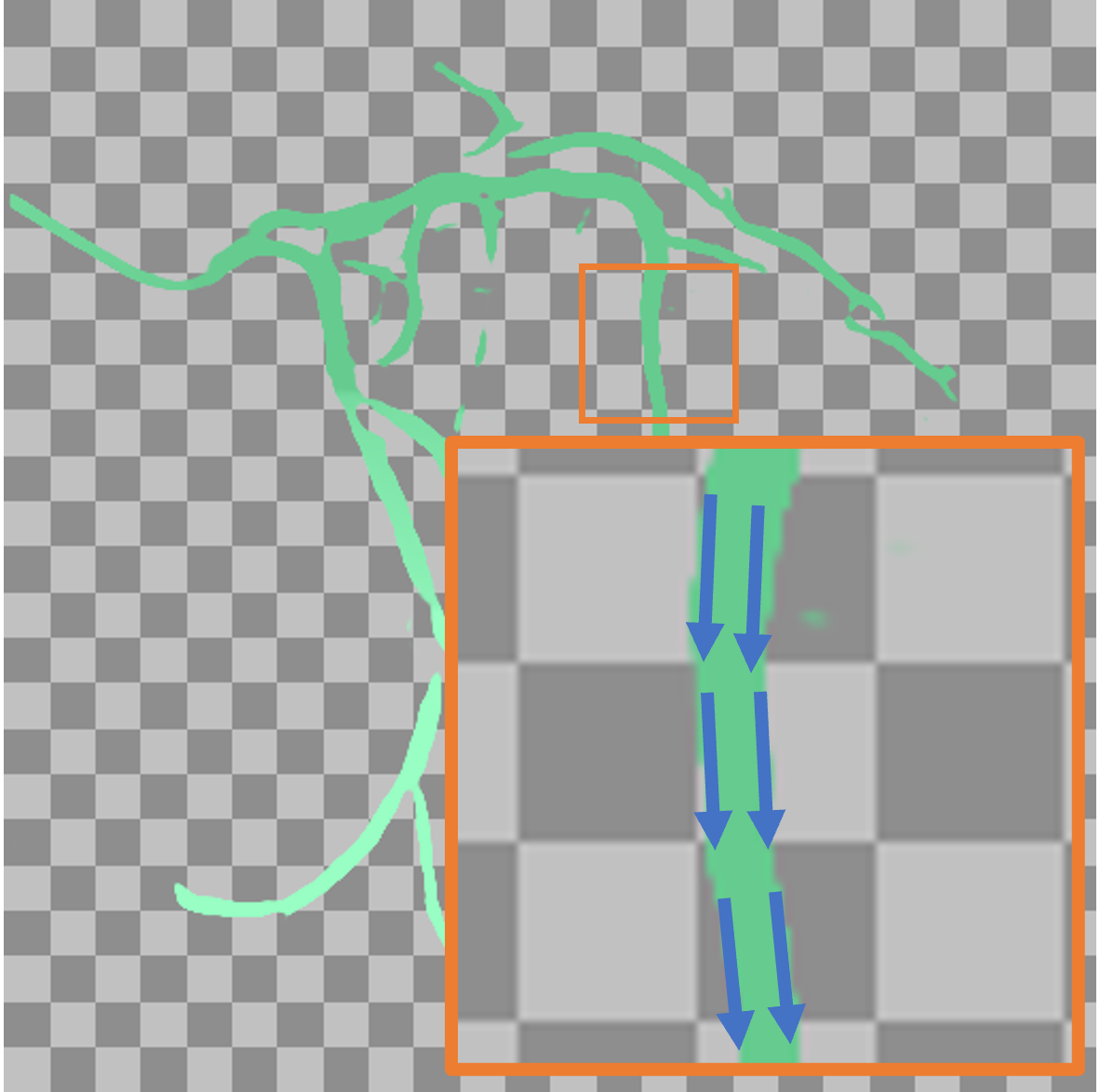} \\
(a) Vessel mask & (b) Distance transform & (c) Vessel motion \\
\end{tabular}%
}
\vspace{-2mm}
\caption{
\textbf{Parallel vessel motion loss.} Aligns flow direction with vessel mask direction. Uses skeletonization and distance transform to determine gradient directions. The predicted vessel motion should be perpendicular to these gradients (blue arrows).
}
\label{fig:parallel}
\end{figure}

\vspace{3pt}  \noindent {\bf Parallel loss.}
Clearly, the direction of blood flow should align with the course of blood vessels (\cref{fig:parallel}). Hence, we design the parallel loss to achieve a parallel alignment between them. Initially, we conduct skeletonization and distance transform on the masks obtained from \cref{sec:preprocess}, and calculate pixel-wise cosine similarity between these transformed masks and the predicted flow:
% \begin{equation}
%  \mathcal{G}_\text{x}(x,y) = \nabla_x D(x,y)
%  , \mathcal{G}_\text{y}(x,y) = \nabla_y D(x,y),
% \end{equation}
% \begin{equation}
%  \mathcal{V}(x,y) =  (\mathcal{G}_\text{x}(x,y) ,  \mathcal{G}_\text{y}(x,y)),
% \end{equation}
% \begin{equation}
%  \mathcal{V}(x) =  (\nabla_u D(x),  \nabla_v D(x)),
% \end{equation}
\begin{equation}
\begin{gathered}
    \mathcal{L}_\text{parallel} = \sum_{x} \frac{\left| \mathcal{V}(x) \cdot \hat{F}(x) \right|}{\|\mathcal{V}(x)\| \cdot \|\hat{F}(x)\|}, \\
    \mathcal{V}(x) =  (\nabla_u D(x),  \nabla_v D(x)),
\end{gathered}
\end{equation}
where $D(x)$ represents the value obtained from the distance transform at pixel coordinate $x$, $\nabla_u$ and $\nabla_v$ are the image gradients from the two spatial directions, and $\hat{F}(x)$ denotes the predicted flow value at position $x$.

\vspace{3pt}  \noindent {\bf Flow warp loss.}
To maintain consistency in the predicted flow for both the foreground vessel and background, we introduce the flow warp loss:
\begin{equation}
    \mathcal{L}_\text{warp} = \sum_{\ell \in [f, b],x} M_t^\ell(x)\cdot \frac{\| \hat{F}_t^\ell(x) - \hat{F}_{t+1}^\ell(F_{t\rightarrow t+1}(x)) \|}{s_t^\ell + s_{t+1}^\ell},
\end{equation}
where $\hat{F}_t^\ell$ is the predicted flow at time $t$, $M_t$ represents the mask for frame t, $F_{t\rightarrow t+1}$ denotes RAFT optical flow computed from frame at time $t$ to $t+1$, $\ell$ denotes the background layer or foreground layer, and $s_t^\ell$ is the scale of $\hat{F}_t^\ell$.
The flow warp loss encourages the flow between nearby frames of both foreground vessel and background layers to follow the guidance from flow predicted by RAFT.

\vspace{3pt}  \noindent {\bf Mask consistency loss.}
Our current method processes a short video clip that lasts only about three seconds. Thus, we suppose the topology of the vessels remains the same during this short time period. For the predicted masks, we compare the mask at time $t$ with the deformed mask at time $t+1$ to ensure consistency across frames. We introduce the mask consistency loss $\mathcal{L}_\text{mask}$:
\begin{multline}
    \mathcal{L}_\text{mask} = \sum_{x} \big| M_t^f(x) - M_{t+1}^f(F_{t\rightarrow t+1}(x)) \big| \\
    + \big| M_t^b(x) - M_{t+1}^b(F_{t\rightarrow t+1}(x)) \big|.
\end{multline}

\vspace{3pt}  \noindent {\bf Reconstruction loss.}
We use the L1 distance between the predicted image and the original image for Reconstruction loss calculation:
\begin{equation}
    \mathcal{L}_\text{rec} = \left\lVert \hat{I}_t - I_t \right\rVert_1.
\end{equation}

Our final loss function is applied to train all components shown as trainable in \cref{fig:pipeline} (c):
\begin{equation}
\begin{split}
        \mathcal{L}_\text{final} = \lambda_\text{prior}\mathcal{L}_\text{prior} + \lambda_\text{parallel}\mathcal{L}_\text{parallel} + \lambda_\text{warp}\mathcal{L}_\text{warp} \\
        + \lambda_\text{mask}\mathcal{L}_\text{mask}
    + \lambda_\text{rec}\mathcal{L}_\text{rec}.
\end{split}
\end{equation}
\section{Experiments}

\begin{figure}[t]
\centering
\small
\setlength{\tabcolsep}{1pt}
\renewcommand{\arraystretch}{1}
\resizebox{1.0\columnwidth}{!}
{
\begin{tabular}{cc:cc:cc}
\includegraphics[width=0.2\columnwidth]{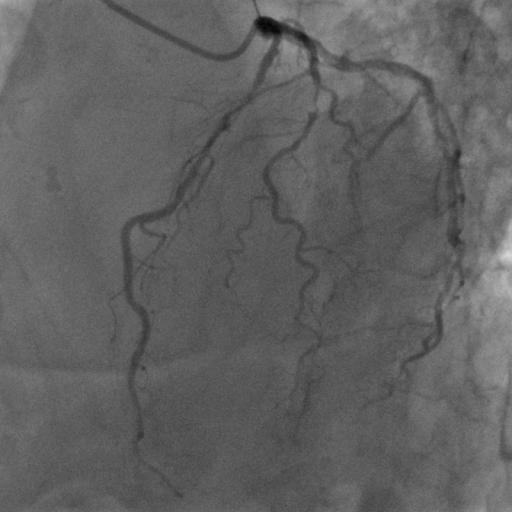} & 
\includegraphics[width=0.2\columnwidth]{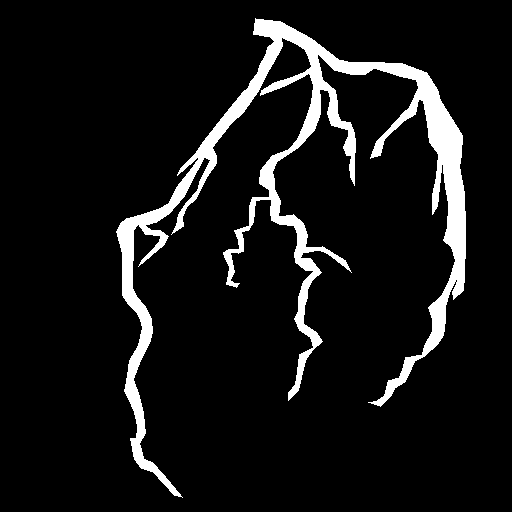} &
\includegraphics[width=0.2\columnwidth]{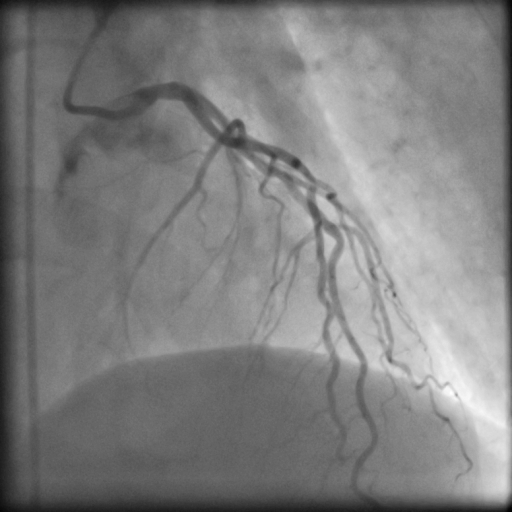} & \includegraphics[width=0.2\columnwidth]{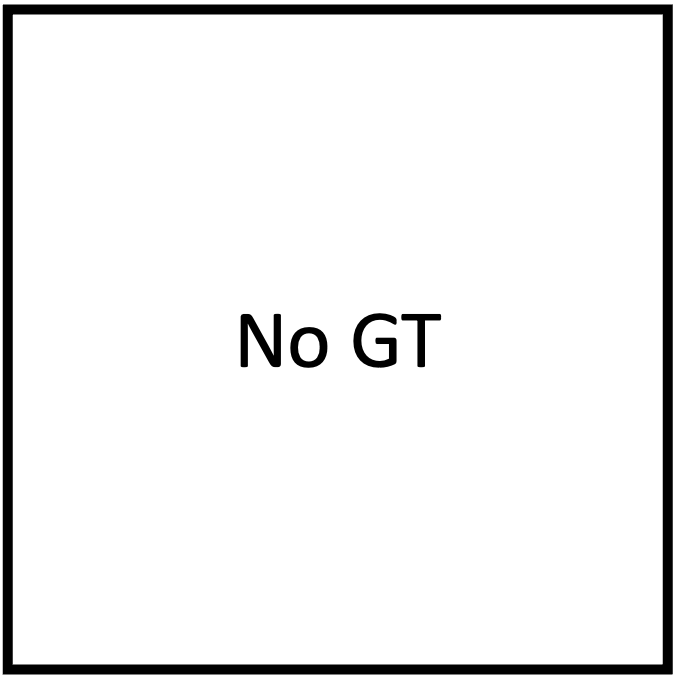} &
\includegraphics[width=0.2\columnwidth]{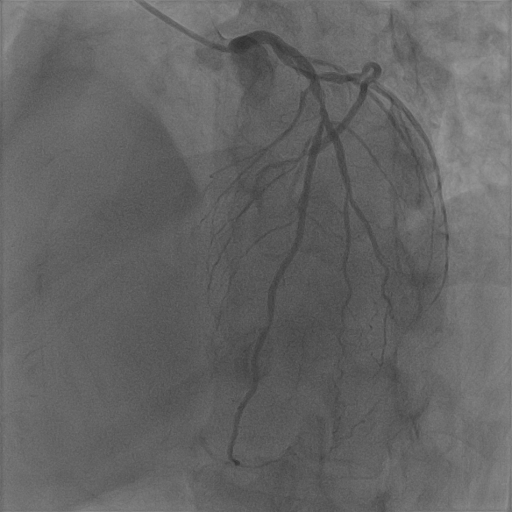} & 
\includegraphics[width=0.2\columnwidth]{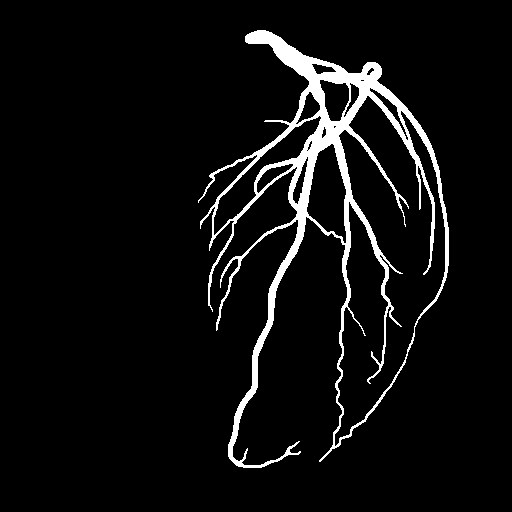} \\
\includegraphics[width=0.2\columnwidth]{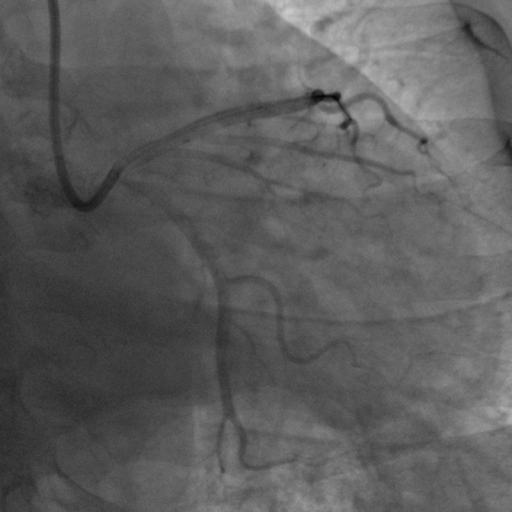} & 
\includegraphics[width=0.2\columnwidth]{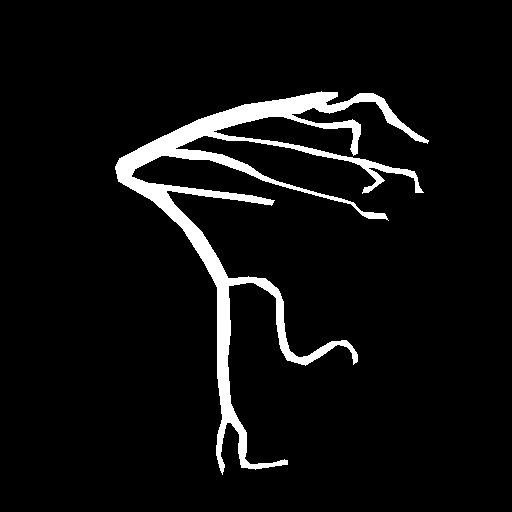} &
\includegraphics[width=0.2\columnwidth]{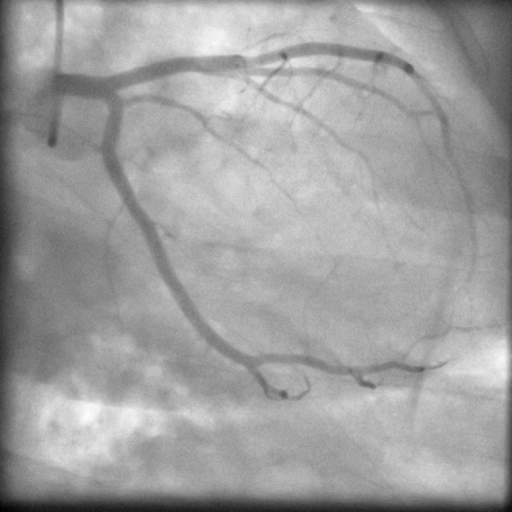} &
\includegraphics[width=0.2\columnwidth]{figures/noGT.png} &
\includegraphics[width=0.2\columnwidth]{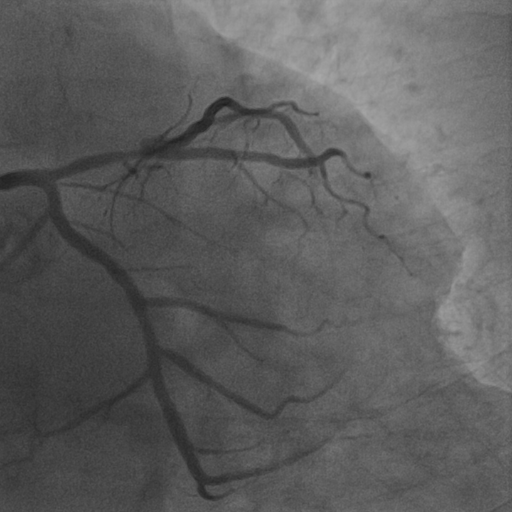} & 
\includegraphics[width=0.2\columnwidth]{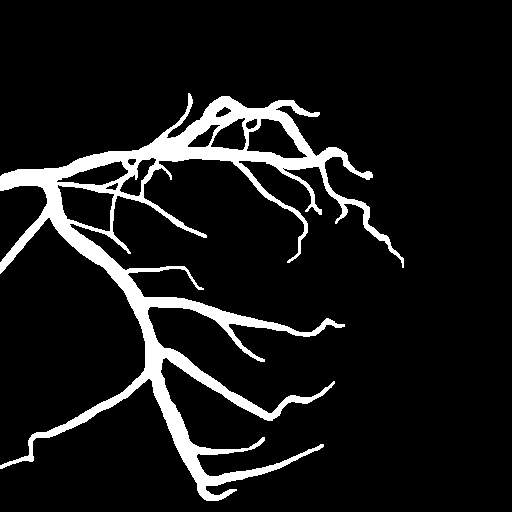}\\

% Image & Ground truth & Video frame & Ground truth & Video frame & Ground truth \\
Image & Ground truth & Video frame & Ground truth & Video frame & Ground truth \\
\multicolumn{2}{c:}{(a) XCAD} & \multicolumn{2}{c:}{(b) CADICA} & \multicolumn{2}{c}{(c) \textbf{Our XACV dataset}}\\
\end{tabular}%
}
\vspace{-2mm}
\caption{
% \textbf{Our dataset provides high-quality manually-labeled vessel segmentation ground truth.} \yulunliu{Show one or two examples of similar X-ray images compared to the XCAD dataset or other datasets. But our GT is much more accurate and detailed.}
\textbf{Comparisons between XCAD~\citep{ma2021self}, CADICA~\citep{jimenez2024cadica}, and our XACV dataset.}
(a) The images from XCAD with their corresponding GTs. (b) The CADICA dataset provides video frames but without corresponding ground truth. (c) Our XACV dataset with GTs was meticulously labeled by experienced radiologists. Our dataset not only provides GTs with greater accuracy and detail, which is evident in the more nuanced vessel delineations, but also features frames of superior quality, facilitating finer and more precise segmentation results.
}
\label{fig:dataset_compare}
\end{figure}

\subsection{XACV Dataset}
We collect 111 complete records of coronary artery X-ray videos from 59 patients, encompassing the injection, flow through the blood vessels around the heart, and dissipation of the contrast agent. Subsequently, we establish the XACV (X-ray Angiography Coronary Video) dataset. Each video consists of an average of 86 frames of high-resolution $512\times512$ coronary artery X-ray images, with an equal distribution of left and right coronary arteries. 
% We invite experienced radiologists to annotate the vascular regions, focusing on one or two frames where the contrast agent is most prominent in each video.
We invite experienced radiologists to annotate the vascular regions, focusing on one or two frames where the contrast agent is most prominent in each video. These annotations are used only for evaluation in our method, not for training, maintaining the unsupervised nature of our approach.
The data collection protocol involves several key steps, including patient preparation with informed consent and metal object removal, image capture using a Philips Allura Xper FD20 machine for standardized frontal (PA) and lateral views, DICOM file storage, and de-identification for patient privacy. Experienced radiologists perform diagnostic annotations using standardized tools and methods, with multiple annotations to enhance accuracy. Quality control measures, secure data management, and strict adherence to ethical guidelines and privacy regulations are implemented throughout the process.
The XCAD dataset contains only a single image, and the CADICA video dataset does not provide corresponding ground truth. Therefore, in the following experiments, we conduct all the analyses on our collected XACV dataset and the corresponding GT for each sequence.
In \cref{fig:dataset_compare}, we show that compared to other publicly available datasets,  XCAD~\citep{ma2021self} and CADICA~\citep{jimenez2024cadica}, our dataset exhibits finer annotations in the vascular regions, providing an advantage for future related tasks. \emph{The development and use of our dataset have been approved by our institution's IRB.}
We include 5 out of 111 sequences of our XACV dataset in the supplementary material to showcase the high quality of both our X-ray frames and manual ground truth annotations.
We will make the XACV dataset publicly available.
% The XCAD dataset contains only a single image, and the CADICA video dataset does not provide corresponding ground truth. Therefore, in the following experiments, we conduct all the analyses on our collected XACV dataset and the corresponding GT for each sequence.
% In Figure~\ref{fig:dataset_compare}, we show that compared to other publicly available datasets,  XCAD~\citep{ma2021self} and CADICA~\citep{jimenez2024cadica}, our dataset exhibits finer annotations in the vascular regions, providing an advantage for future related tasks. \emph{The development and use of our dataset have been approved by our institution's IRB.}

\begin{table*}[t]
\centering
% \small
% \setlength{\tabcolsep}{2pt}
% \renewcommand{\arraystretch}{0.4}
\small
\caption{\textbf{Quantitative evaluation with different methods on the XACV dataset.} Method categories: S: Supervised, T: traditional, SS: Self-supervised, U: unsupervised. Bold indicates the best performance among traditional, self-supervised, and unsupervised methods. Our unsupervised method (DeNVeR) aims to outperform existing non-supervised approaches.}
\vspace{-2mm}
\label{tab:quantitative}
\resizebox{\textwidth}{!}{%
\begin{tabular}{lllccccccc}
\toprule
 & Input & Method & clDice & NSD & Jaccard & Dice & Acc. & Sn. & Sp.\\
\midrule
T & Image & Hessian~\citep{frangi1998multiscale} & $0.577_{\pm 0.062}$ & $0.321_{\pm 0.066}$ & $0.415_{\pm 0.055}$ & $0.584_{\pm 0.055}$ & $0.929_{\pm 0.015}$ & $0.451_{\pm 0.062}$ & $\bf{0.990}_{\pm 0.008}$ \\
\midrule
S & Image & U-Net~\citep{ronneberger2015unet} & $ 0.757_{\pm 0.114}$ & $0.603_{\pm 0.126}$ & $0.638_{\pm 0.126}$ & $0.771_{\pm 0.107}$ & $0.956_{\pm 0.015}$ & $0.711_{\pm 0.151}$ & $0.986_{\pm 0.008}$ \\
\midrule
\multirow{3}{*}{SS} & Image & SSVS~\citep{ma2021self} & $0.408_{\pm 0.057}$ & $0.216_{\pm 0.039} $ & $0.355_{\pm 0.046}$ & $0.522_{\pm 0.050}$ & $0.905_{\pm 0.013}$ & $0.471_{\pm 0.056}$ & $0.960_{\pm 0.009}$ \\
 & Image & DARL~\citep{kim2022diffusion} & $0.605_{\pm 0.065}$ & $0.300_{\pm 0.058}$ & $0.464_{\pm 0.064}$ & $0.631_{\pm 0.060}$ & $0.929_{\pm 0.014}$ & $0.547_{\pm 0.060}$ & $0.978_{\pm 0.014}$ \\
 & Image & FreeCOS~\citep{shi2023freecos} & $0.639_{\pm 0.101}$ & $0.461_{\pm 0.087}$ & $0.506_{\pm 0.135}$ & $0.660_{\pm 0.131}$ & $0.941_{\pm 0.015}$ & $0.554_{\pm 0.152}$ & $0.988_{\pm 0.004}$ \\
\midrule
U & Video & \textbf{DeNVeR(Ours}) & $\bf{0.704}_{\pm 0.081}$ & $\bf{0.515}_{\pm 0.101}$ & $\bf{0.584}_{\pm 0.082}$ & $\bf{0.733}_{\pm 0.066}$ & $\bf{0.947}_{\pm 0.014}$ & $\bf{0.656}_{\pm 0.091}$ & $0.985_{\pm 0.006}$ \\
\bottomrule
\end{tabular}%
}
\end{table*}

\begin{figure*}[t]
\small
\centering
\setlength{\tabcolsep}{1pt}
\renewcommand{\arraystretch}{1}
\resizebox{1.0\textwidth}{!}{%
\begin{tabular}{cc}
\includegraphics[width=0.5\textwidth]{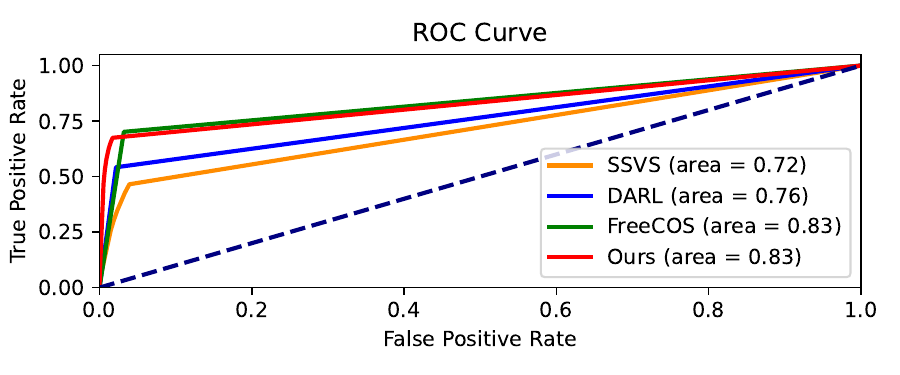}  &
\includegraphics[width=0.5\textwidth]{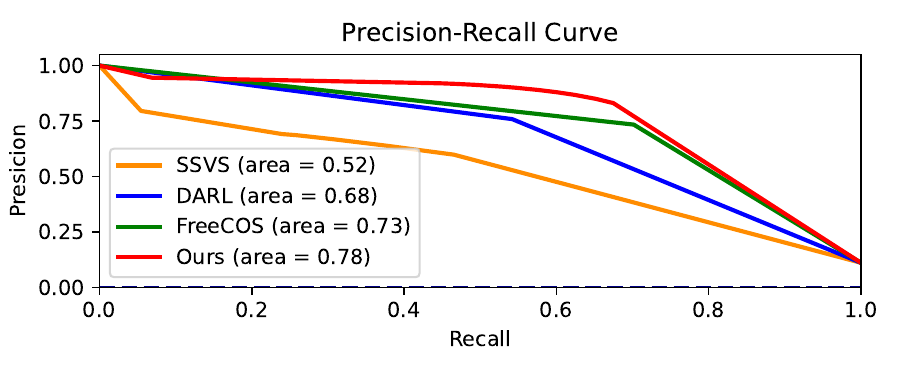}\\
\end{tabular}%
}
\vspace{-4mm}
\caption{\textbf{AUROC and AUPRC results.} Our model performs favorably against other methods on both AUROC and AUPRC.} 
\label{fig:auroc}
\end{figure*}

\subsection{Baseline Methods and Evaluation Metrics}
We compare DeNVeR's performance on the XACV dataset against state-of-the-art methods, including self-supervised (SSVS~\citep{ma2021self}, DARL~\citep{kim2022diffusion}, FreeCOS~\citep{shi2023freecos}), traditional (Hessian~\citep{frangi1998multiscale}), and supervised (U-Net~\citep{ronneberger2015unet}) approaches.
For a fair comparison, we apply the same thresholding and region-growing steps to all methods and optimize heuristic thresholds using Dice scores. 
Following~\citep{ma2021self, kim2022diffusion, shi2023freecos}, we use standard metrics (Jaccard Index, Dice Coefficient, accuracy, sensitivity, specificity) and advanced metrics (NSD~\citep{Reinke_2024}, clDice\citep{cldice2021}, AUROC~\citep{bradley1997use}, AUPRC~\citep{boyd2013area}) for evaluation. 
Due to a lack of publicly available implementations, we couldn't compare with video-based vessel segmentation methods for coronary arteries.

\subsection{Implementation Details}
In this paper, we implement the entire deep learning architecture using PyTorch~\citep{paszke2019pytorch} and train it with Adam optimizer~\citep{kingma2014adam} on a single NVIDIA GeForce RTX 4090 GPU. The entire testing process, including model training and inference, takes approximately 20 minutes and utilizes 18GB of RAM. In the preprocessing stage, we compute the optical flow using RAFT~\citep{teed2020raft}. 

Masks obtained from preprocessing are typically discontinuous and noisy. Therefore, we utilize deep learning methods for training. To simplify the task of vessel segmentation, we divide it into two stages. In stage 1, we use MLPs to acquire the background canonical image, with $\lambda_\text{limit}=0.02$ and $\lambda_\text{smooth}=0.02$. In stage 2, we employ U-Net~\citep{ronneberger2015unet} to predict masks, B-spline models, and foreground canonical images. Initially, we use a warm start U-Net~\citep{ronneberger2015unet} network with $\mathcal{L}_\text{prior}$ to generate a coarse mask, with $\mathcal{L}_\text{prior}$ weight set to 0.5. Then, we gradually incorporate $\mathcal{L}_\text{parallel}$ ($\lambda_\text{parallel}=0.05$), $\mathcal{L}_\text{warp}$ ($\lambda_\text{warp}=0.1$), $\mathcal{L}_\text{mask}$ ($\lambda_\text{mask}=0.1$), and $\mathcal{L}_\text{rec}$ ($\lambda_\text{rec}=0.5$) to optimize DeNVeR. Specifically, our model requires 20 minutes of runtime to process a video sequence of 80 frames. However, our method provides fully automatic segmentation without manual annotations, potentially saving significant time and resources in the long term. 
% We conduct hyperparameter analysis in the supplementary material.

% \subsection{Experimental Results}

\begin{figure*}[t]
\centering
\small
\setlength{\tabcolsep}{1pt}
\renewcommand{\arraystretch}{1}
\resizebox{1.0\textwidth}{!} 
{
\begin{tabular}{cc:c:c:ccc:c}
\includegraphics[width=0.15\textwidth]{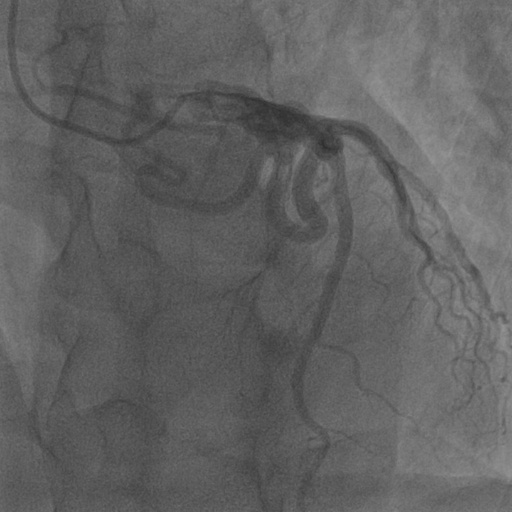} & 
\includegraphics[width=0.15\textwidth]{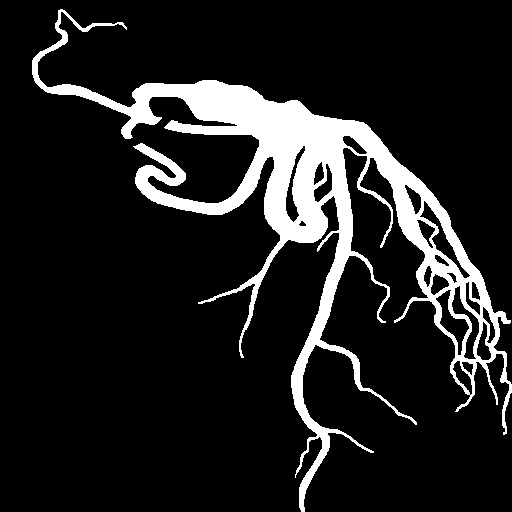} & 
\includegraphics[width=0.15\textwidth]{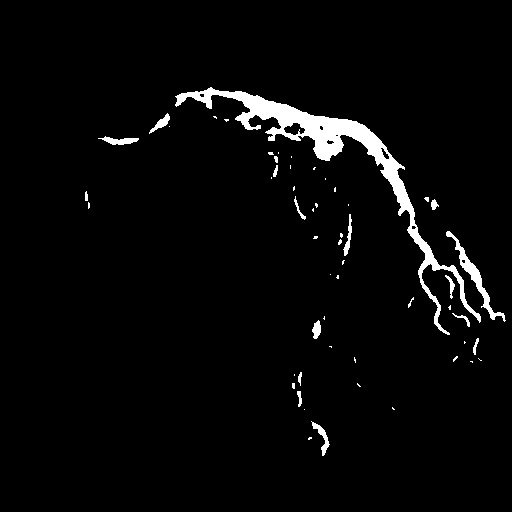} & 
\includegraphics[width=0.15\textwidth]{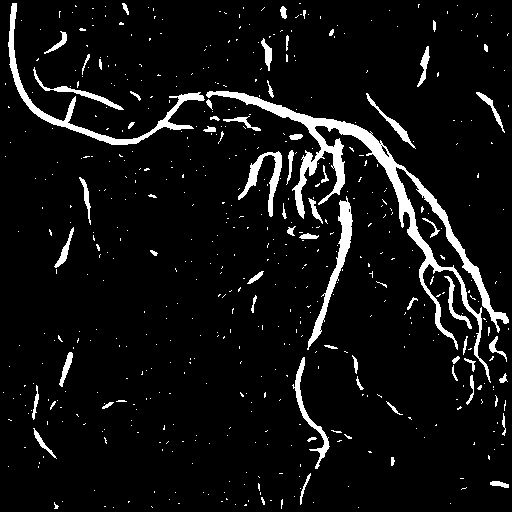} & 
\includegraphics[width=0.15\textwidth]{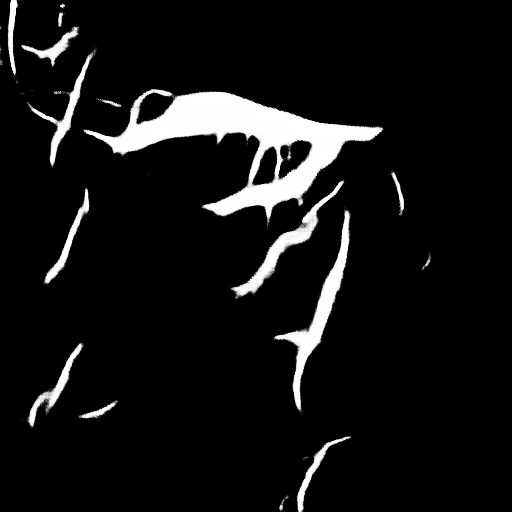} & 
\includegraphics[width=0.15\textwidth]{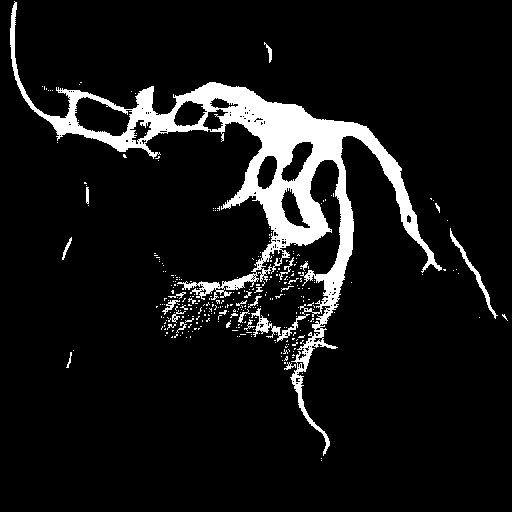} & 
\includegraphics[width=0.15\textwidth]{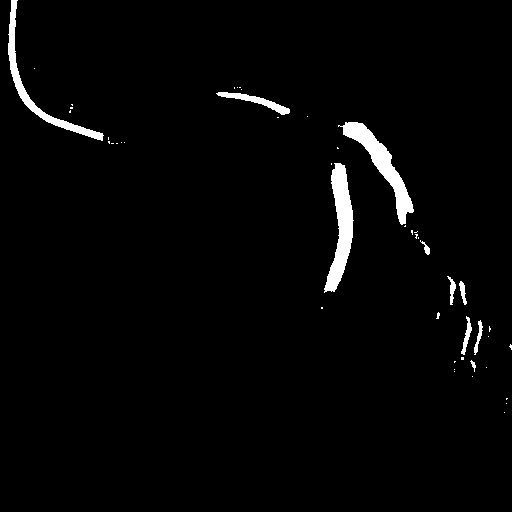} & 
\includegraphics[width=0.15\textwidth]{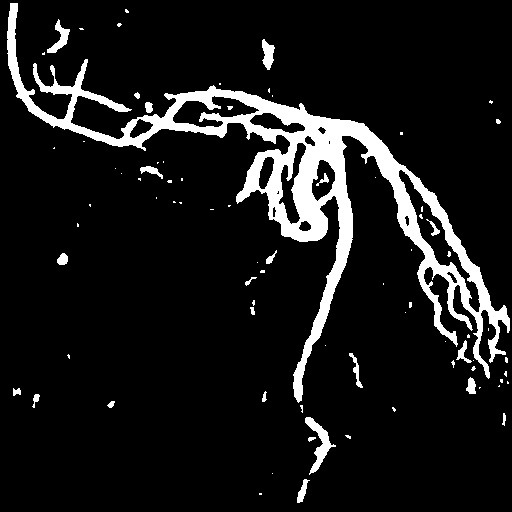} \\
\includegraphics[width=0.15\textwidth]{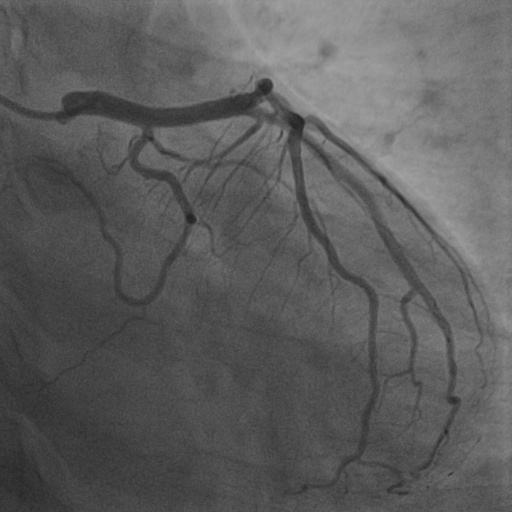} & 
\includegraphics[width=0.15\textwidth]{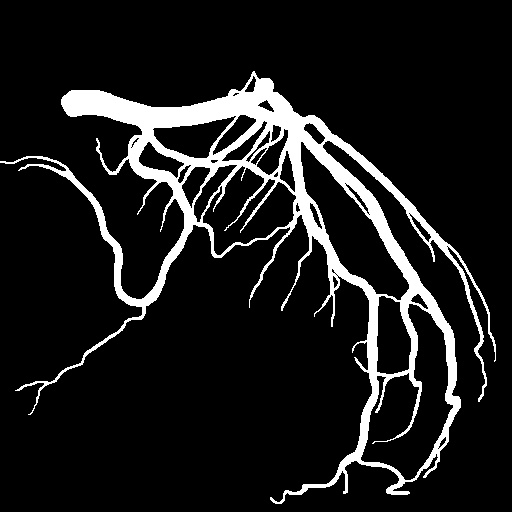} & 
\includegraphics[width=0.15\textwidth]{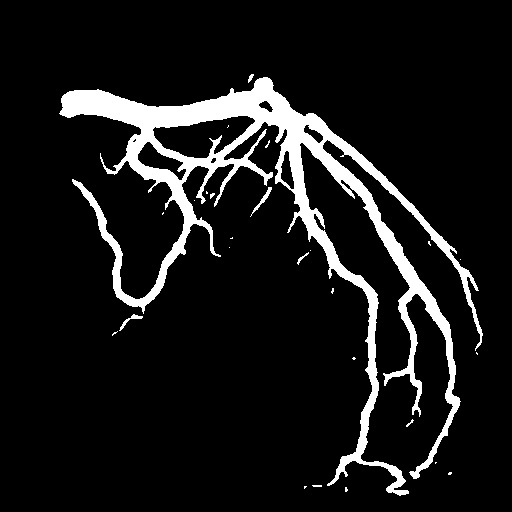} & 
\includegraphics[width=0.15\textwidth]{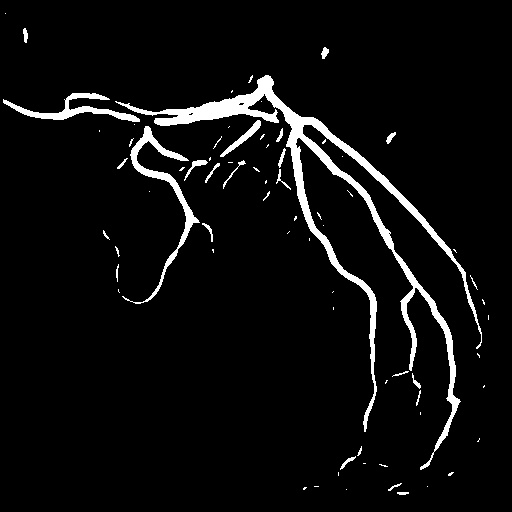} & 
\includegraphics[width=0.15\textwidth]{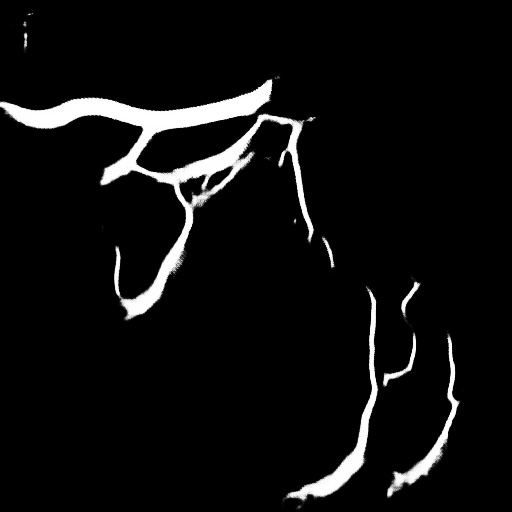} & 
\includegraphics[width=0.15\textwidth]{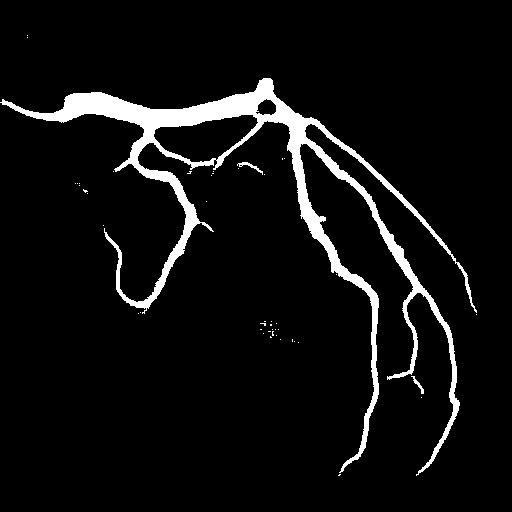} & 
\includegraphics[width=0.15\textwidth]{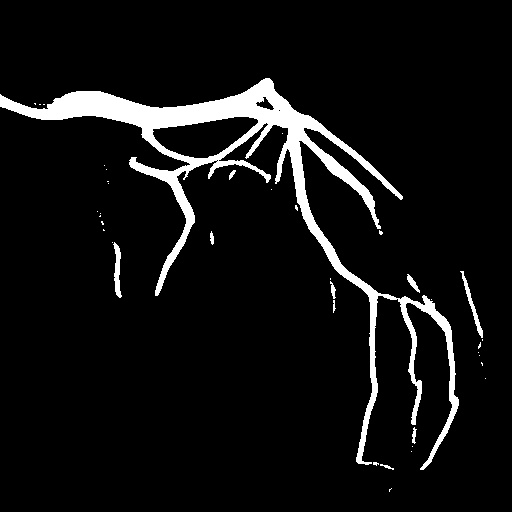} & 
\includegraphics[width=0.15\textwidth]{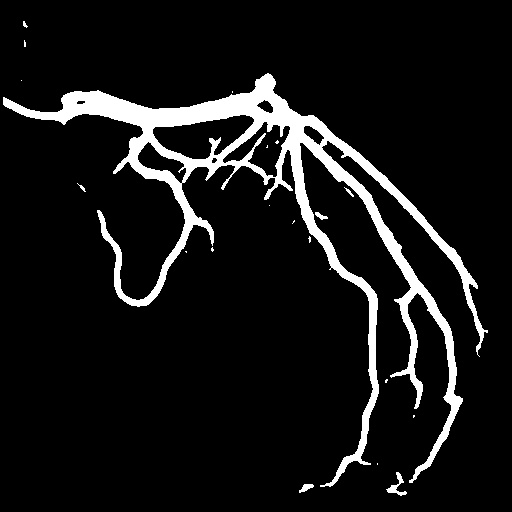} \\
\includegraphics[width=0.15\textwidth]{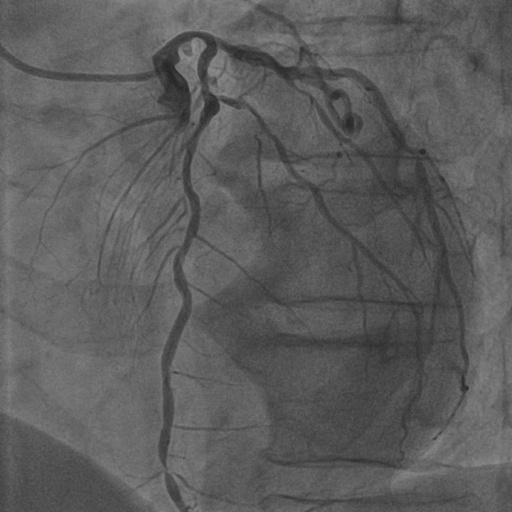} & 
\includegraphics[width=0.15\textwidth]{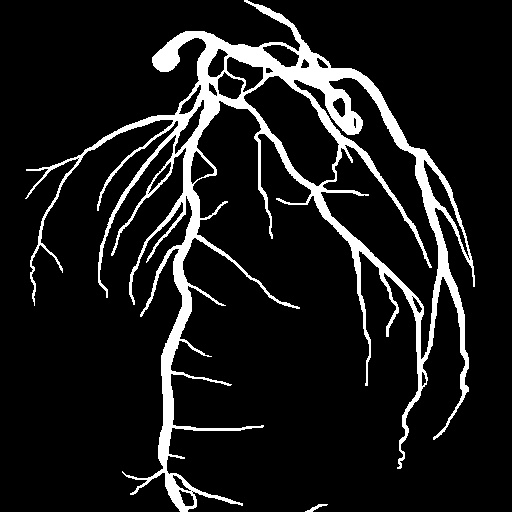} & 
\includegraphics[width=0.15\textwidth]{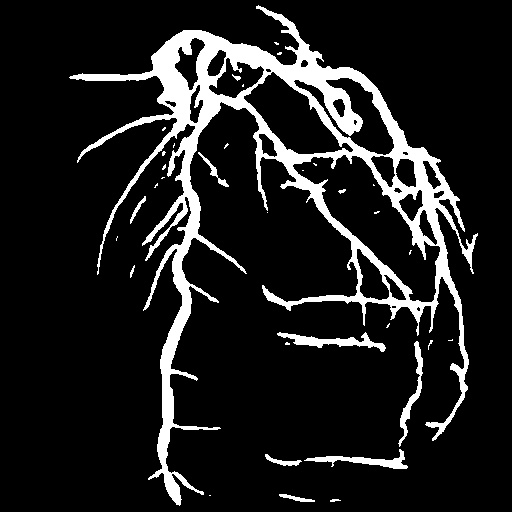} & 
\includegraphics[width=0.15\textwidth]{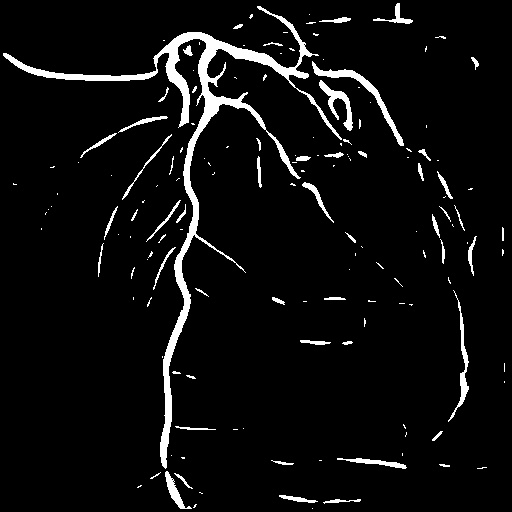} & 
\includegraphics[width=0.15\textwidth]{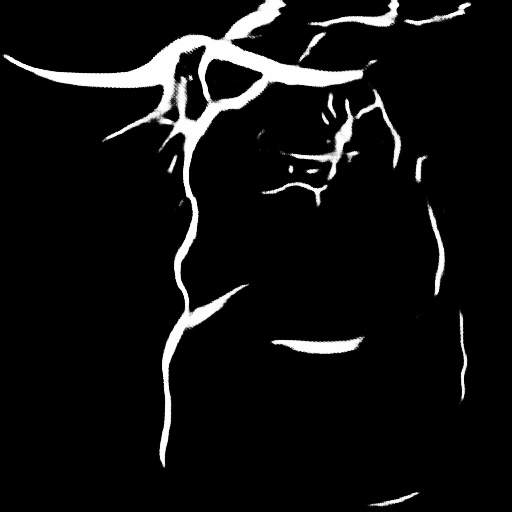} & 
\includegraphics[width=0.15\textwidth]{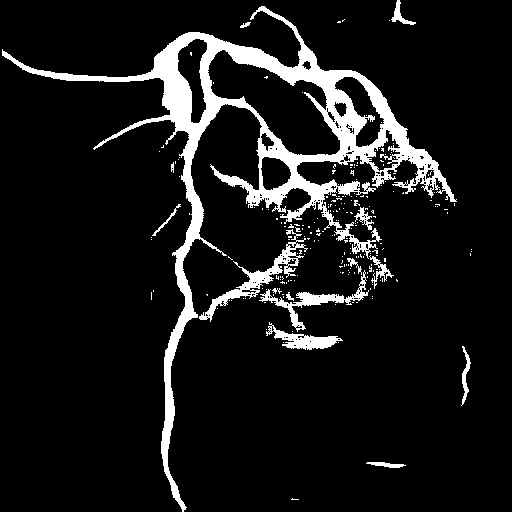} & 
\includegraphics[width=0.15\textwidth]{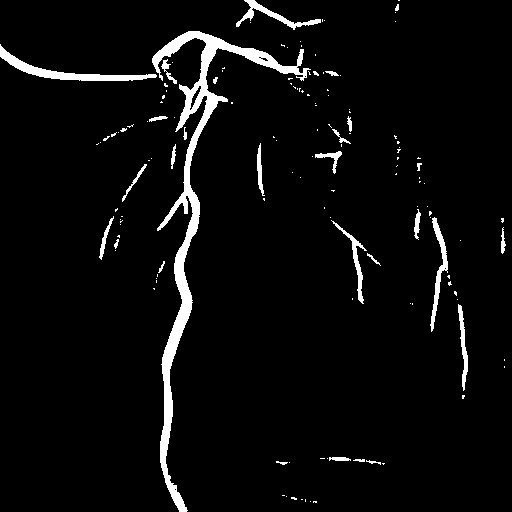} & 
\includegraphics[width=0.15\textwidth]{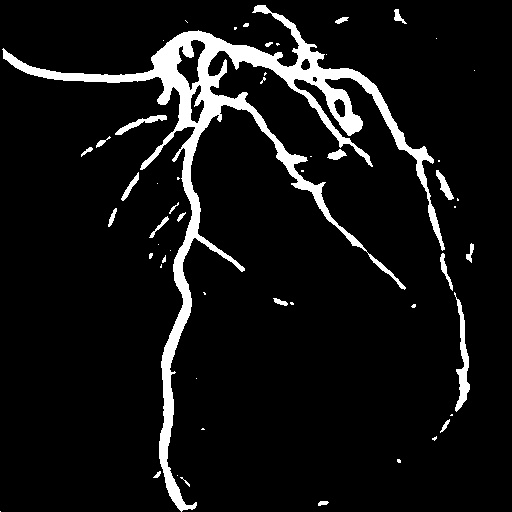} \\
\includegraphics[width=0.15\textwidth]{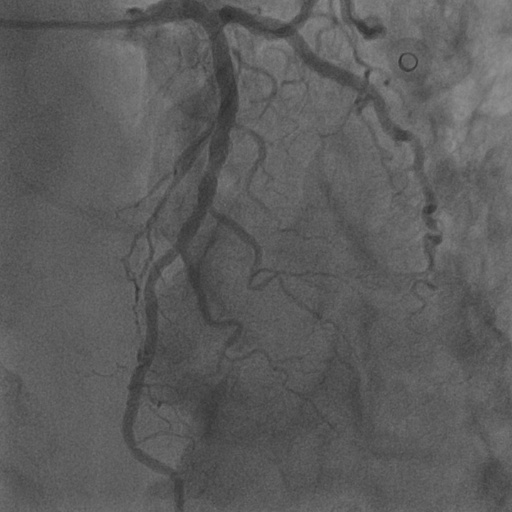} & 
\includegraphics[width=0.15\textwidth]{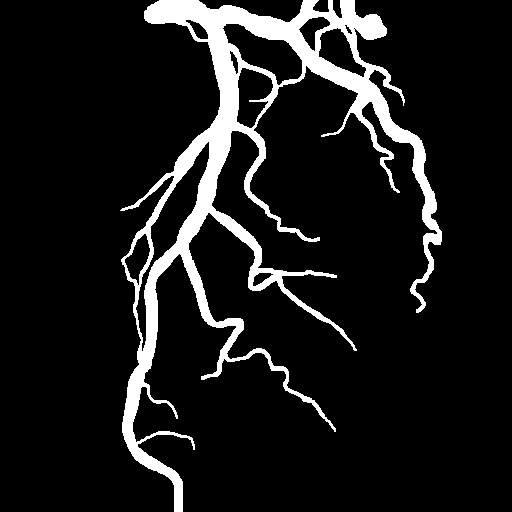} & 
\includegraphics[width=0.15\textwidth]{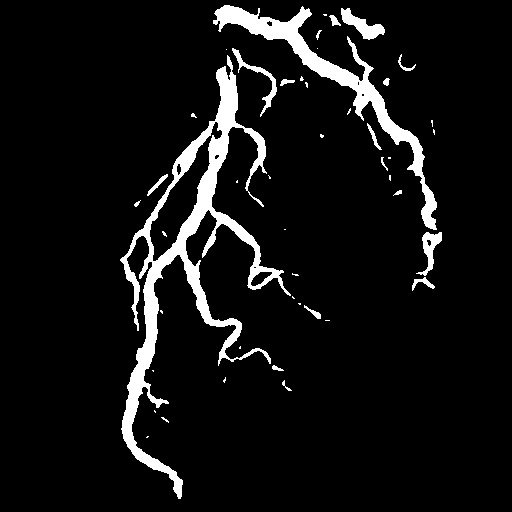} & 
\includegraphics[width=0.15\textwidth]{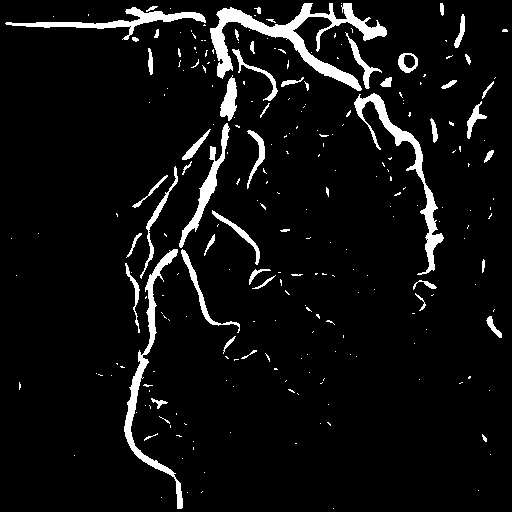} & 
\includegraphics[width=0.15\textwidth]{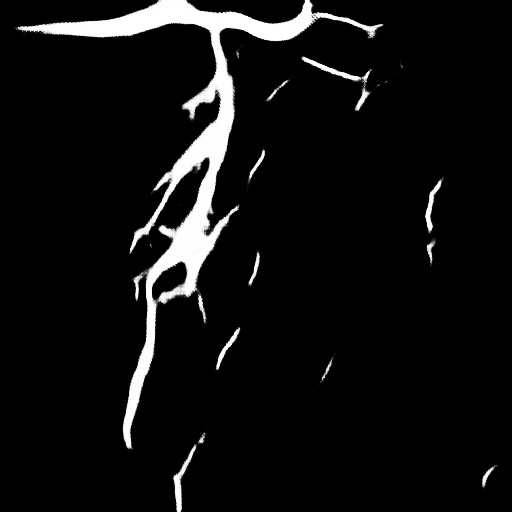} & 
\includegraphics[width=0.15\textwidth]{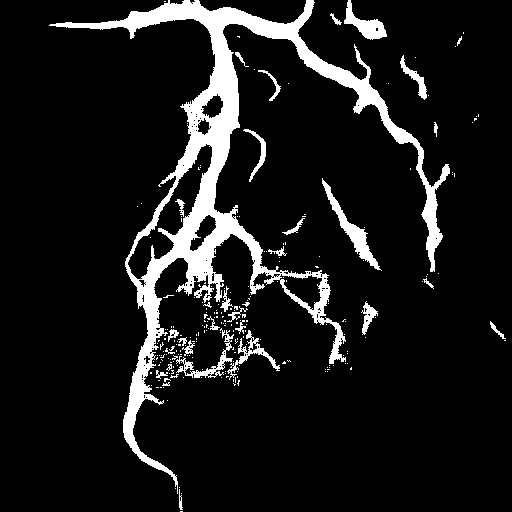} & 
\includegraphics[width=0.15\textwidth]{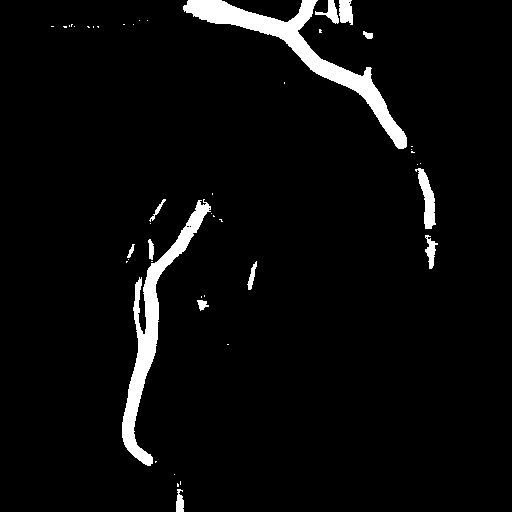} & 
\includegraphics[width=0.15\textwidth]{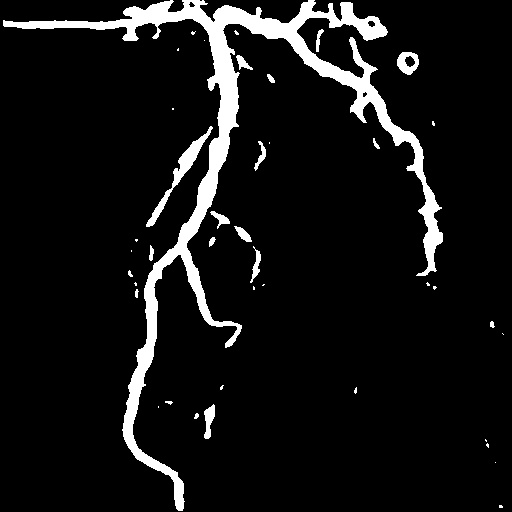} \\
\includegraphics[width=0.15\textwidth]{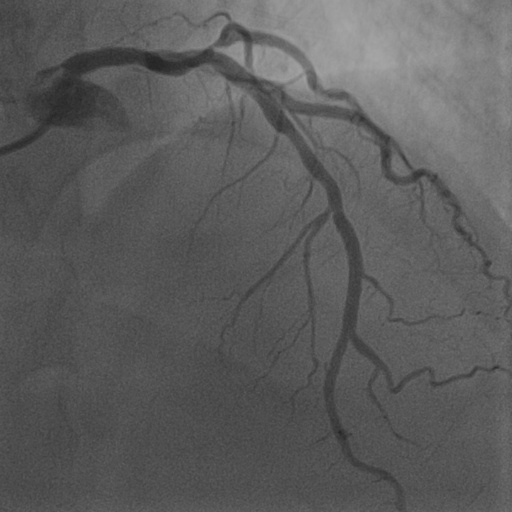} & 
\includegraphics[width=0.15\textwidth]{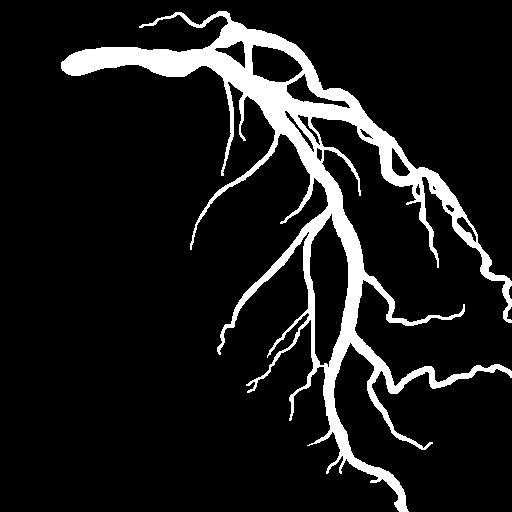} & 
\includegraphics[width=0.15\textwidth]{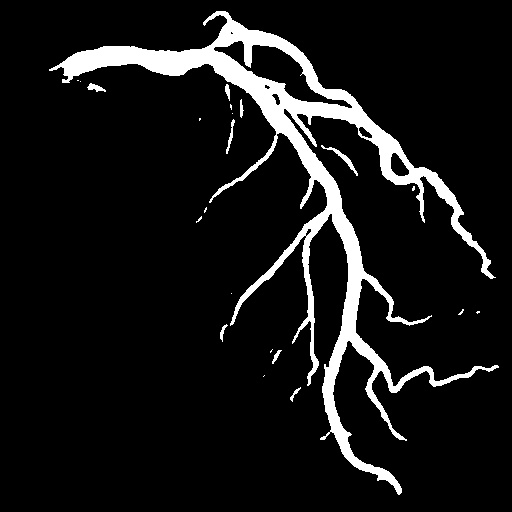} & 
\includegraphics[width=0.15\textwidth]{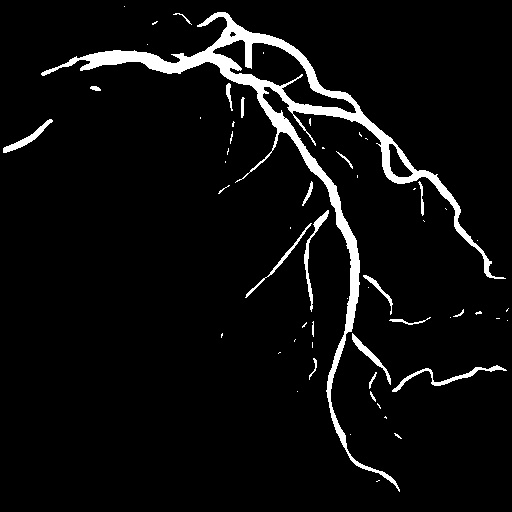} & 
\includegraphics[width=0.15\textwidth]{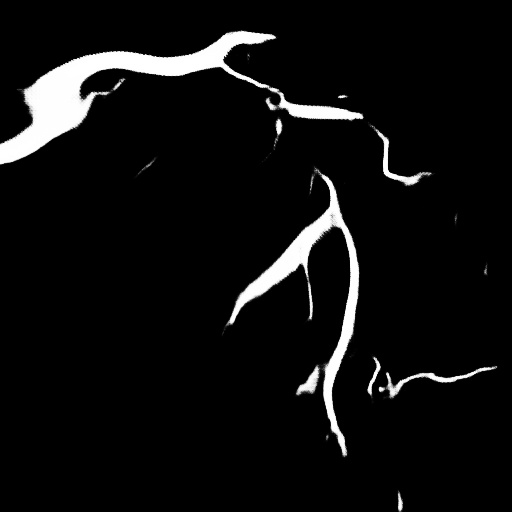} & 
\includegraphics[width=0.15\textwidth]{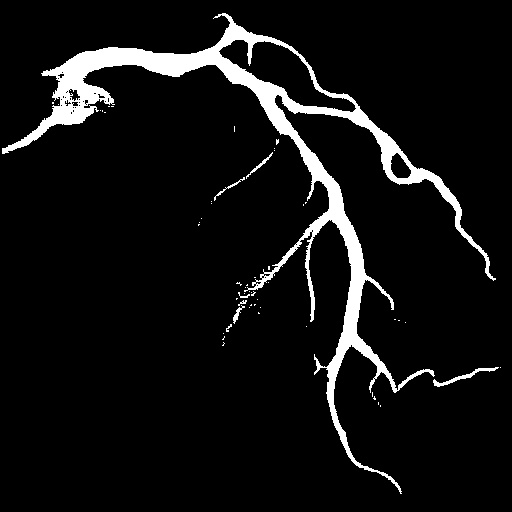} & 
\includegraphics[width=0.15\textwidth]{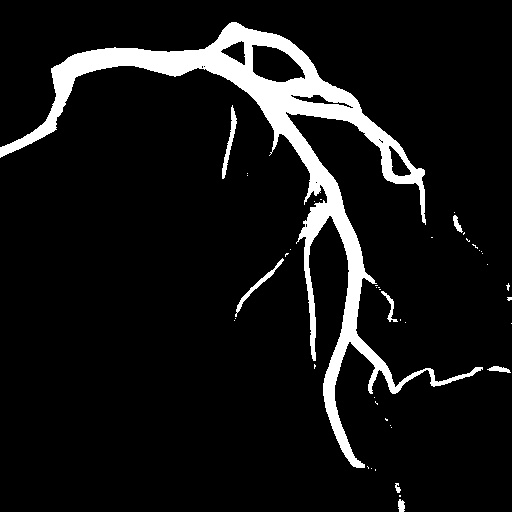} & 
\includegraphics[width=0.15\textwidth]{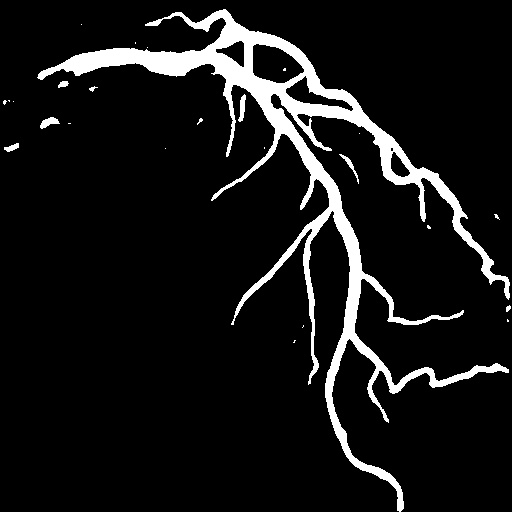} \\
% Image & GT & U-Net (image) & U-Net (video) & Hessian & SSVS & DARL & FreeCOS & \textbf{Ours} \\
% & & \multicolumn{2}{c:}{\textbf{Supervised} (upper bound)} & \textbf{Traditional} & \multicolumn{3}{c:}{\textbf{Self-supervised}} & \textbf{Unsupervised} \\
% Image & GT  & Hessian & SSVS~\citep{ma2021self} & DARL ~\citep{kim2022diffusion}& FreeCOS~\citep{shi2023freecos} & \textbf{Ours} \\
Image & GT & Hessian & U-Net & SSVS & DARL & FreeCOS & \textbf{Ours} \\
&  & \textbf{Traditional} & \textbf{Supervised} & \multicolumn{3}{c:}{\textbf{Self-supervised}} & \textbf{Unsupervised} \\
\end{tabular}%
}
\vspace{-2mm}
\caption{\textbf{Visualization results on the vessel segmentation.}}
% \vspace{-1mm}
\label{fig:qualitative}
\end{figure*}

\begin{figure}[t]
\centering
\small
\setlength{\tabcolsep}{1pt}
\renewcommand{\arraystretch}{1}
\resizebox{1.0\columnwidth}{!}
{
\begin{tabular}{c:c:ccc:c}
\includegraphics[width=0.2\columnwidth]{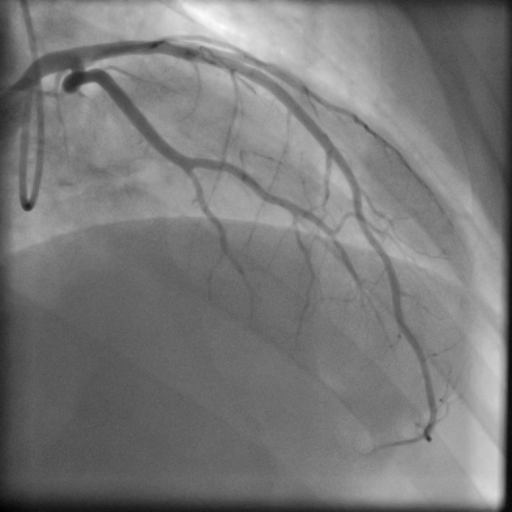} & 
\includegraphics[width=0.2\columnwidth]{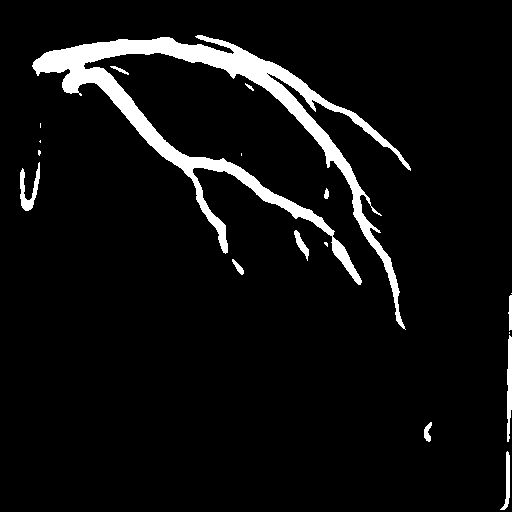} &
\includegraphics[width=0.2\columnwidth]{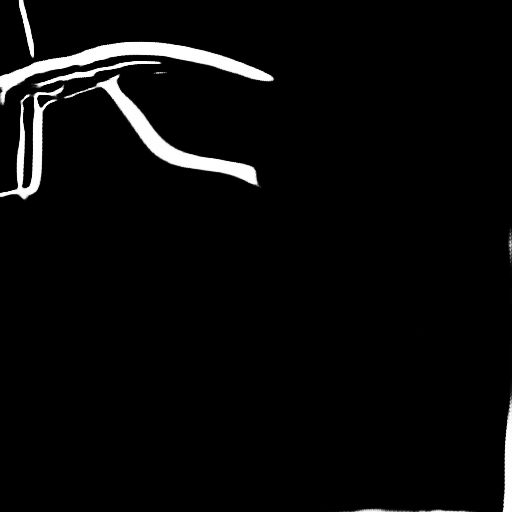} &
\includegraphics[width=0.2\columnwidth]{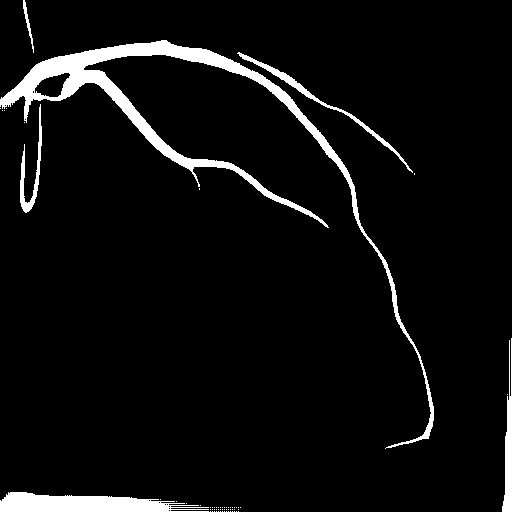} &
\includegraphics[width=0.2\columnwidth]{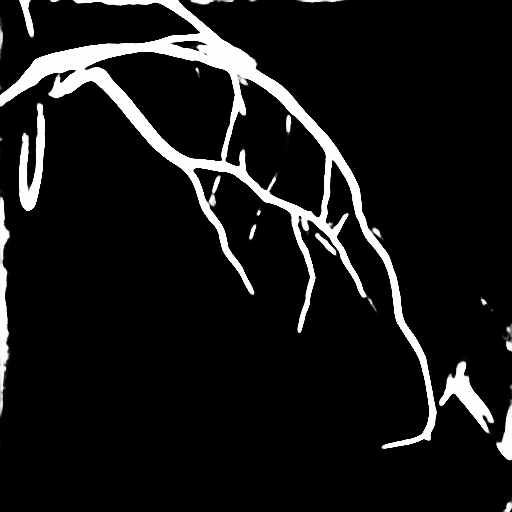} & 
\includegraphics[width=0.2\columnwidth]{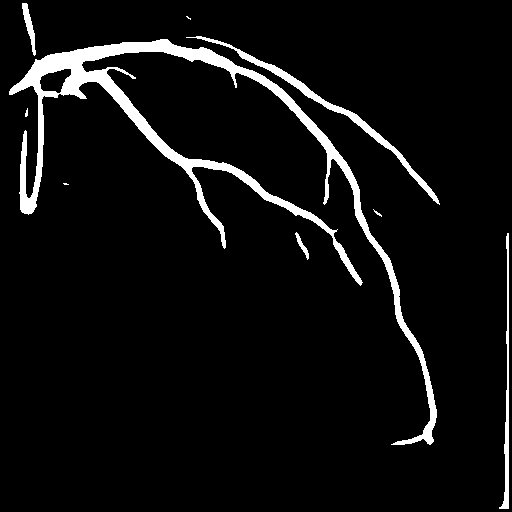} \\
\includegraphics[width=0.2\columnwidth]{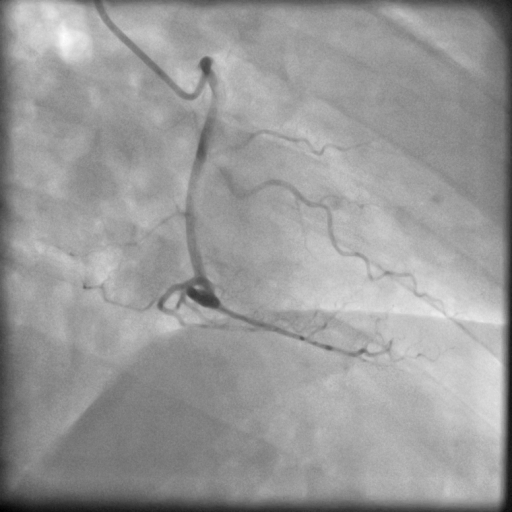} & 
\includegraphics[width=0.2\columnwidth]{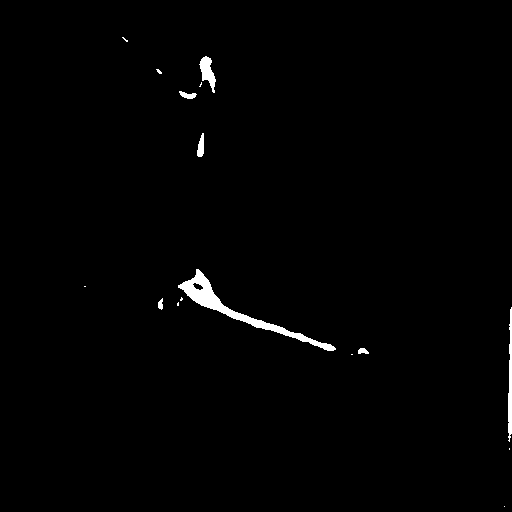} & 
\includegraphics[width=0.2\columnwidth]{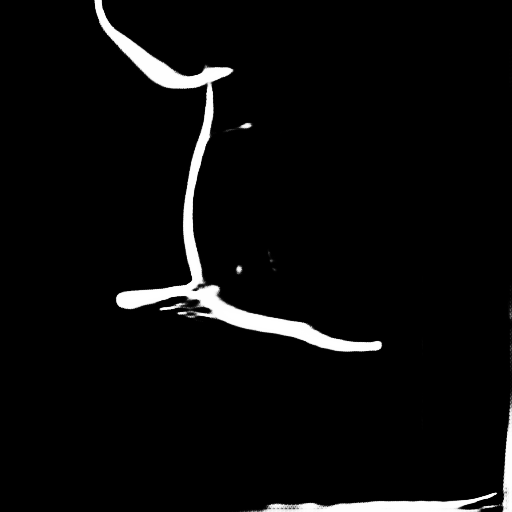} & 
\includegraphics[width=0.2\columnwidth]{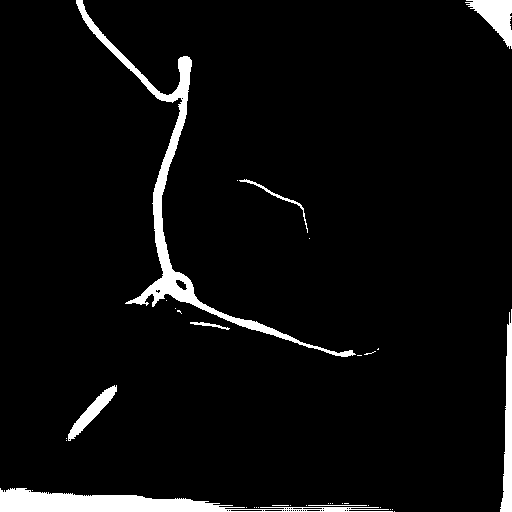} & 
\includegraphics[width=0.2\columnwidth]{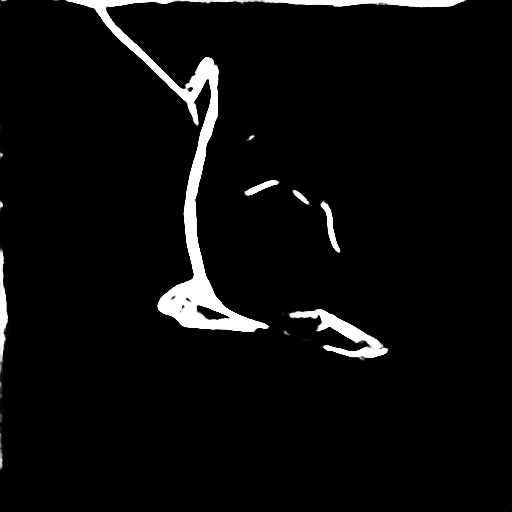} & 
\includegraphics[width=0.2\columnwidth]{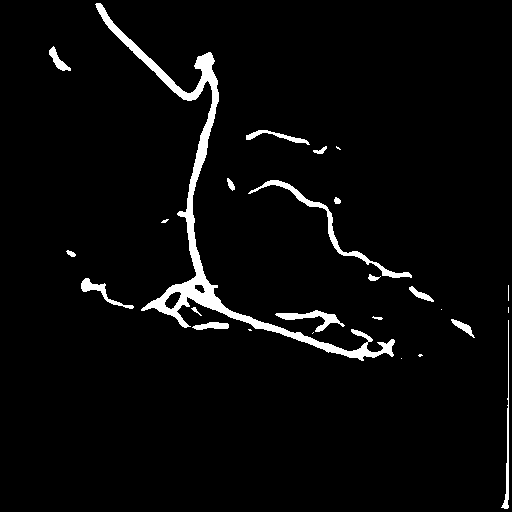} \\
\includegraphics[width=0.2\columnwidth]{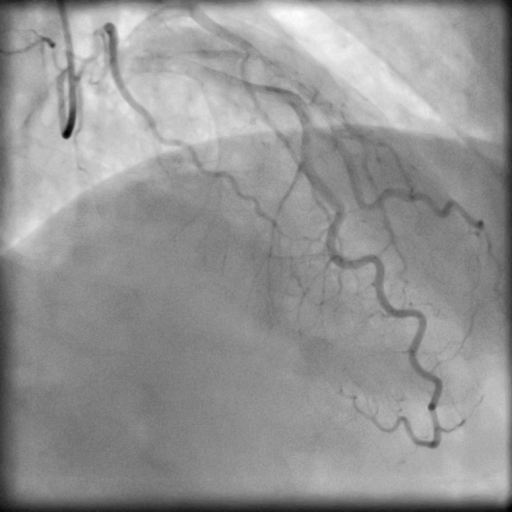} & 
\includegraphics[width=0.2\columnwidth]{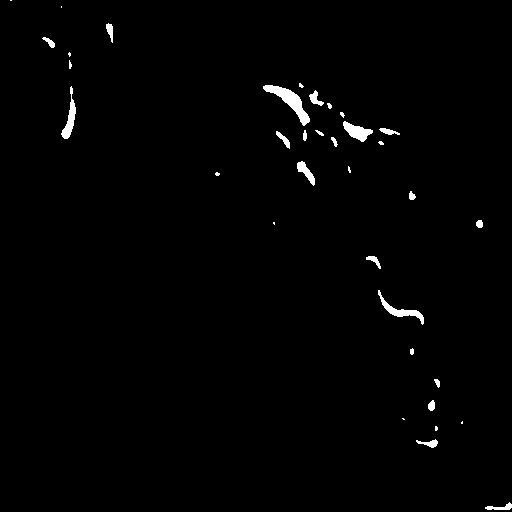} & 
\includegraphics[width=0.2\columnwidth]{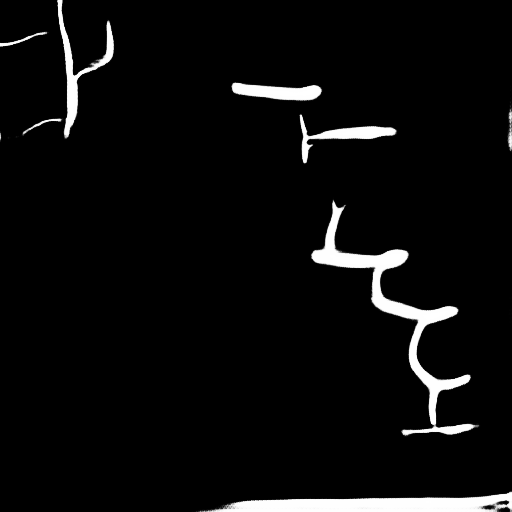} & 
\includegraphics[width=0.2\columnwidth]{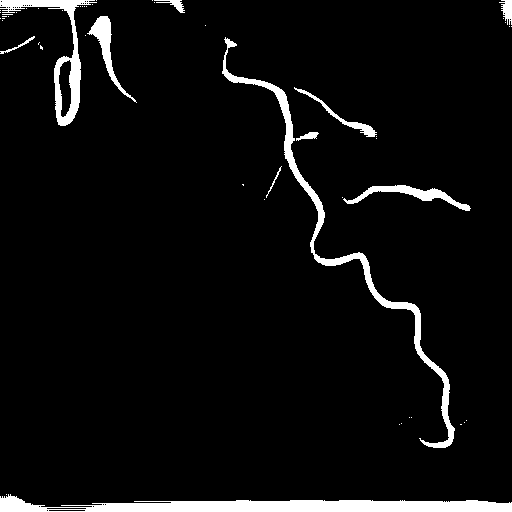} & 
\includegraphics[width=0.2\columnwidth]{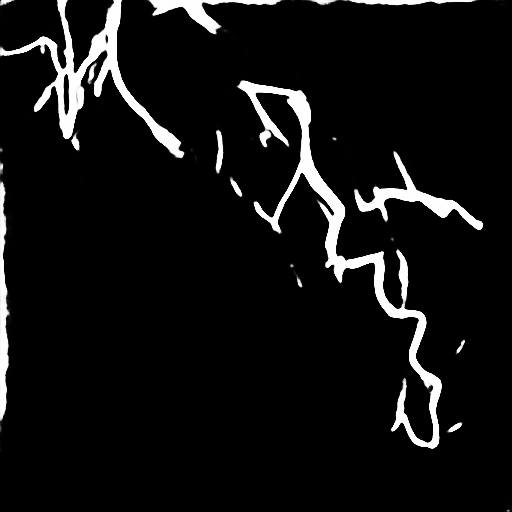} & 
\includegraphics[width=0.2\columnwidth]{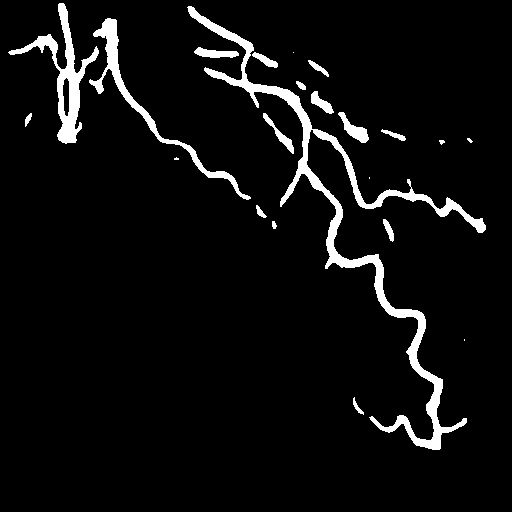} \\
Image & U-Net & SSVS & DARL & FreeCOS & \textbf{Ours} \\
& \textbf{Supervised} & \multicolumn{3}{c:}{\textbf{Self-supervised}} & \textbf{Unsupervised} \\
\end{tabular}%
}
\vspace{-2mm}
\caption{\textbf{Results on CADICA~\citep{jimenez2024cadica} dataset.} Supervised methods cannot generalize well to a new dataset and suffer from the domain gaps between training (our XACV) and testing datasets (CADICA). Our method, although requiring test-time training, can adapt to various datasets in an unsupervised manner. The CADICA dataset does not contain GT and is the only video vessel dataset publicly available. Therefore, we demonstrate qualitative comparisons.
% ~\yulunliu{Need FreeCOS.}
% ~\wujh{top to down : rgb - freecos - ours}
}
% \vspace{-1mm}
\label{fig:cadica_visual_result}
\end{figure}

\begin{figure*}[t]
\centering
% \small
\setlength{\tabcolsep}{1pt}
\renewcommand{\arraystretch}{1}
\resizebox{1.0\textwidth}{!} 
{
\begin{tabular}{ccc:ccc:ccc}
\includegraphics[width=0.125\textwidth]{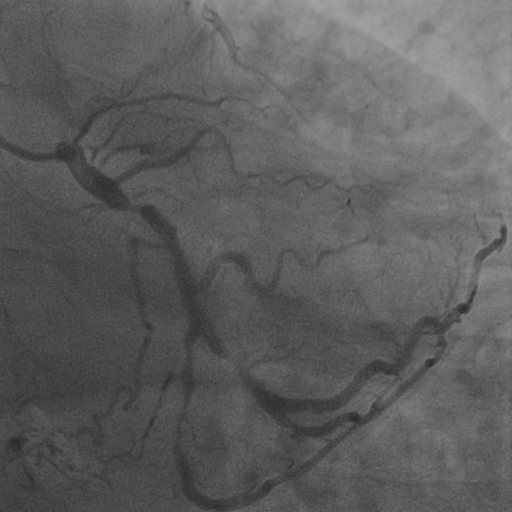} &
\includegraphics[width=0.125\textwidth]{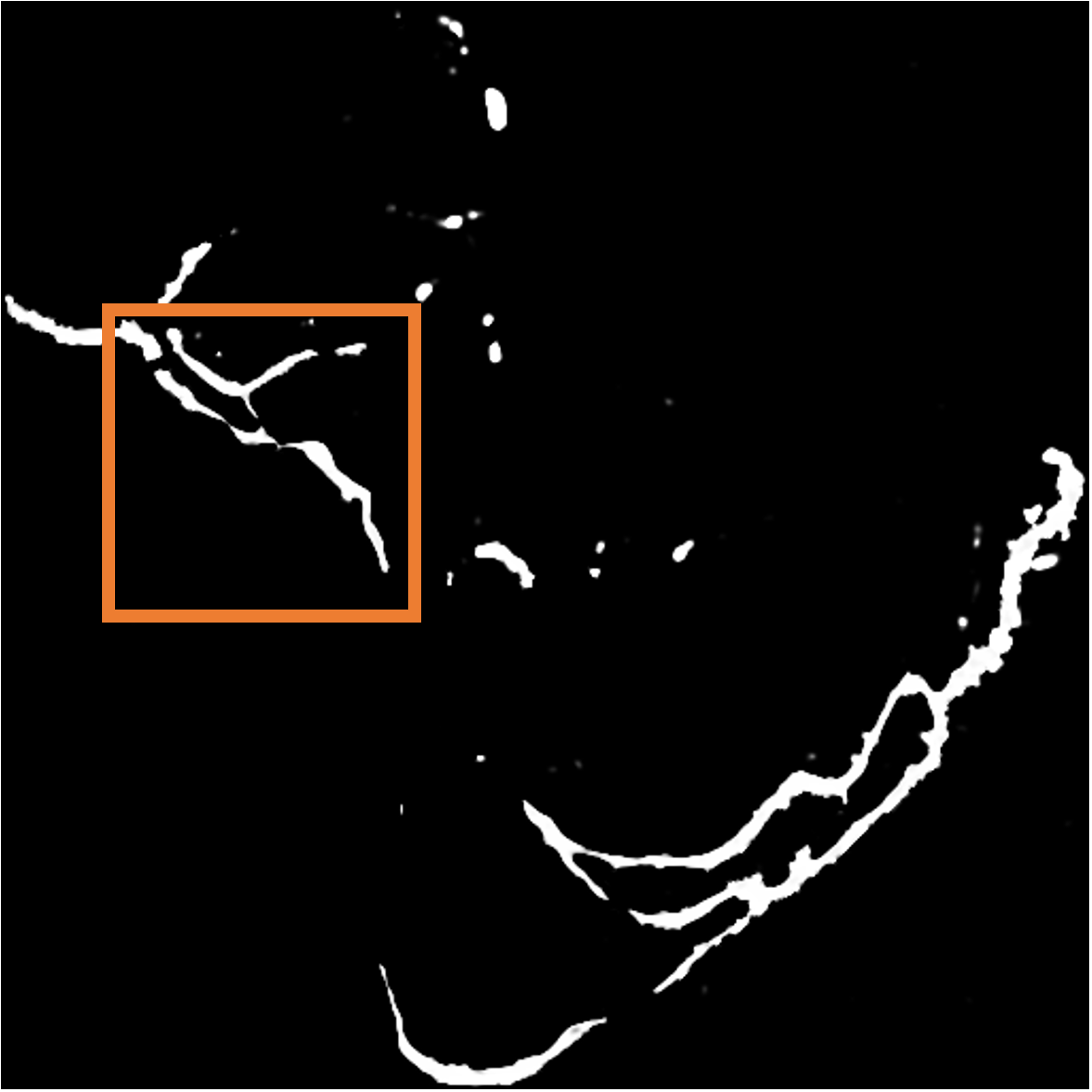} &
\includegraphics[width=0.125\textwidth]{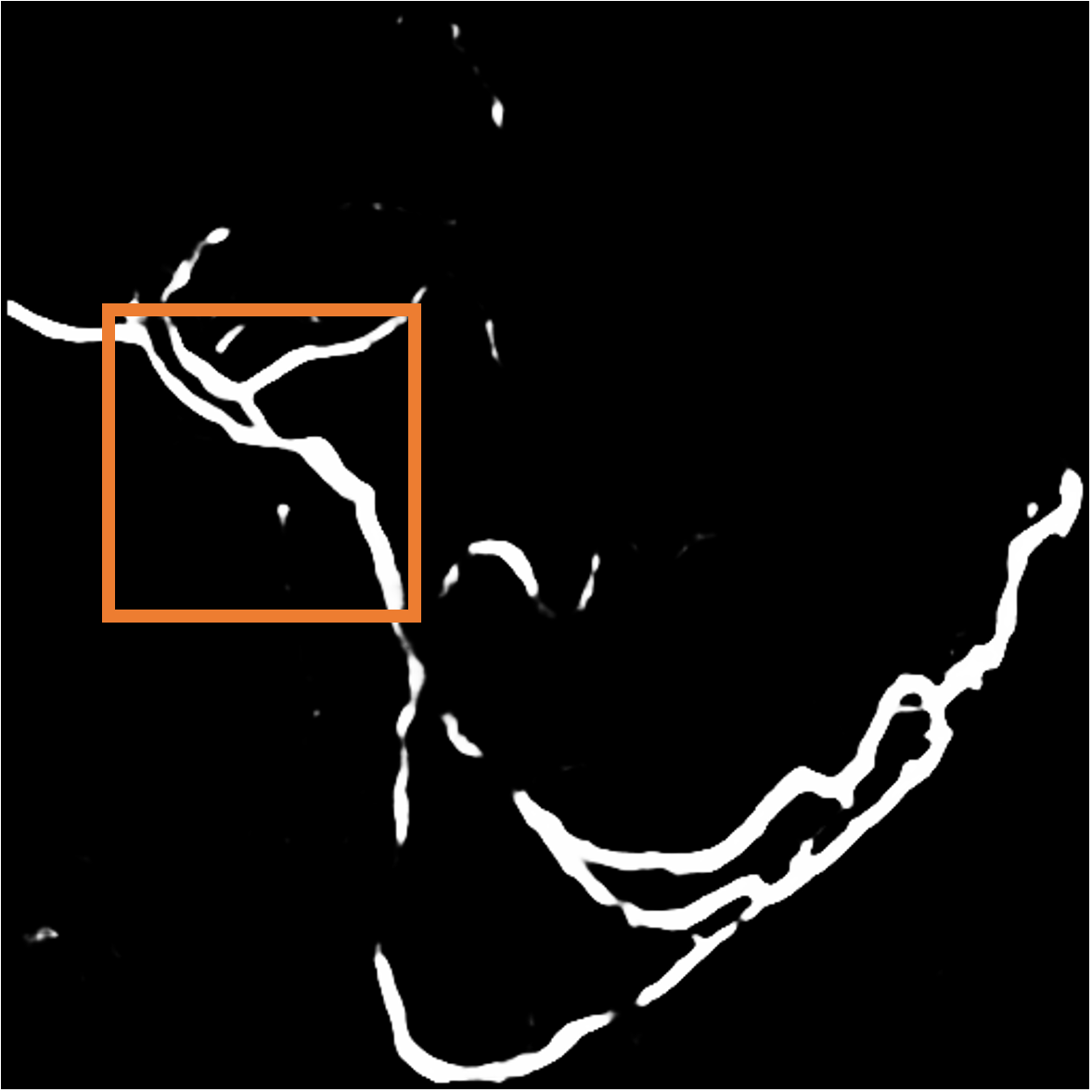} &
\includegraphics[width=0.125\textwidth]{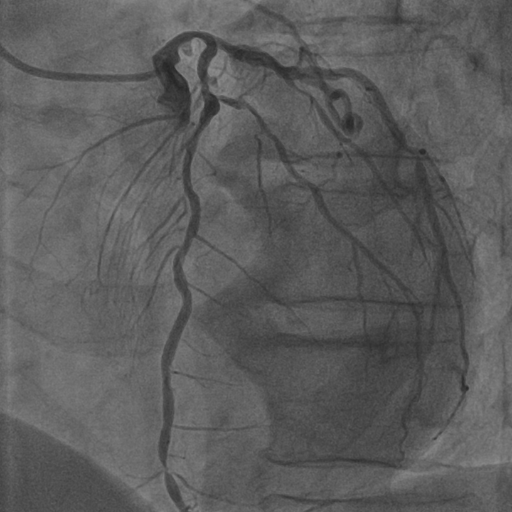} &
\includegraphics[width=0.125\textwidth]{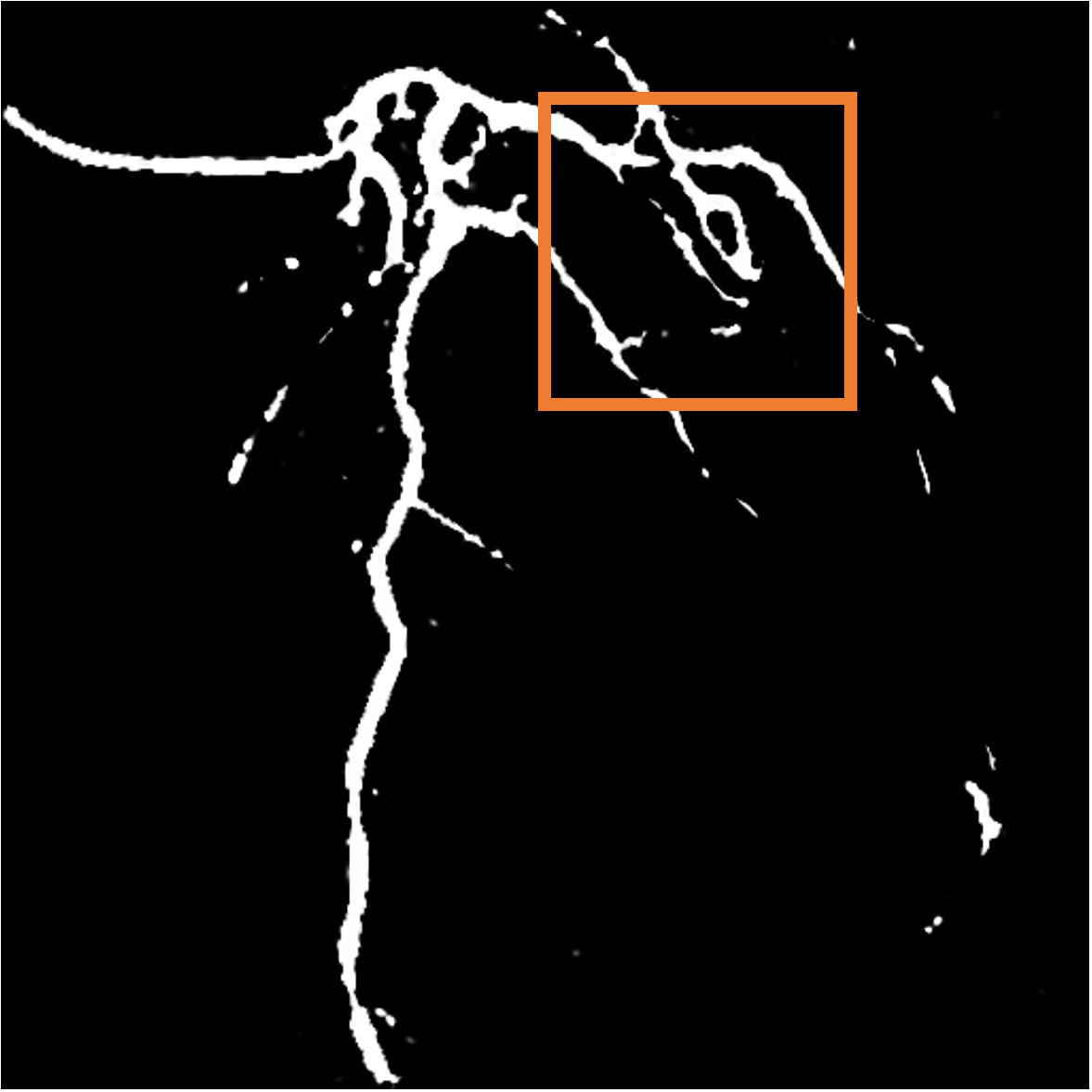} &
\includegraphics[width=0.125\textwidth]{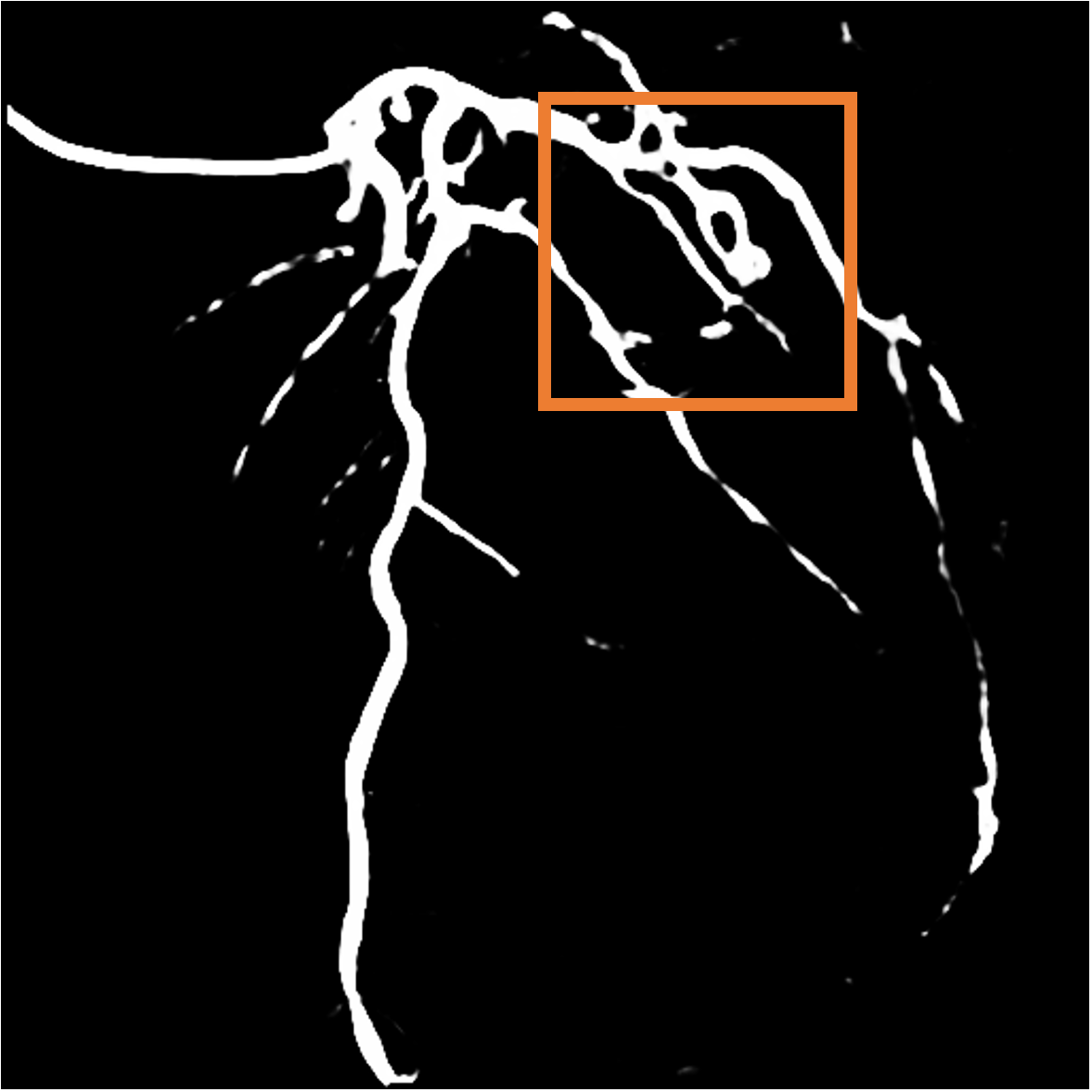} &
\includegraphics[width=0.125\textwidth]{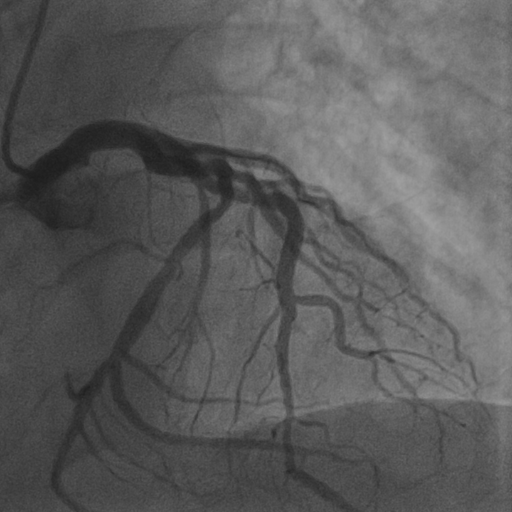} &
\includegraphics[width=0.125\textwidth]{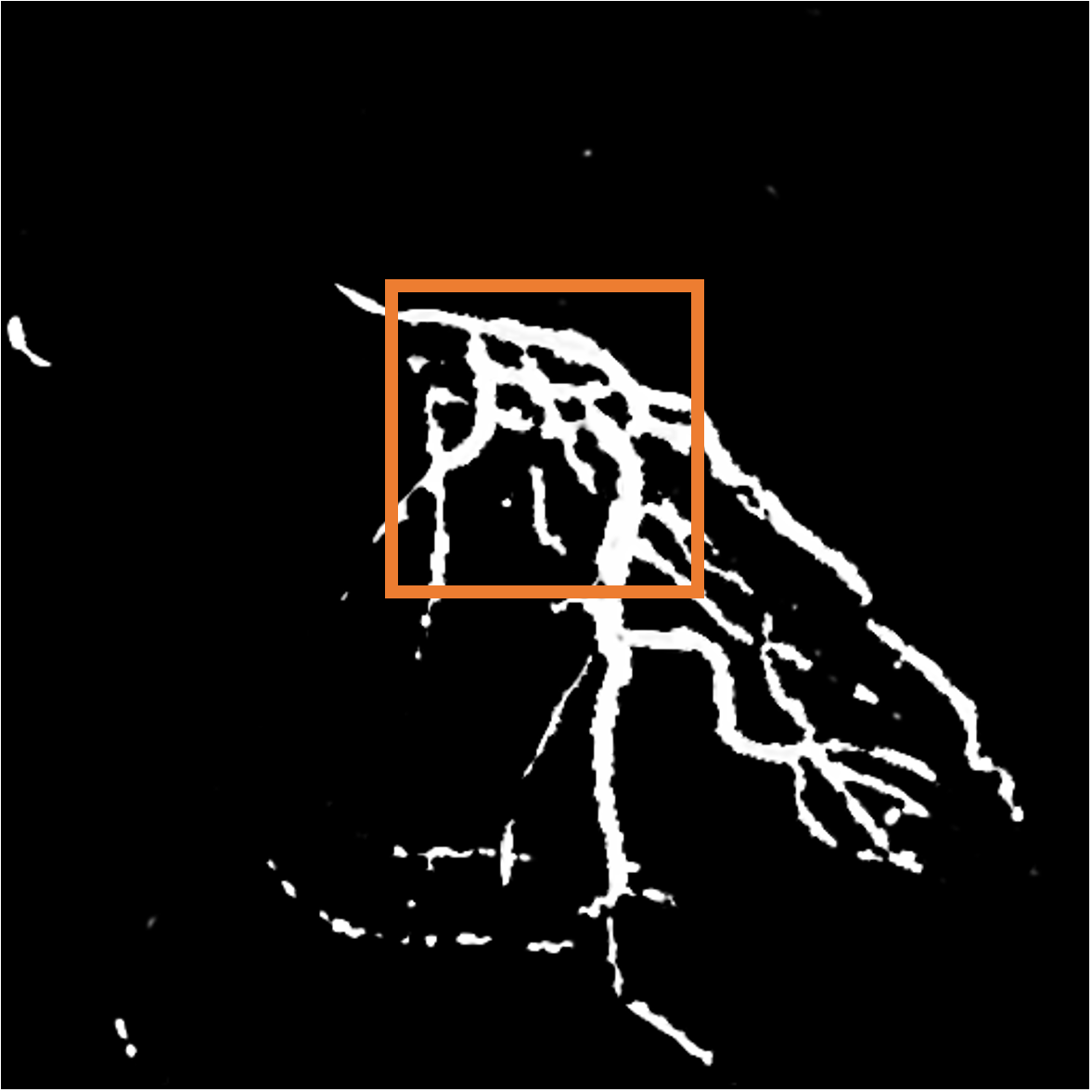} &
\includegraphics[width=0.125\textwidth]{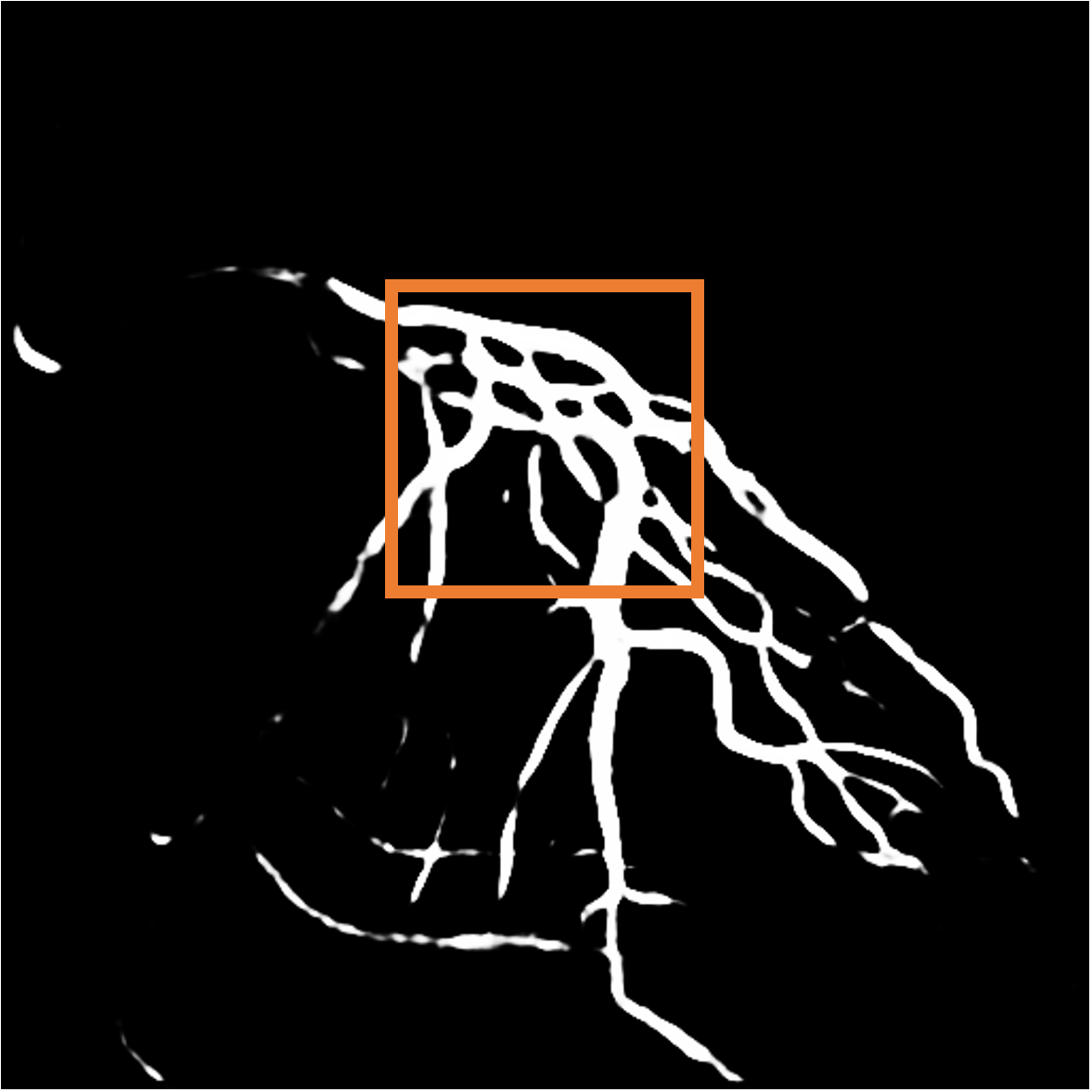} \\
Input frame & \emph{Without} & \emph{With} & Input frame & \emph{Without} & \emph{With} & Input frame & \emph{Without} & \emph{With} \\
\multicolumn{3}{c:}{(a) Layer separation boostrapping} & \multicolumn{3}{c:}{(b) Hessian prior loss $\mathcal{L}_\text{prior}$} & \multicolumn{3}{c}{(c) Parallel vessel motion loss $\mathcal{L}_\text{parallel}$} \\
\end{tabular}%
}
\vspace{-3mm}
\caption{\textbf{Visual comparisons of ablation studies.}}
% \vspace{-1mm}
\label{fig:ablation_result}
\end{figure*}

\subsection{Comparison}
% \wujh{sp score similar problem}
Table~\ref{tab:quantitative} reports the performance of video vessel segmentation on the XACV dataset between our proposed DeNVeR and the baseline methods. Although our method is unsupervised, for comparison with other supervised or self-supervised methods, we still partition the entire dataset into training and testing sets in an 8:2 ratio. All the results recorded in \cref{tab:quantitative} are obtained on the testing set. 

Since supervised training and testing data are from the same dataset (in-domain setting), its performance will be better than that of self-supervised or unsupervised methods. However, it is worth noting that in this scenario, our method does not require any labels and can still outperform existing self-supervised methods. Also, we test the CADICA dataset to compare the generalization ability of supervised training and our proposed unsupervised training in \cref{fig:cadica_visual_result}. We find that supervised methods are limited by the domain of their training data and thus struggle to generalize well. Our method, while requiring test-time training, can adapt to various datasets in an unsupervised manner. This allows for greater flexibility and generalization across different types of vascular video data.

In comparison with the traditional Hessian-based filter, our method achieves a 16.9\% improvement in the Jaccard Index and a 14.9\% increase in the Dice Score, indicating a significant enhancement in performance while utilizing it as a prior. While our method is more complex than supervised approaches, it eliminates the need for costly and time-consuming manual annotations. The test-time training phase, though computationally intensive, is a one-time process per video. For self-supervised methods, we follow their tutorials to augment the training dataset and generate synthetic masks for training. Each model is trained for at least 100 epochs. The results indicate that FreeCOS~\citep{shi2023freecos} performs the best among them, but our approach still shows a 7.7\% improvement in the Jaccard Index and a 7.3\% improvement in the Dice Score compared to it. It is worth noting that, due to the sensitivity of the Hessian-based approach to the chosen threshold and its greater bias, under our intentionally selected optimal conditions, the performance of SSVS may be slightly lower than that of the Hessian-based filter.

We calculate the AUROC and AUPRC in \cref{fig:auroc}. We normalize the model's final layer output to [0, 1] to use it as the probability for calculating AUROC and AUPRC. Our model performs favorably against other methods on both AUROC and AUPRC.
We also provide visual comparison results in \cref{fig:qualitative}, demonstrating that our vessel segmentation results are more accurate, complete, and closer to the ground truth masks. Moreover, in some sequences, our method even performs on par with supervised U-Net~\cite{ronneberger2015unet}, as U-Net might face an overfitting problem with insufficient training data. 
Additionally, we provide visual comparisons on the CADICA~\cite{jimenez2024cadica} dataset, which is also a coronary artery X-ray video dataset but without ground truth labeling. 
\cref{fig:cadica_visual_result} demonstrates that our test-time training scheme generalizes better than existing methods.
Due to space limitations, we provide more visual comparisons in the supplementary material.

\begin{table}[t]
\centering
\small
\caption{\textbf{Ablation study.}}
\vspace{-3mm}
% \wujh{without eulerian dice score $>$ origin model}
% \vspace{1mm}
\label{tab:ablation_study}
\resizebox{\columnwidth}{!}{%
\begin{tabular}{ccc|ccccc}
\toprule
Bootstrap & $\mathcal{L}_\text{prior}$ & $\mathcal{L}_\text{parallel}$ & Jaccard & Dice & Acc. & Sn. & Sp.\\
\midrule
% - & \checkmark & \checkmark & ${0.5776}$ & ${0.7284}$ & ${0.9471}$ & ${0.6514}$ & ${0.9852}$ \\ 
%  \checkmark & - & \checkmark & ${0.5764}$ & ${0.7276}$ & ${0.9469}$ & ${0.6489}$ & $\bf{0.9857}$ \\ 
%  \checkmark & \checkmark & - & ${0.5783}$ & ${0.7292}$ & ${0.9472}$ & ${0.6502}$ & ${0.9856}$ \\ 
%  \checkmark & \checkmark & \checkmark &  $\bf{0.5840}$ & $\bf{0.7339}$ & $\bf{0.9479}$ & $\bf{0.6567}$ & $0.9855$ \\
- & \checkmark & \checkmark & ${0.4821}$ & ${0.6462}$ & ${0.9359}$ & ${0.5722}$ & ${0.9814}$ \\ 
 \checkmark & - & \checkmark & ${0.4630}$ & ${0.6333}$ & ${0.9321}$ & ${0.5319}$ & $\bf{0.9857}$ \\ 
 \checkmark & \checkmark & - & ${0.5394}$ & ${0.6971}$ & ${0.9428}$ & ${0.6049}$ & ${0.9856}$ \\ 
 \checkmark & \checkmark & \checkmark &  $\bf{0.5840}$ & $\bf{0.7339}$ & $\bf{0.9479}$ & $\bf{0.6567}$ & $0.9855$ \\
\bottomrule
\end{tabular}%
}
\end{table}

\begin{figure}[t]
\centering
\small
\setlength{\tabcolsep}{1pt}
\renewcommand{\arraystretch}{1}
\resizebox{1.0\columnwidth}{!}{%
\begin{tabular}{cccc}
\includegraphics[width=0.25\columnwidth]{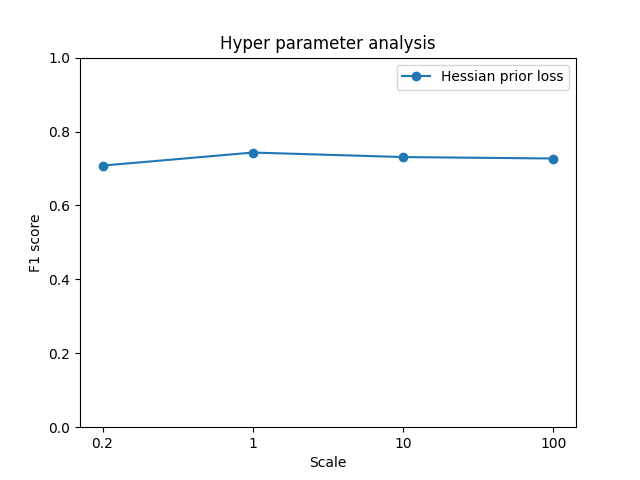} & 
\includegraphics[width=0.25\columnwidth]{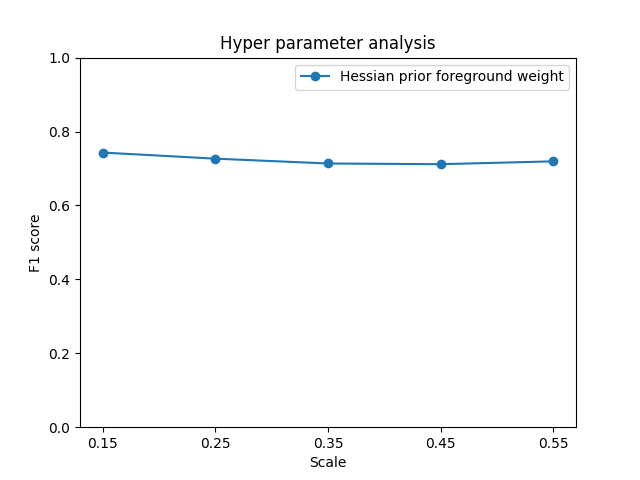} &
\includegraphics[width=0.25\columnwidth]{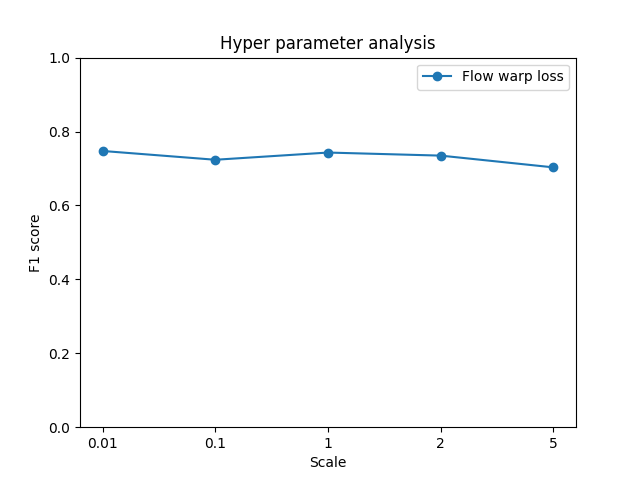} & 
\includegraphics[width=0.25\columnwidth]{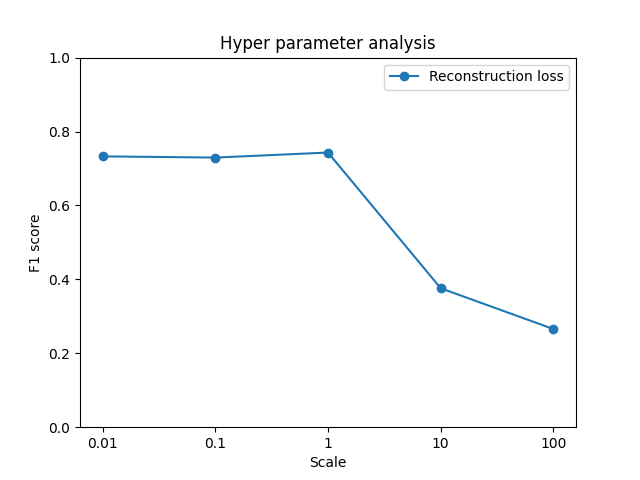} \\
Hessian prior & Hessian prior & Flow warp & Reconstruct \\
loss $\lambda_\text{prior}$ & foreground weight $\alpha$ & loss $\lambda_\text{warp}$ & loss $\lambda_\text{rec}$ \\
\end{tabular}%
}
\vspace{-3mm}
\caption{\textbf{Hyperparameters sensitivity analysis.} Including weights of various losses. Our method is robust and not sensitive to different hyperparameters.}
\label{fig:hyper}
\end{figure}

\subsection{Ablation Study}

\noindent {\bf Layer separation bootstrapping.}
To validate the effectiveness of the layer separation bootstrapping, we train foreground and background canonical images using the same representation. The results are shown in \cref{tab:ablation_study}, where optimizing both foreground and background canonical images simultaneously leads to a decrease in the Dice score by 0.0877. The comparison is shown in \cref{fig:ablation_result} (a), where the orange area indicates the difference between without and with Layer separation bootstrapping. The bottom-right corner shows a zoom-in patch, highlighting the significant effect of the bootstrapping step.

\vspace{3pt}  \noindent {\bf Hessian prior loss $\mathcal{L}_\text{prior}$.}
To test the effect of the Hessian prior loss, we remove the Hessian prior loss. As a result, the segmentation performance, as shown in \cref{tab:ablation_study}, also decreases the Dice score by 0.1006. The comparison between the without and with $\mathcal{L}_\text{prior}$ is shown in \cref{fig:ablation_result} (b), where the orange area indicates the difference between them. The zoom-in patch shows that our model predicts less noticeable vascular regions incorporating the Hessian prior loss $\mathcal{L}_\text{prior}$.

\vspace{3pt}  \noindent {\bf Parallel vessel motion loss $\mathcal{L}_\text{parallel}$.}
We conduct an experiment to assess the effect of the parallel vessel motion loss by removing it from the training pipeline.
% To assess the impact of the parallel loss, we experiment by removing it. 
As shown in Table \ref{tab:ablation_study}, the segmentation performance decreases the Dice score by 0.0368. Without this loss to enforce parallelism between blood and vessels, the segmentation results are negatively affected. In addition, the comparison between without and with $\mathcal{L}_\text{parallel}$ is shown in \cref{fig:ablation_result} (c). The zoom-in patch shows that the image with $\mathcal{L}_\text{parallel}$ has clearer segmented vascular regions.

The improvements from individual components may appear marginal. However, their cumulative effect leads to overall superior performance compared to baselines. In \cref{fig:ablation_result}, we provide visual comparisons of ablation studies, demonstrating that these components are essential for clear and complete vessel segmentations. These components help connect disconnected or over-segmented vessels in specific cases.

\vspace{3pt}  \noindent {\bf Hyperparameter analysis.}
We analyze model sensitivity to key loss weights: Hessian prior ($\lambda_\text{prior}$), prior foreground ($\alpha$), flow warp ($\lambda_\text{warp}$), and reconstruction ($\lambda_\text{rec}$). \cref{fig:hyper} shows F1 score variations for each parameter while fixing others, demonstrating robust performance across different settings.
\section{Conclusion}
\label{sec:conclusion}

% In conclusion, this paper introduces a novel unsupervised approach for video vessel segmentation, leveraging the novel DeNVeR framework. Demonstrating superior performance against existing self-supervised methods, even without any training images or videos, paves the way for significant advancements in medical imaging analysis. The work addresses critical challenges in X-ray angiography, offering a robust solution for enhancing disease diagnosis and treatment planning. Future directions include refining the model to tackle more complex imaging scenarios, potentially broadening its applicability across various medical disciplines. This research holds promise for significantly impacting patient care and medical research.~\yulunliu{Rewrite.}
This paper introduces DeNVeR, an unsupervised test-time training framework for vessel segmentation in X-ray video data. DeNVeR utilizes optical flow and layer separation techniques to accurately segment vessels without requiring annotated datasets.
Quantitative and qualitative evaluations on the XACV and CADICA datasets show that DeNVeR outperforms existing image-based self-supervised methods, offering precise delineation of vessel boundaries critical for medical diagnosis and treatment.

\noindent {\bf Limitations.}
DeNVeR's current limitations include sensitivity to preprocessing filters that may misidentify non-vessel structures, significant computational overhead requiring 20 minutes to process 80 frames, and restriction to contrast-enhanced vessel videos, making it unsuitable for static vessel images like retinal scans.
% Our unsupervised method is sensitive to preprocessing filters, potentially misidentifying non-vascular structures as vessels. DeNVeR's runtime (20 minutes for 80 frames) and computational requirements are also limiting factors. Additionally, our motion-based approach does not apply to datasets without contrast agent flow, such as retinal vessel images.

\newpage
% \clearpage

\paragraph{Acknowledgements.}
This research was funded by the National Science and Technology Council, Taiwan, under Grants NSTC 112-2222-E-A49-004-MY2 and 113-2628-E-A49-023-. The authors are grateful to Google, NVIDIA, and MediaTek Inc. for their generous donations. Yu-Lun Liu acknowledges the Yushan Young Fellow Program by the MOE in Taiwan.

{
    \small
    \bibliographystyle{ieeenat_fullname}
    \bibliography{main}

\begin{thebibliography}{76}
\providecommand{\natexlab}[1]{#1}
\providecommand{\url}[1]{\texttt{#1}}
\expandafter\ifx\csname urlstyle\endcsname\relax
  \providecommand{\doi}[1]{doi: #1}\else
  \providecommand{\doi}{doi: \begingroup \urlstyle{rm}\Url}\fi

\bibitem[Abdal et~al.(2021)Abdal, Zhu, Mitra, and Wonka]{abdal2021labels4free}
Rameen Abdal, Peihao Zhu, Niloy~J Mitra, and Peter Wonka.
\newblock Labels4free: Unsupervised segmentation using stylegan.
\newblock In \emph{ICCV}, 2021.

\bibitem[Alonso et~al.(2021)Alonso, Sabater, Ferstl, Montesano, and Murillo]{alonso2021semi}
Inigo Alonso, Alberto Sabater, David Ferstl, Luis Montesano, and Ana~C Murillo.
\newblock Semi-supervised semantic segmentation with pixel-level contrastive learning from a class-wise memory bank.
\newblock In \emph{ICCV}, 2021.

\bibitem[Bar et~al.(2022)Bar, Wang, Kantorov, Reed, Herzig, Chechik, Rohrbach, Darrell, and Globerson]{bar2022detreg}
Amir Bar, Xin Wang, Vadim Kantorov, Colorado~J Reed, Roei Herzig, Gal Chechik, Anna Rohrbach, Trevor Darrell, and Amir Globerson.
\newblock Detreg: Unsupervised pretraining with region priors for object detection.
\newblock In \emph{CVPR}, 2022.

\bibitem[Benaim et~al.(2020)Benaim, Ephrat, Lang, Mosseri, Freeman, Rubinstein, Irani, and Dekel]{benaim2020speednet}
Sagie Benaim, Ariel Ephrat, Oran Lang, Inbar Mosseri, William~T Freeman, Michael Rubinstein, Michal Irani, and Tali Dekel.
\newblock Speednet: Learning the speediness in videos.
\newblock In \emph{CVPR}, 2020.

\bibitem[Black and Anandan(1991)]{black1991robust}
Michael~J Black and Padmanabhan Anandan.
\newblock Robust dynamic motion estimation over time.
\newblock In \emph{CVPR}, 1991.

\bibitem[Boyd et~al.(2013)Boyd, Eng, and Page]{boyd2013area}
Kendrick Boyd, Kevin~H Eng, and C~David Page.
\newblock Area under the precision-recall curve: point estimates and confidence intervals.
\newblock In \emph{Machine Learning and Knowledge Discovery in Databases: European Conference, ECML PKDD 2013, Prague, Czech Republic, September 23-27, 2013, Proceedings, Part III 13}, 2013.

\bibitem[Bradley(1997)]{bradley1997use}
Andrew~P Bradley.
\newblock The use of the area under the roc curve in the evaluation of machine learning algorithms.
\newblock \emph{Pattern recognition}, 1997.

\bibitem[Brox and Malik(2010)]{brox2010object}
Thomas Brox and Jitendra Malik.
\newblock Object segmentation by long term analysis of point trajectories.
\newblock In \emph{ECCV}, 2010.

\bibitem[Chen et~al.(2019)Chen, Arti{\`e}res, and Denoyer]{chen2019unsupervised}
Micka{\"e}l Chen, Thierry Arti{\`e}res, and Ludovic Denoyer.
\newblock Unsupervised object segmentation by redrawing.
\newblock In \emph{NeurIPS}, 2019.

\bibitem[Chen et~al.(2024)Chen, Chan, Shiu, Yen, Yeh, and Liu]{chen2024narcan}
Ting-Hsuan Chen, Jie~Wen Chan, Hau-Shiang Shiu, Shih-Han Yen, Changhan Yeh, and Yu-Lun Liu.
\newblock Narcan: Natural refined canonical image with integration of diffusion prior for video editing.
\newblock In \emph{NeurIPS}, 2024.

\bibitem[Do et~al.(2021)Do, Tran, and Venkatesh]{do2021clustering}
Kien Do, Truyen Tran, and Svetha Venkatesh.
\newblock Clustering by maximizing mutual information across views.
\newblock In \emph{ICCV}, 2021.

\bibitem[Dodge~Jr et~al.(1992)Dodge~Jr, Brown, Bolson, and Dodge]{dodge1992lumen}
J~Theodore Dodge~Jr, B~Greg Brown, Edward~L Bolson, and Harold~T Dodge.
\newblock Lumen diameter of normal human coronary arteries. influence of age, sex, anatomic variation, and left ventricular hypertrophy or dilation.
\newblock \emph{Circulation}, 1992.

\bibitem[Doersch et~al.(2015)Doersch, Gupta, and Efros]{doersch2015unsupervised}
Carl Doersch, Abhinav Gupta, and Alexei~A Efros.
\newblock Unsupervised visual representation learning by context prediction.
\newblock In \emph{ICCV}, 2015.

\bibitem[Fan et~al.(2019)Fan, Mo, Qiu, Li, Zhu, Li, Hu, Rong, and Chen]{fan2019accurate}
Zhun Fan, Jiajie Mo, Benzhang Qiu, Wenji Li, Guijie Zhu, Chong Li, Jianye Hu, Yibiao Rong, and Xinjian Chen.
\newblock Accurate retinal vessel segmentation via octave convolution neural network.
\newblock \emph{arXiv preprint arXiv:1906.12193}, 2019.

\bibitem[Felfelian et~al.(2016)Felfelian, Fazlali, Karimi, Soroushmehr, Samavi, Nallamothu, and Najarian]{felfelian2016vessel}
Banafsheh Felfelian, Hamid~R Fazlali, Nader Karimi, S~Mohamad~R Soroushmehr, Shadrokh Samavi, B Nallamothu, and Kayvan Najarian.
\newblock Vessel segmentation in low contrast x-ray angiogram images.
\newblock In \emph{ICIP}, 2016.

\bibitem[Frangi et~al.(1998)Frangi, Niessen, Vincken, and Viergever]{frangi1998multiscale}
Alejandro~F Frangi, Wiro~J Niessen, Koen~L Vincken, and Max~A Viergever.
\newblock Multiscale vessel enhancement filtering.
\newblock In \emph{MICCAI}, 1998.

\bibitem[Gidaris et~al.(2018)Gidaris, Singh, and Komodakis]{gidaris2018unsupervised}
Spyros Gidaris, Praveer Singh, and Nikos Komodakis.
\newblock Unsupervised representation learning by predicting image rotations.
\newblock In \emph{ICLR}, 2018.

\bibitem[G{\"o}tberg et~al.(2017)G{\"o}tberg, Christiansen, Gudmundsdottir, Sandhall, Danielewicz, Jakobsen, Olsson, {\"O}hagen, Olsson, Omerovic, et~al.]{gotberg2017instantaneous}
Matthias G{\"o}tberg, Evald~H Christiansen, Ingibj{\"o}rg~J Gudmundsdottir, Lennart Sandhall, Mikael Danielewicz, Lars Jakobsen, Sven-Erik Olsson, Patrik {\"O}hagen, Hans Olsson, Elmir Omerovic, et~al.
\newblock Instantaneous wave-free ratio versus fractional flow reserve to guide pci.
\newblock \emph{New England Journal of Medicine}, 2017.

\bibitem[Hao et~al.(2020)Hao, Ding, Qiu, Lv, Fei, Zhu, and Qin]{hao2020sequential}
Dongdong Hao, Song Ding, Linwei Qiu, Yisong Lv, Baowei Fei, Yueqi Zhu, and Binjie Qin.
\newblock Sequential vessel segmentation via deep channel attention network.
\newblock \emph{Neural Networks}, 2020.

\bibitem[Holynski et~al.(2021)Holynski, Curless, Seitz, and Szeliski]{holynski2021animating}
Aleksander Holynski, Brian~L Curless, Steven~M Seitz, and Richard Szeliski.
\newblock Animating pictures with eulerian motion fields.
\newblock In \emph{CVPR}, 2021.

\bibitem[Isensee et~al.(2018)Isensee, Petersen, Klein, Zimmerer, Jaeger, Kohl, Wasserthal, Koehler, Norajitra, Wirkert, and Maier-Hein]{isensee2018nnunet}
Fabian Isensee, Jens Petersen, Andre Klein, David Zimmerer, Paul~F. Jaeger, Simon Kohl, Jakob Wasserthal, Gregor Koehler, Tobias Norajitra, Sebastian Wirkert, and Klaus~H. Maier-Hein.
\newblock nnu-net: Self-adapting framework for u-net-based medical image segmentation, 2018.

\bibitem[Iyer et~al.(2023)Iyer, Nallamothu, Figueroa, and Nadakuditi]{iyer2023multi}
Kritika Iyer, Brahmajee~K Nallamothu, C~Alberto Figueroa, and Raj~R Nadakuditi.
\newblock A multi-stage neural network approach for coronary 3d reconstruction from uncalibrated x-ray angiography images.
\newblock 2023.

\bibitem[Ji et~al.(2019)Ji, Henriques, and Vedaldi]{ji2019invariant}
Xu Ji, Joao~F Henriques, and Andrea Vedaldi.
\newblock Invariant information clustering for unsupervised image classification and segmentation.
\newblock In \emph{ICCV}, 2019.

\bibitem[Jim{\'e}nez-Partinen et~al.(2024)Jim{\'e}nez-Partinen, Molina-Cabello, Thurnhofer-Hemsi, Palomo, Rodr{\'\i}guez-Capit{\'a}n, Molina-Ramos, and Jim{\'e}nez-Navarro]{jimenez2024cadica}
Ariadna Jim{\'e}nez-Partinen, Miguel~A Molina-Cabello, Karl Thurnhofer-Hemsi, Esteban~J Palomo, Jorge Rodr{\'\i}guez-Capit{\'a}n, Ana~I Molina-Ramos, and Manuel Jim{\'e}nez-Navarro.
\newblock Cadica: a new dataset for coronary artery disease detection by using invasive coronary angiography.
\newblock \emph{arXiv preprint arXiv:2402.00570}, 2024.

\bibitem[Jojic and Frey(2001)]{jojic2001learning}
Nebojsa Jojic and Brendan~J Frey.
\newblock Learning flexible sprites in video layers.
\newblock In \emph{CVPR}, 2001.

\bibitem[Khan et~al.(2020)Khan, Siddique, Ahmad, Mazzara, et~al.]{khan2020hybrid}
Khan~Bahadar Khan, Muhammad~Shahbaz Siddique, Muhammad Ahmad, Manuel Mazzara, et~al.
\newblock A hybrid unsupervised approach for retinal vessel segmentation.
\newblock \emph{BioMed Research International}, 2020, 2020.

\bibitem[Khowaja et~al.(2019)Khowaja, Khuwaja, and Ismaili]{khowaja2019framework}
Sunder~Ali Khowaja, Parus Khuwaja, and Imdad~Ali Ismaili.
\newblock A framework for retinal vessel segmentation from fundus images using hybrid feature set and hierarchical classification.
\newblock \emph{Signal, Image and Video Processing}, 2019.

\bibitem[Kim et~al.(2023)Kim, Oh, and Ye]{kim2022diffusion}
Boah Kim, Yujin Oh, and Jong~Chul Ye.
\newblock Diffusion adversarial representation learning for self-supervised vessel segmentation.
\newblock In \emph{ICLR}, 2023.

\bibitem[Kingma and Ba(2015)]{kingma2014adam}
Diederik~P Kingma and Jimmy Ba.
\newblock Adam: A method for stochastic optimization.
\newblock In \emph{ICLR}, 2015.

\bibitem[Larsson et~al.(2017)Larsson, Maire, and Shakhnarovich]{larsson2017colorization}
Gustav Larsson, Michael Maire, and Gregory Shakhnarovich.
\newblock Colorization as a proxy task for visual understanding.
\newblock In \emph{CVPR}, 2017.

\bibitem[Law and Chung(2008)]{law2008three}
Max~WK Law and Albert~CS Chung.
\newblock Three dimensional curvilinear structure detection using optimally oriented flux.
\newblock In \emph{ECCV}, 2008.

\bibitem[Ledig et~al.(2017)Ledig, Theis, Husz{\'a}r, Caballero, Cunningham, Acosta, Aitken, Tejani, Totz, Wang, et~al.]{ledig2017photo}
Christian Ledig, Lucas Theis, Ferenc Husz{\'a}r, Jose Caballero, Andrew Cunningham, Alejandro Acosta, Andrew Aitken, Alykhan Tejani, Johannes Totz, Zehan Wang, et~al.
\newblock Photo-realistic single image super-resolution using a generative adversarial network.
\newblock In \emph{CVPR}, 2017.

\bibitem[Li et~al.(2021)Li, Hu, Liu, Peng, Zhou, and Peng]{li2021contrastive}
Yunfan Li, Peng Hu, Zitao Liu, Dezhong Peng, Joey~Tianyi Zhou, and Xi Peng.
\newblock Contrastive clustering.
\newblock In \emph{AAAI}, 2021.

\bibitem[Lin and Ching(2005)]{lin2005extraction}
Chih-Yang Lin and Yu-Tai Ching.
\newblock Extraction of coronary arterial tree using cine x-ray angiograms.
\newblock \emph{Biomedical Engineering: Applications, Basis and Communications}, 2005.

\bibitem[Liu et~al.(2020{\natexlab{a}})Liu, Li, Jin, Qiu, Xia, and Sun]{liu2020dynamic}
Fan Liu, Dongxiao Li, Xinyu Jin, Wenyuan Qiu, Qi Xia, and Bin Sun.
\newblock Dynamic cardiac mri reconstruction using motion aligned locally low rank tensor (mallrt).
\newblock \emph{Magnetic resonance imaging}, 2020{\natexlab{a}}.

\bibitem[Liu et~al.(2020{\natexlab{b}})Liu, Lai, Yang, Chuang, and Huang]{liu2020learning}
Yu-Lun Liu, Wei-Sheng Lai, Ming-Hsuan Yang, Yung-Yu Chuang, and Jia-Bin Huang.
\newblock Learning to see through obstructions.
\newblock In \emph{CVPR}, 2020{\natexlab{b}}.

\bibitem[Liu et~al.(2021{\natexlab{a}})Liu, Lai, Yang, Chuang, and Huang]{liu2021hybrid}
Yu-Lun Liu, Wei-Sheng Lai, Ming-Hsuan Yang, Yung-Yu Chuang, and Jia-Bin Huang.
\newblock Hybrid neural fusion for full-frame video stabilization.
\newblock In \emph{ICCV}, 2021{\natexlab{a}}.

\bibitem[Liu et~al.(2021{\natexlab{b}})Liu, Lai, Yang, Chuang, and Huang]{liu2021learning}
Yu-Lun Liu, Wei-Sheng Lai, Ming-Hsuan Yang, Yung-Yu Chuang, and Jia-Bin Huang.
\newblock Learning to see through obstructions with layered decomposition.
\newblock \emph{IEEE TPAMI}, 2021{\natexlab{b}}.

\bibitem[Liu et~al.(2023)Liu, Gao, Meuleman, Tseng, Saraf, Kim, Chuang, Kopf, and Huang]{liu2023robust}
Yu-Lun Liu, Chen Gao, Andreas Meuleman, Hung-Yu Tseng, Ayush Saraf, Changil Kim, Yung-Yu Chuang, Johannes Kopf, and Jia-Bin Huang.
\newblock Robust dynamic radiance fields.
\newblock In \emph{CVPR}, 2023.

\bibitem[Ma et~al.(2021)Ma, Hua, Deng, Song, Wang, Xue, Cao, Ma, and Guan]{ma2021self}
Yuxin Ma, Yang Hua, Hanming Deng, Tao Song, Hao Wang, Zhengui Xue, Heng Cao, Ruhui Ma, and Haibing Guan.
\newblock Self-supervised vessel segmentation via adversarial learning.
\newblock In \emph{ICCV}, 2021.

\bibitem[Maglaveras et~al.(2001)Maglaveras, Haris, Efstratiadis, Gourassas, and Louridas]{maglaveras2001artery}
N Maglaveras, K Haris, SN Efstratiadis, J Gourassas, and G Louridas.
\newblock Artery skeleton extraction using topographic and connected component labeling.
\newblock In \emph{Computers in Cardiology 2001. Vol. 28 (Cat. No. 01CH37287)}, 2001.

\bibitem[Memari et~al.(2019)Memari, Ramli, Saripan, Mashohor, and Moghbel]{memari2019retinal}
Nogol Memari, Abd~Rahman Ramli, M~Iqbal~Bin Saripan, Syamsiah Mashohor, and Mehrdad Moghbel.
\newblock Retinal blood vessel segmentation by using matched filtering and fuzzy c-means clustering with integrated level set method for diabetic retinopathy assessment.
\newblock \emph{Journal of Medical and Biological Engineering}, 2019.

\bibitem[Misra and Maaten(2020)]{misra2020self}
Ishan Misra and Laurens van~der Maaten.
\newblock Self-supervised learning of pretext-invariant representations.
\newblock In \emph{CVPR}, 2020.

\bibitem[Misra et~al.(2016)Misra, Zitnick, and Hebert]{misra2016shuffle}
Ishan Misra, C~Lawrence Zitnick, and Martial Hebert.
\newblock Shuffle and learn: unsupervised learning using temporal order verification.
\newblock In \emph{ECCV}, 2016.

\bibitem[Nam et~al.(2022)Nam, Brubaker, and Brown]{nam2022neural}
Seonghyeon Nam, Marcus~A Brubaker, and Michael~S Brown.
\newblock Neural image representations for multi-image fusion and layer separation.
\newblock In \emph{ECCV}, 2022.

\bibitem[Nasr-Esfahani et~al.(2016)Nasr-Esfahani, Samavi, Karimi, Soroushmehr, Ward, Jafari, Felfeliyan, Nallamothu, and Najarian]{nasr2016vessel}
Ebrahim Nasr-Esfahani, Shadrokh Samavi, Nader Karimi, SM~Reza Soroushmehr, Kevin Ward, Mohammad~H Jafari, Banafsheh Felfeliyan, B Nallamothu, and Kayvan Najarian.
\newblock Vessel extraction in x-ray angiograms using deep learning.
\newblock In \emph{EMBC}, 2016.

\bibitem[Noroozi and Favaro(2016)]{noroozi2016unsupervised}
Mehdi Noroozi and Paolo Favaro.
\newblock Unsupervised learning of visual representations by solving jigsaw puzzles.
\newblock In \emph{ECCV}, 2016.

\bibitem[Ost et~al.(2021)Ost, Mannan, Thuerey, Knodt, and Heide]{ost2021neural}
Julian Ost, Fahim Mannan, Nils Thuerey, Julian Knodt, and Felix Heide.
\newblock Neural scene graphs for dynamic scenes.
\newblock In \emph{CVPR}, 2021.

\bibitem[Otsu et~al.(1975)]{otsu1975threshold}
Nobuyuki Otsu et~al.
\newblock A threshold selection method from gray-level histograms.
\newblock \emph{Automatica}, 1975.

\bibitem[Park et~al.(2020)Park, Efros, Zhang, and Zhu]{park2020contrastive}
Taesung Park, Alexei~A Efros, Richard Zhang, and Jun-Yan Zhu.
\newblock Contrastive learning for unpaired image-to-image translation.
\newblock In \emph{ECCV}, 2020.

\bibitem[Paszke et~al.(2019)Paszke, Gross, Massa, Lerer, Bradbury, Chanan, Killeen, Lin, Gimelshein, Antiga, et~al.]{paszke2019pytorch}
Adam Paszke, Sam Gross, Francisco Massa, Adam Lerer, James Bradbury, Gregory Chanan, Trevor Killeen, Zeming Lin, Natalia Gimelshein, Luca Antiga, et~al.
\newblock Pytorch: An imperative style, high-performance deep learning library.
\newblock In \emph{NeurIPS}, 2019.

\bibitem[Pathak et~al.(2016)Pathak, Krahenbuhl, Donahue, Darrell, and Efros]{pathak2016context}
Deepak Pathak, Philipp Krahenbuhl, Jeff Donahue, Trevor Darrell, and Alexei~A Efros.
\newblock Context encoders: Feature learning by inpainting.
\newblock In \emph{CVPR}, 2016.

\bibitem[Reinke et~al.(2024)Reinke, Tizabi, Baumgartner, Eisenmann, Heckmann-Nötzel, Kavur, Rädsch, Sudre, Acion, Antonelli, Arbel, Bakas, Benis, Buettner, Cardoso, Cheplygina, Chen, Christodoulou, Cimini, Farahani, Ferrer, Galdran, van Ginneken, Glocker, Godau, Hashimoto, Hoffman, Huisman, Isensee, Jannin, Kahn, Kainmueller, Kainz, Karargyris, Kleesiek, Kofler, Kooi, Kopp-Schneider, Kozubek, Kreshuk, Kurc, Landman, Litjens, Madani, Maier-Hein, Martel, Meijering, Menze, Moons, Müller, Nichyporuk, Nickel, Petersen, Rafelski, Rajpoot, Reyes, Riegler, Rieke, Saez-Rodriguez, Sánchez, Shetty, Summers, Taha, Tiulpin, Tsaftaris, Van~Calster, Varoquaux, Yaniv, Jäger, and Maier-Hein]{Reinke_2024}
Annika Reinke, Minu~D. Tizabi, Michael Baumgartner, Matthias Eisenmann, Doreen Heckmann-Nötzel, A.~Emre Kavur, Tim Rädsch, Carole~H. Sudre, Laura Acion, Michela Antonelli, Tal Arbel, Spyridon Bakas, Arriel Benis, Florian Buettner, M.~Jorge Cardoso, Veronika Cheplygina, Jianxu Chen, Evangelia Christodoulou, Beth~A. Cimini, Keyvan Farahani, Luciana Ferrer, Adrian Galdran, Bram van Ginneken, Ben Glocker, Patrick Godau, Daniel~A. Hashimoto, Michael~M. Hoffman, Merel Huisman, Fabian Isensee, Pierre Jannin, Charles~E. Kahn, Dagmar Kainmueller, Bernhard Kainz, Alexandros Karargyris, Jens Kleesiek, Florian Kofler, Thijs Kooi, Annette Kopp-Schneider, Michal Kozubek, Anna Kreshuk, Tahsin Kurc, Bennett~A. Landman, Geert Litjens, Amin Madani, Klaus Maier-Hein, Anne~L. Martel, Erik Meijering, Bjoern Menze, Karel G.~M. Moons, Henning Müller, Brennan Nichyporuk, Felix Nickel, Jens Petersen, Susanne~M. Rafelski, Nasir Rajpoot, Mauricio Reyes, Michael~A. Riegler, Nicola Rieke, Julio Saez-Rodriguez, Clara~I. Sánchez,
  Shravya Shetty, Ronald~M. Summers, Abdel~A. Taha, Aleksei Tiulpin, Sotirios~A. Tsaftaris, Ben Van~Calster, Gaël Varoquaux, Ziv~R. Yaniv, Paul~F. Jäger, and Lena Maier-Hein.
\newblock Understanding metric-related pitfalls in image analysis validation.
\newblock \emph{Nature Methods}, 21\penalty0 (2):\penalty0 182–194, 2024.

\bibitem[Ren and Lee(2018)]{ren2018cross}
Zhongzheng Ren and Yong~Jae Lee.
\newblock Cross-domain self-supervised multi-task feature learning using synthetic imagery.
\newblock In \emph{CVPR}, 2018.

\bibitem[Revaud et~al.(2016)Revaud, Weinzaepfel, Harchaoui, and Schmid]{revaud2016deepmatching}
Jerome Revaud, Philippe Weinzaepfel, Zaid Harchaoui, and Cordelia Schmid.
\newblock Deepmatching: Hierarchical deformable dense matching.
\newblock \emph{IJCV}, 2016.

\bibitem[Ronneberger et~al.(2015)Ronneberger, Fischer, and Brox]{ronneberger2015unet}
Olaf Ronneberger, Philipp Fischer, and Thomas Brox.
\newblock U-net: Convolutional networks for biomedical image segmentation.
\newblock In \emph{MICCAI}, 2015.

\bibitem[Shi and Malik(1998)]{shi1998motion}
Jianbo Shi and Jitendra Malik.
\newblock Motion segmentation and tracking using normalized cuts.
\newblock In \emph{ICCV}, 1998.

\bibitem[Shi et~al.(2023)Shi, Ding, Zhang, and Yang]{shi2023freecos}
Tianyi Shi, Xiaohuan Ding, Liang Zhang, and Xin Yang.
\newblock Freecos: Self-supervised learning from fractals and unlabeled images for curvilinear object segmentation.
\newblock In \emph{ICCV}, 2023.

\bibitem[Shit et~al.(2021)Shit, Paetzold, Sekuboyina, Ezhov, Unger, Zhylka, Pluim, Bauer, and Menze]{cldice2021}
Suprosanna Shit, Johannes~C Paetzold, Anjany Sekuboyina, Ivan Ezhov, Alexander Unger, Andrey Zhylka, Josien~PW Pluim, Ulrich Bauer, and Bjoern~H Menze.
\newblock cldice-a novel topology-preserving loss function for tubular structure segmentation.
\newblock In \emph{CVPR}, 2021.

\bibitem[Soomro et~al.(2019)Soomro, Afifi, Shah, Soomro, Baloch, Zheng, Yin, and Gao]{soomro2019impact}
Toufique~Ahmed Soomro, Ahmed~J Afifi, Ahmed~Ali Shah, Shafiullah Soomro, Gulsher~Ali Baloch, Lihong Zheng, Ming Yin, and Junbin Gao.
\newblock Impact of image enhancement technique on cnn model for retinal blood vessels segmentation.
\newblock \emph{IEEE Access}, 2019.

\bibitem[Stables et~al.(2022)Stables, Mullen, Elguindy, Nicholas, Aboul-Enien, Kemp, O’Kane, Hobson, Johnson, Khan, et~al.]{stables2022routine}
Rodney~H Stables, Liam~J Mullen, Mostafa Elguindy, Zoe Nicholas, Yousra~H Aboul-Enien, Ian Kemp, Peter O’Kane, Alex Hobson, Thomas~W Johnson, Sohail~Q Khan, et~al.
\newblock Routine pressure wire assessment versus conventional angiography in the management of patients with coronary artery disease: the ripcord 2 trial.
\newblock \emph{Circulation}, 2022.

\bibitem[Su et~al.(2024)Su, Hu, Tsai, Lee, Lin, and Liu]{su2024boostmvsnerfs}
Chih-Hai Su, Chih-Yao Hu, Shr-Ruei Tsai, Jie-Ying Lee, Chin-Yang Lin, and Yu-Lun Liu.
\newblock Boostmvsnerfs: Boosting mvs-based nerfs to generalizable view synthesis in large-scale scenes.
\newblock In \emph{ACM SIGGRAPH 2024 Conference Papers}, 2024.

\bibitem[Teed and Deng(2020)]{teed2020raft}
Zachary Teed and Jia Deng.
\newblock Raft: Recurrent all-pairs field transforms for optical flow.
\newblock In \emph{ECCV}, 2020.

\bibitem[Toth et~al.(2014)Toth, Hamilos, Pyxaras, Mangiacapra, Nelis, De~Vroey, Di~Serafino, Muller, Van~Mieghem, Wyffels, et~al.]{toth2014evolving}
Gabor Toth, Michalis Hamilos, Stylianos Pyxaras, Fabio Mangiacapra, Olivier Nelis, Frederic De~Vroey, Luigi Di~Serafino, Olivier Muller, Carlos Van~Mieghem, Eric Wyffels, et~al.
\newblock Evolving concepts of angiogram: fractional flow reserve discordances in 4000 coronary stenoses.
\newblock \emph{European heart journal}, 2014.

\bibitem[Ulyanov et~al.(2018)Ulyanov, Vedaldi, and Lempitsky]{ulyanov2018deep}
Dmitry Ulyanov, Andrea Vedaldi, and Victor Lempitsky.
\newblock Deep image prior.
\newblock In \emph{CVPR}, 2018.

\bibitem[Wang and Chung(2020)]{wang2020higher}
Jierong Wang and Albert~CS Chung.
\newblock Higher-order flux with spherical harmonics transform for vascular analysis.
\newblock In \emph{MICCAI}, 2020.

\bibitem[Wang and Adelson(1994)]{wang1994representing}
John~YA Wang and Edward~H Adelson.
\newblock Representing moving images with layers.
\newblock \emph{IEEE TIP}, 1994.

\bibitem[Wang et~al.(2022)Wang, Zhao, Zhang, Ding, Wang, and Shen]{wang2022contrastmask}
Xuehui Wang, Kai Zhao, Ruixin Zhang, Shouhong Ding, Yan Wang, and Wei Shen.
\newblock Contrastmask: Contrastive learning to segment every thing.
\newblock In \emph{CVPR}, 2022.

\bibitem[Wu et~al.(2021)Wu, Qu, Lin, Zhou, Qiao, Zhang, Xie, and Ma]{wu2021contrastive}
Haiyan Wu, Yanyun Qu, Shaohui Lin, Jian Zhou, Ruizhi Qiao, Zhizhong Zhang, Yuan Xie, and Lizhuang Ma.
\newblock Contrastive learning for compact single image dehazing.
\newblock In \emph{CVPR}, 2021.

\bibitem[Xie et~al.(2021)Xie, Ding, Wang, Zhan, Xu, Sun, Li, and Luo]{xie2021detco}
Enze Xie, Jian Ding, Wenhai Wang, Xiaohang Zhan, Hang Xu, Peize Sun, Zhenguo Li, and Ping Luo.
\newblock Detco: Unsupervised contrastive learning for object detection.
\newblock In \emph{ICCV}, 2021.

\bibitem[Xu et~al.(2019)Xu, Xiao, Zhao, Shao, Xie, and Zhuang]{xu2019self}
Dejing Xu, Jun Xiao, Zhou Zhao, Jian Shao, Di Xie, and Yueting Zhuang.
\newblock Self-supervised spatiotemporal learning via video clip order prediction.
\newblock In \emph{CVPR}, 2019.

\bibitem[Yang et~al.(2018)Yang, Yang, Wang, Yang, Ai, and Wang]{yang2018automatic}
Siyuan Yang, Jian Yang, Yachen Wang, Qi Yang, Danni Ai, and Yongtian Wang.
\newblock Automatic coronary artery segmentation in x-ray angiograms by multiple convolutional neural networks.
\newblock In \emph{Proceedings of the 3rd international conference on multimedia and image processing}, 2018.

\bibitem[Yang et~al.(2019{\natexlab{a}})Yang, Kweon, Roh, Lee, Kang, Park, Kim, Yang, Hur, Kang, et~al.]{yang2019deep}
Su Yang, Jihoon Kweon, Jae-Hyung Roh, Jae-Hwan Lee, Heejun Kang, Lae-Jeong Park, Dong~Jun Kim, Hyeonkyeong Yang, Jaehee Hur, Do-Yoon Kang, et~al.
\newblock Deep learning segmentation of major vessels in x-ray coronary angiography.
\newblock \emph{Scientific reports}, 2019{\natexlab{a}}.

\bibitem[Yang et~al.(2019{\natexlab{b}})Yang, Loquercio, Scaramuzza, and Soatto]{yang2019unsupervised}
Yanchao Yang, Antonio Loquercio, Davide Scaramuzza, and Stefano Soatto.
\newblock Unsupervised moving object detection via contextual information separation.
\newblock In \emph{CVPR}, 2019{\natexlab{b}}.

\bibitem[Ye et~al.(2022)Ye, Li, Tucker, Kanazawa, and Snavely]{ye2022deformable}
Vickie Ye, Zhengqi Li, Richard Tucker, Angjoo Kanazawa, and Noah Snavely.
\newblock Deformable sprites for unsupervised video decomposition.
\newblock In \emph{CVPR}, 2022.

\bibitem[Zhong et~al.(2021)Zhong, Yuan, Wu, Yuan, Peng, and Wang]{zhong2021pixel}
Yuanyi Zhong, Bodi Yuan, Hong Wu, Zhiqiang Yuan, Jian Peng, and Yu-Xiong Wang.
\newblock Pixel contrastive-consistent semi-supervised semantic segmentation.
\newblock In \emph{ICCV}, 2021.

\end{thebibliography}
}

\clearpage
\setcounter{page}{1}
\maketitlesupplementary

\section{Additional Visualization Results}
\begin{figure*}[h!]
\centering
% \small
\setlength{\tabcolsep}{1pt}
\renewcommand{\arraystretch}{1}
\resizebox{1.0\textwidth}{!} 
{
\begin{tabular}{ccccc}
\includegraphics[width=0.2\textwidth]{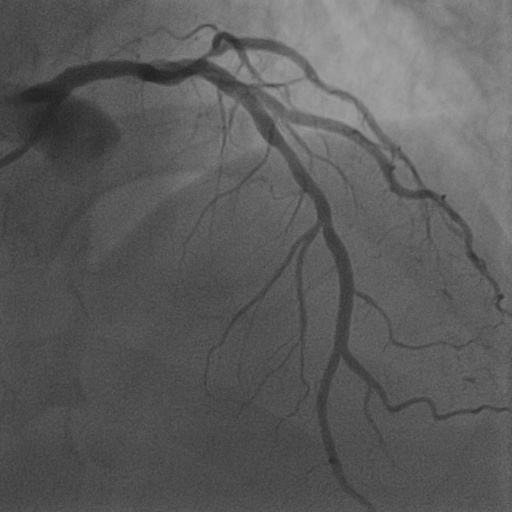} & 
\multicolumn{1}{c:}{\includegraphics[width=0.2\textwidth]{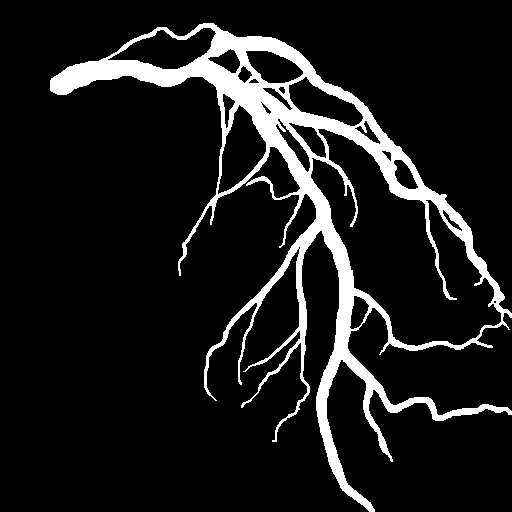}} & 
\includegraphics[width=0.2\textwidth]{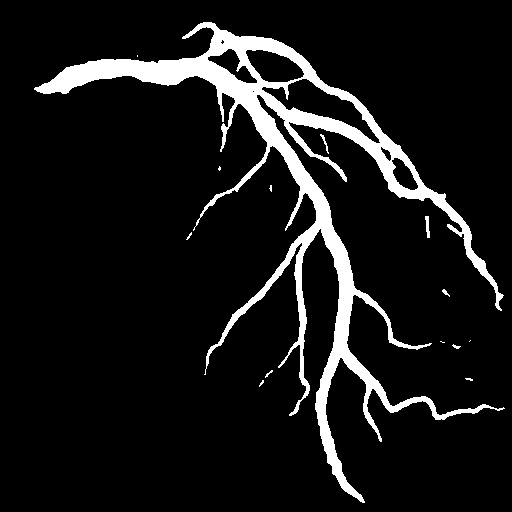} & 
\multicolumn{1}{c:}{\includegraphics[width=0.2\textwidth]{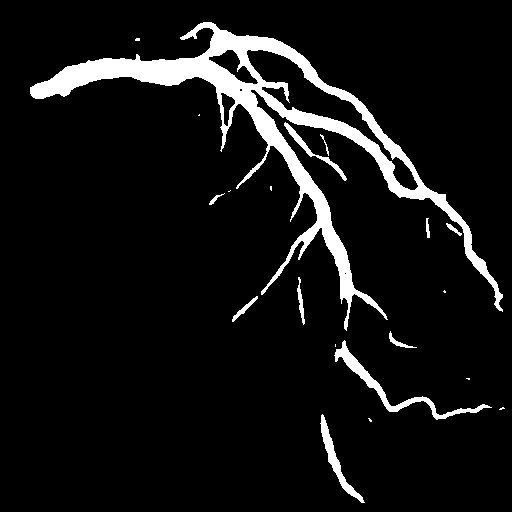}} & 
\includegraphics[width=0.2\textwidth]{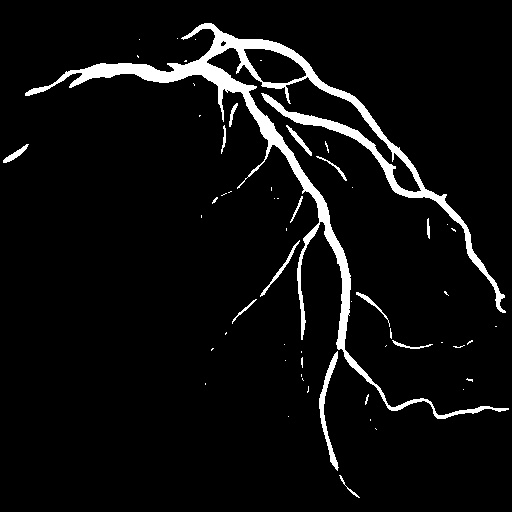} \\
Image & \multicolumn{1}{c:}{Ground truth} & U-Net (image) & \multicolumn{1}{c:}{U-Net (video)} & Hessian \\
& \multicolumn{1}{c:}{} & \multicolumn{2}{c:}{\textbf{Supervised}} \\
&
\includegraphics[width=0.2\textwidth]{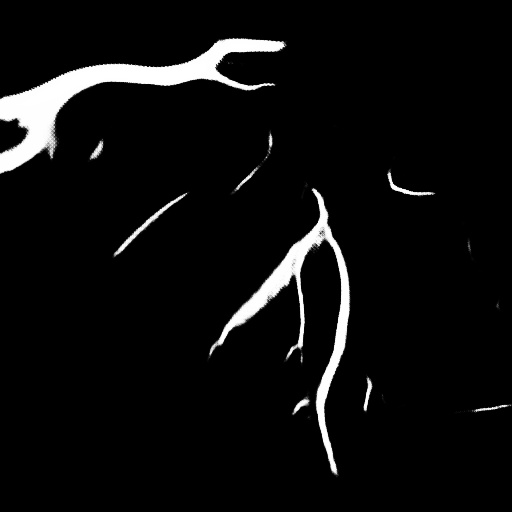} & 
\includegraphics[width=0.2\textwidth]{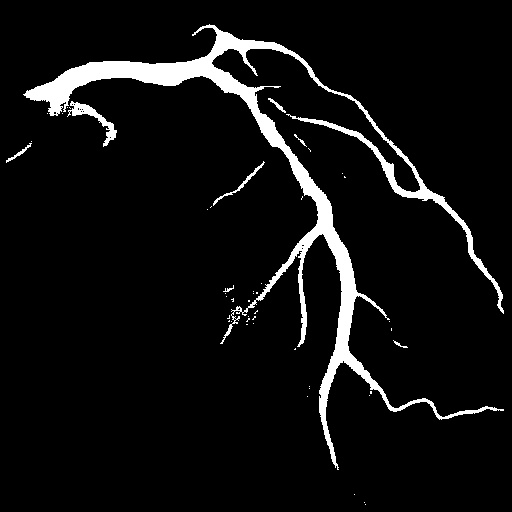} & 
\multicolumn{1}{c:}{\includegraphics[width=0.2\textwidth]{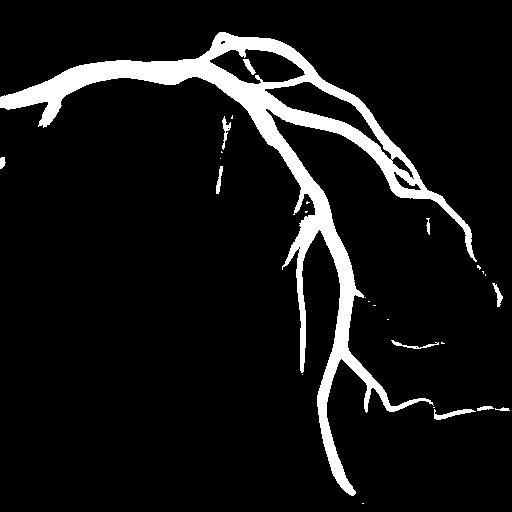}} & 
\includegraphics[width=0.2\textwidth]{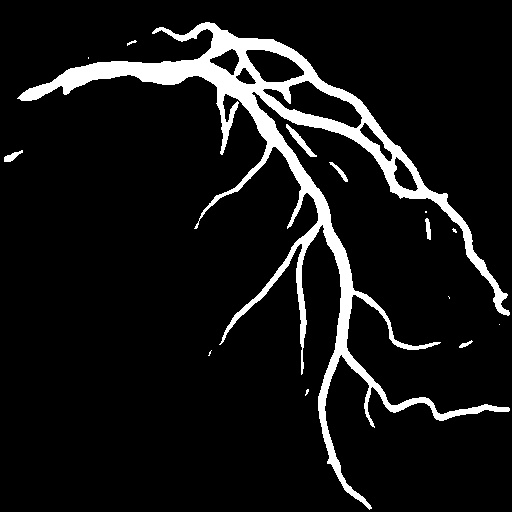} \\
& SSVS & DARL & \multicolumn{1}{c:}{FreeCOS} & \textbf{Ours} \\
& \multicolumn{3}{c:}{\textbf{Self-supervised}} & \textbf{Unsupervised} \\
\includegraphics[width=0.2\textwidth]{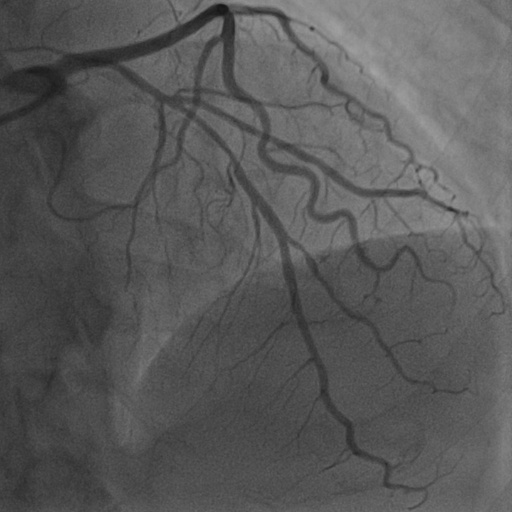} & 
\multicolumn{1}{c:}{\includegraphics[width=0.2\textwidth]{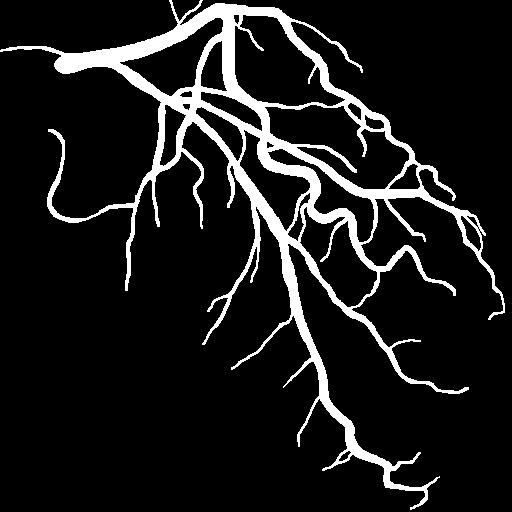}} & 
\includegraphics[width=0.2\textwidth]{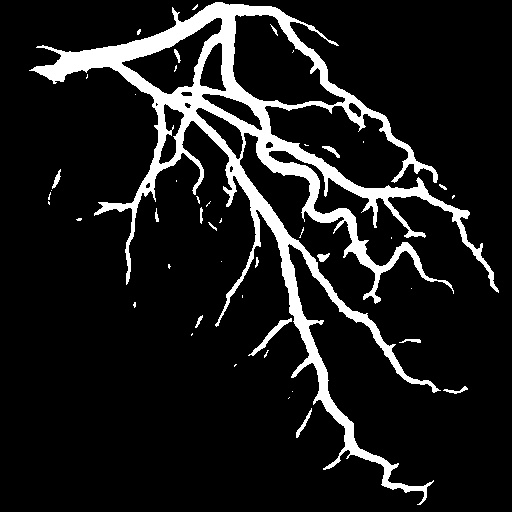} & 
\multicolumn{1}{c:}{\includegraphics[width=0.2\textwidth]{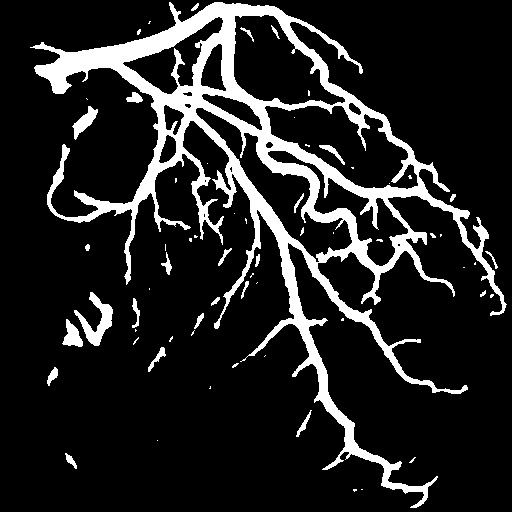}} & 
\includegraphics[width=0.2\textwidth]{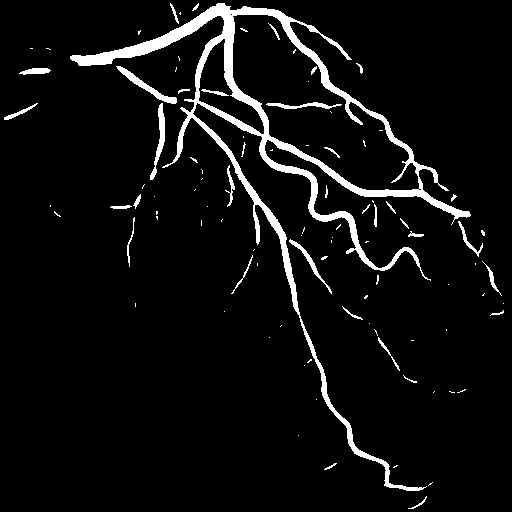} \\
Image & \multicolumn{1}{c:}{Ground truth} & U-Net (image) & \multicolumn{1}{c:}{U-Net (video)} & Hessian \\
& \multicolumn{1}{c:}{} & \multicolumn{2}{c:}{\textbf{Supervised}} \\
&
\includegraphics[width=0.2\textwidth]{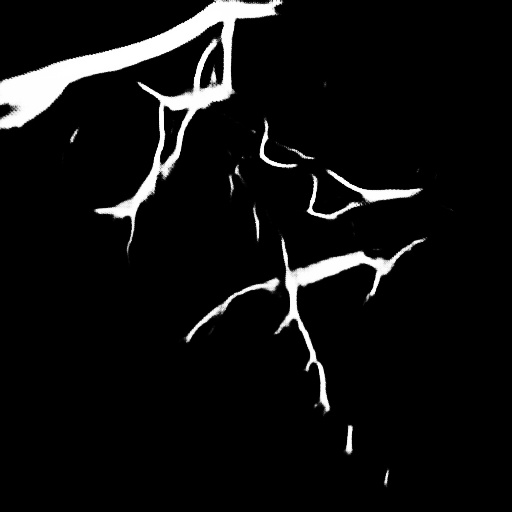} & 
\includegraphics[width=0.2\textwidth]{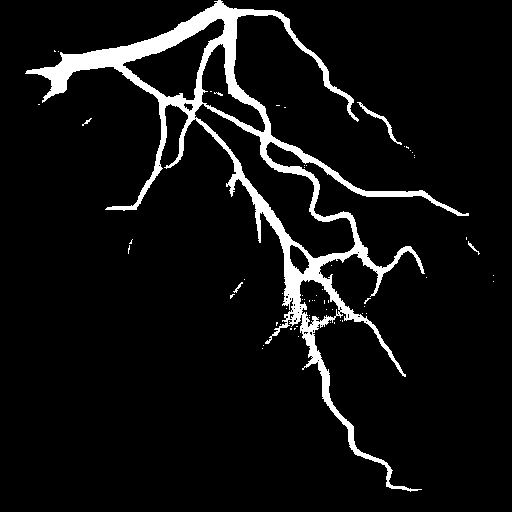} & 
\multicolumn{1}{c:}{\includegraphics[width=0.2\textwidth]{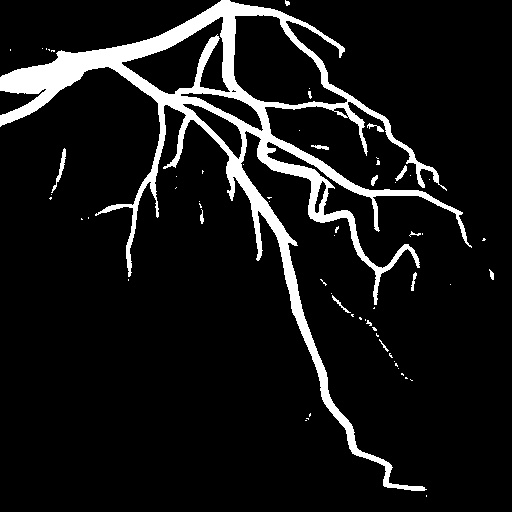}} & 
\includegraphics[width=0.2\textwidth]{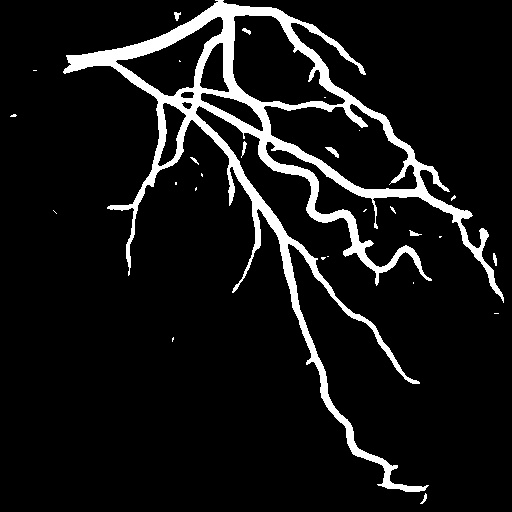} \\
& SSVS & DARL & \multicolumn{1}{c:}{FreeCOS} & \textbf{Ours} \\
& \multicolumn{3}{c:}{\textbf{Self-supervised}} & \textbf{Unsupervised} \\
% \includegraphics[width=0.2\textwidth]{figures/results/00184_Image.jpg} & 
% \multicolumn{1}{c:}{\includegraphics[width=0.2\textwidth]{figures/results/00184_GT.jpg}} & 
% \includegraphics[width=0.2\textwidth]{figures/results/00184_UNet_image.jpg} & 
% \multicolumn{1}{c:}{\includegraphics[width=0.2\textwidth]{figures/results/00184_UNet_video.jpg}} & 
% \includegraphics[width=0.2\textwidth]{figures/results/00184_Hessian.jpg} \\
% Image & \multicolumn{1}{c:}{Ground truth} & U-Net (image) & \multicolumn{1}{c:}{U-Net (video)} & Hessian \\
% & \multicolumn{1}{c:}{} & \multicolumn{2}{c:}{\textbf{Supervised}} \\
% &
% \includegraphics[width=0.2\textwidth]{figures/results/00184_SSVS.jpg} & 
% \includegraphics[width=0.2\textwidth]{figures/results/00184_DARL.jpg} & 
% \multicolumn{1}{c:}{\includegraphics[width=0.2\textwidth]{figures/results/00184_FreeCOS.jpg}} & 
% \includegraphics[width=0.2\textwidth]{figures/results/00184_Ours.jpg} \\
% & SSVS & DARL & \multicolumn{1}{c:}{FreeCOS} & \textbf{Ours} \\
% & \multicolumn{3}{c:}{\textbf{Self-supervised}} & \textbf{Unsupervised} \\
\end{tabular}%
}
% \vspace{-5mm}
\caption{\textbf{Additional visualization results on the vessel segmentation.}}
% \vspace{-1mm}
\label{fig:qualitative_appendix}
\end{figure*}

\begin{figure*}[h!]
\centering
% \small
\setlength{\tabcolsep}{1pt}
\renewcommand{\arraystretch}{1}
\resizebox{1.0\textwidth}{!} 
{
\begin{tabular}{ccccc}
\includegraphics[width=0.2\textwidth]{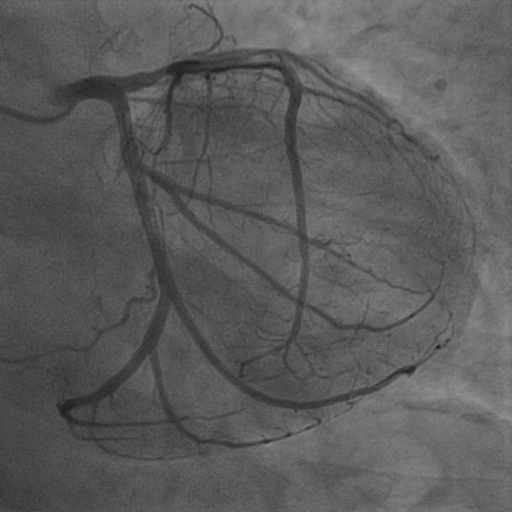} & 
\multicolumn{1}{c:}{\includegraphics[width=0.2\textwidth]{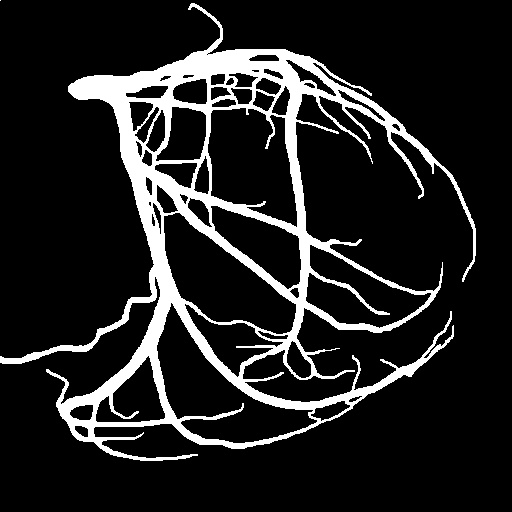}} & 
\includegraphics[width=0.2\textwidth]{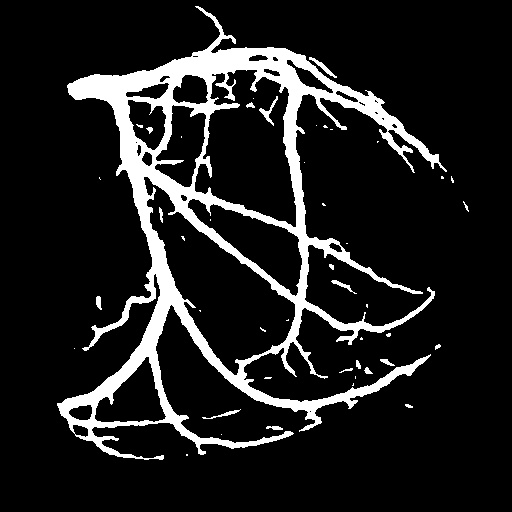} & 
\multicolumn{1}{c:}{\includegraphics[width=0.2\textwidth]{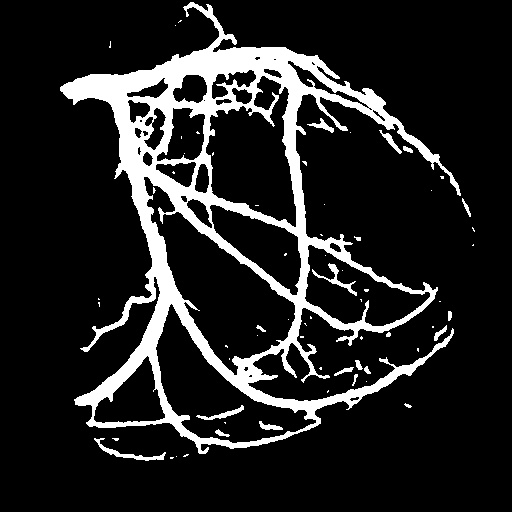}} & 
\includegraphics[width=0.2\textwidth]{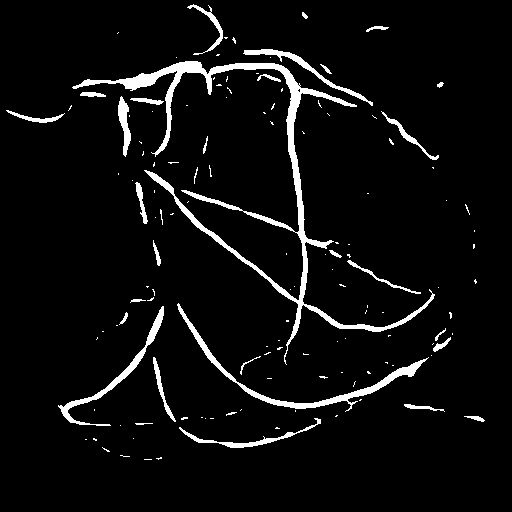} \\
Image & \multicolumn{1}{c:}{Ground truth} & U-Net (image) & \multicolumn{1}{c:}{U-Net (video)} & Hessian \\
& \multicolumn{1}{c:}{} & \multicolumn{2}{c:}{\textbf{Supervised}} \\
&
\includegraphics[width=0.2\textwidth]{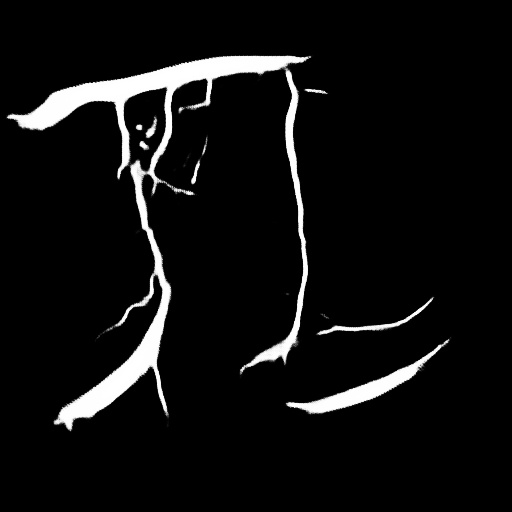} & 
\includegraphics[width=0.2\textwidth]{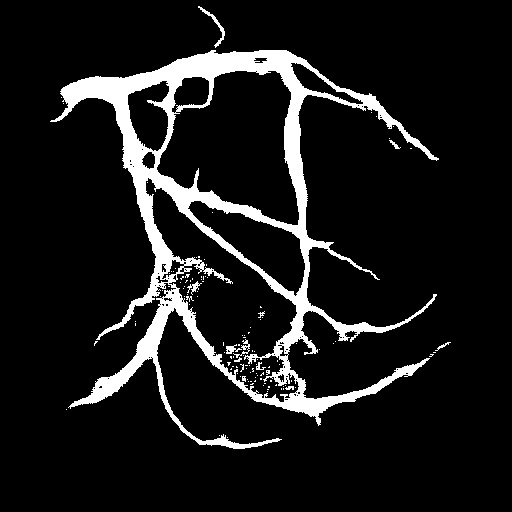} & 
\multicolumn{1}{c:}{\includegraphics[width=0.2\textwidth]{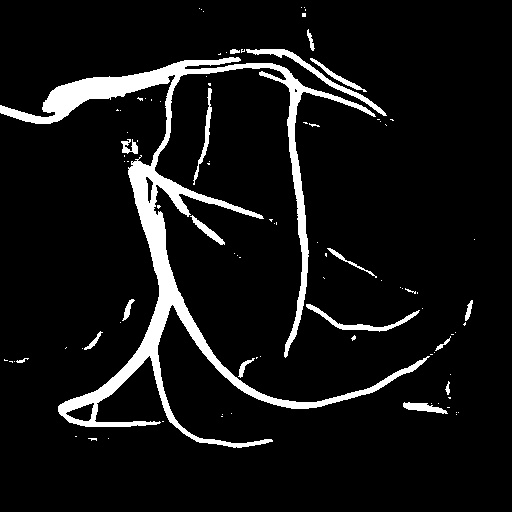}} & 
\includegraphics[width=0.2\textwidth]{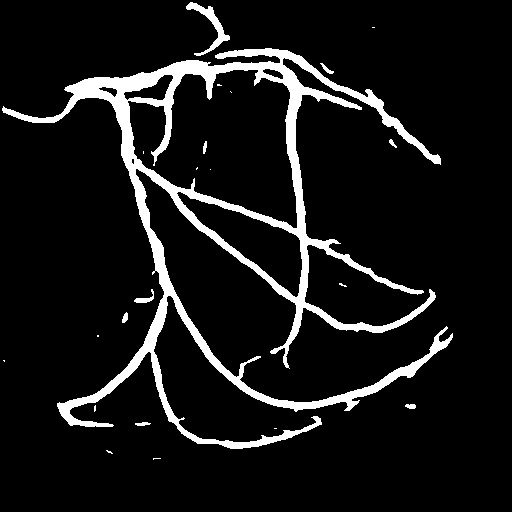} \\
& SSVS & DARL & \multicolumn{1}{c:}{FreeCOS} & \textbf{Ours} \\
& \multicolumn{3}{c:}{\textbf{Self-supervised}} & \textbf{Unsupervised} \\
\includegraphics[width=0.2\textwidth]{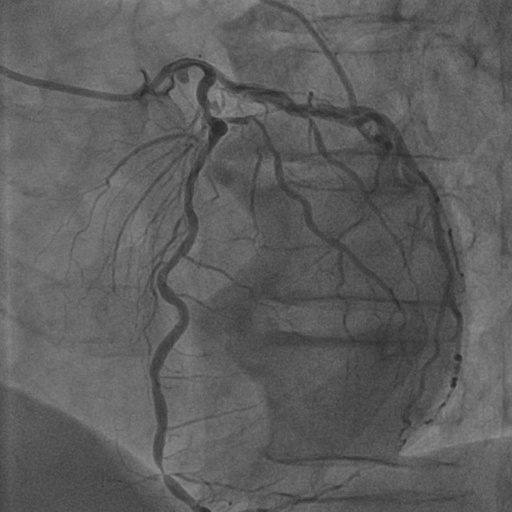} & 
\multicolumn{1}{c:}{\includegraphics[width=0.2\textwidth]{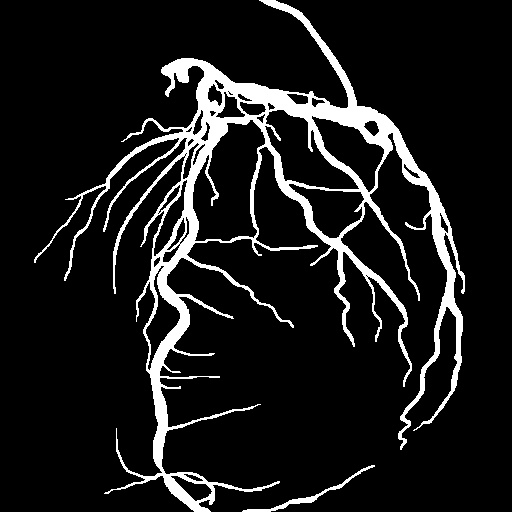}} & 
\includegraphics[width=0.2\textwidth]{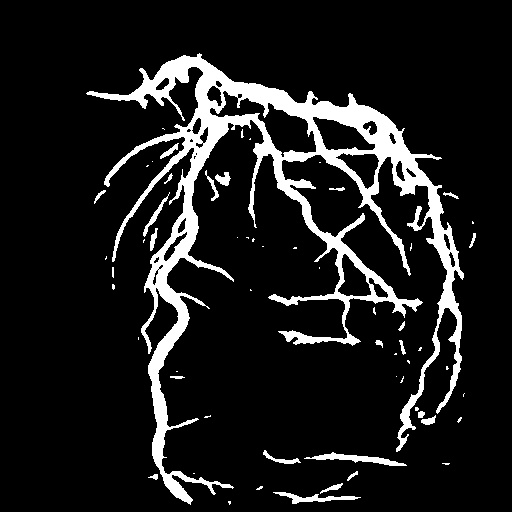} & 
\multicolumn{1}{c:}{\includegraphics[width=0.2\textwidth]{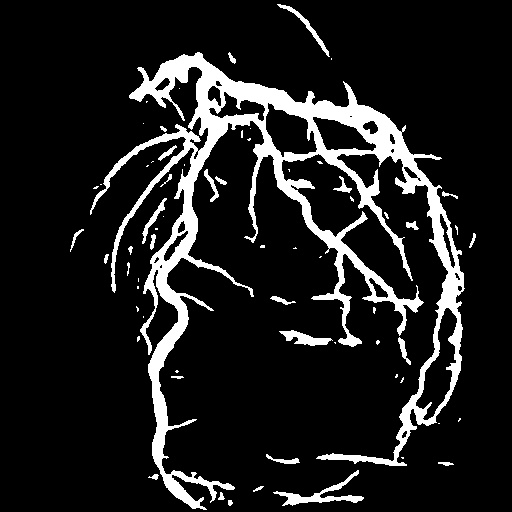}} & 
\includegraphics[width=0.2\textwidth]{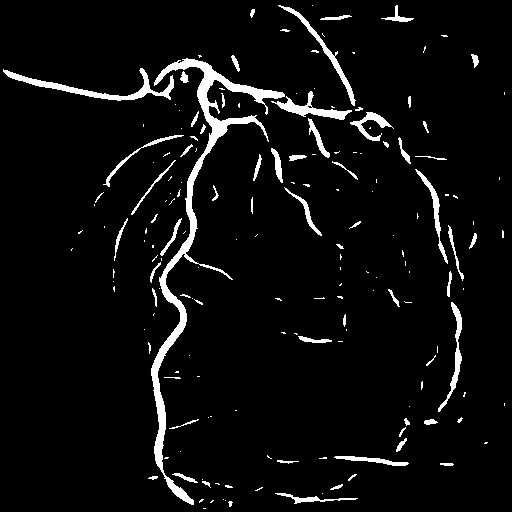} \\
Image & \multicolumn{1}{c:}{Ground truth} & U-Net (image) & \multicolumn{1}{c:}{U-Net (video)} & Hessian \\
& \multicolumn{1}{c:}{} & \multicolumn{2}{c:}{\textbf{Supervised}} \\
&
\includegraphics[width=0.2\textwidth]{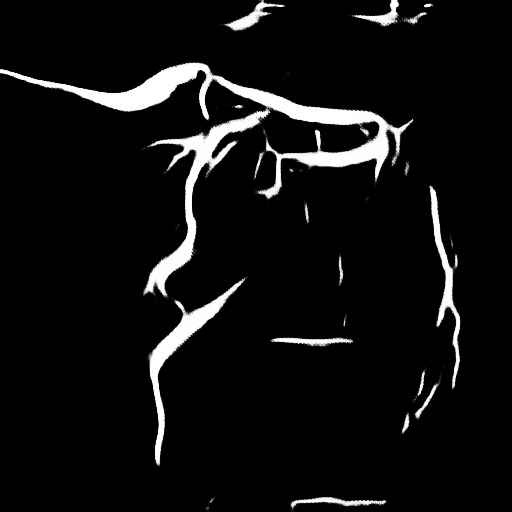} & 
\includegraphics[width=0.2\textwidth]{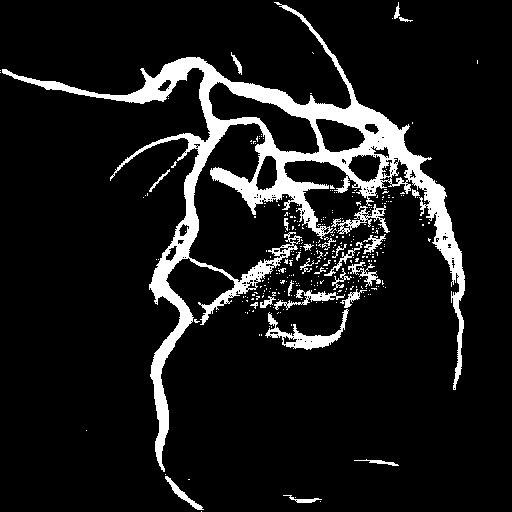} & 
\multicolumn{1}{c:}{\includegraphics[width=0.2\textwidth]{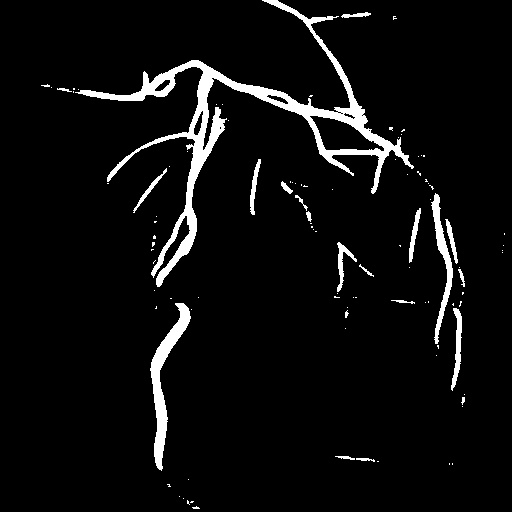}} & 
\includegraphics[width=0.2\textwidth]{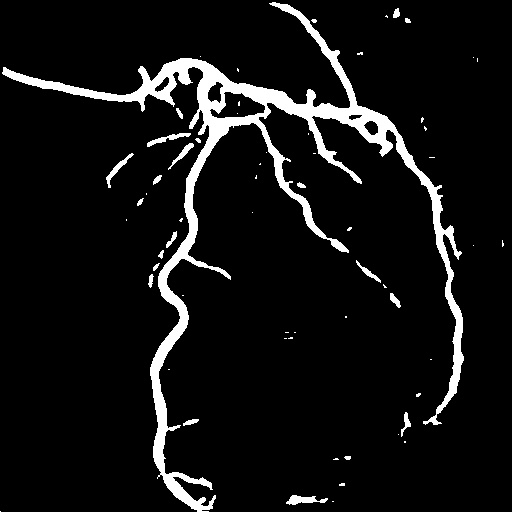} \\
& SSVS & DARL & \multicolumn{1}{c:}{FreeCOS} & \textbf{Ours} \\
& \multicolumn{3}{c:}{\textbf{Self-supervised}} & \textbf{Unsupervised} \\
\end{tabular}%
}
% \vspace{-5mm}
\caption{\textbf{Additional visualization results on the vessel segmentation.}}
% \vspace{-1mm}
\label{fig:qualitative_appendix_2}
\end{figure*}

\begin{figure*}[h!]
\centering
% \small
\setlength{\tabcolsep}{1pt}
\renewcommand{\arraystretch}{1}
\resizebox{1.0\textwidth}{!} 
{
\begin{tabular}{ccccc}
\includegraphics[width=0.2\textwidth]{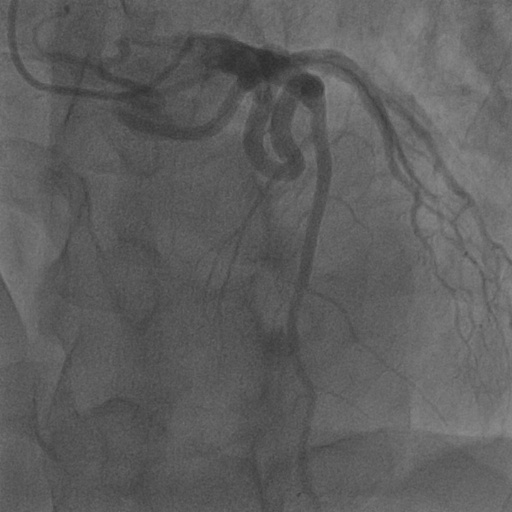} & 
\multicolumn{1}{c:}{\includegraphics[width=0.2\textwidth]{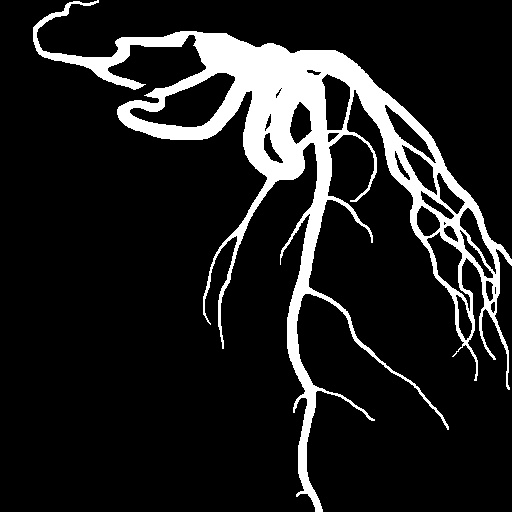}} & 
\includegraphics[width=0.2\textwidth]{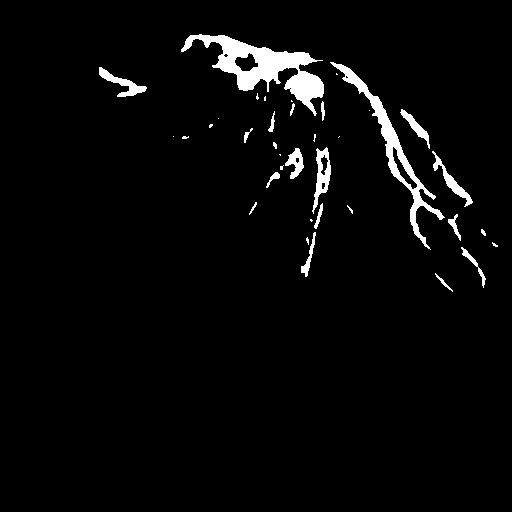} & 
\multicolumn{1}{c:}{\includegraphics[width=0.2\textwidth]{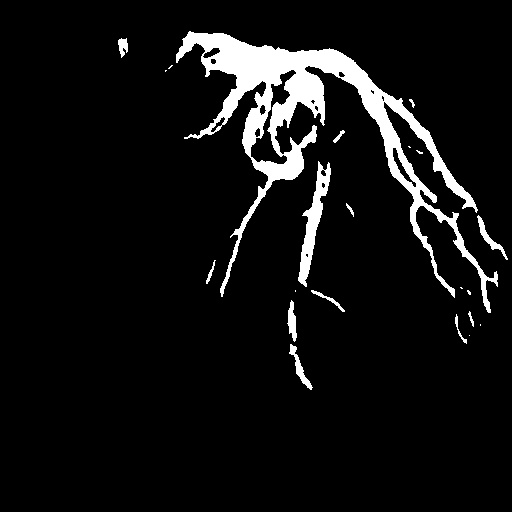}} & 
\includegraphics[width=0.2\textwidth]{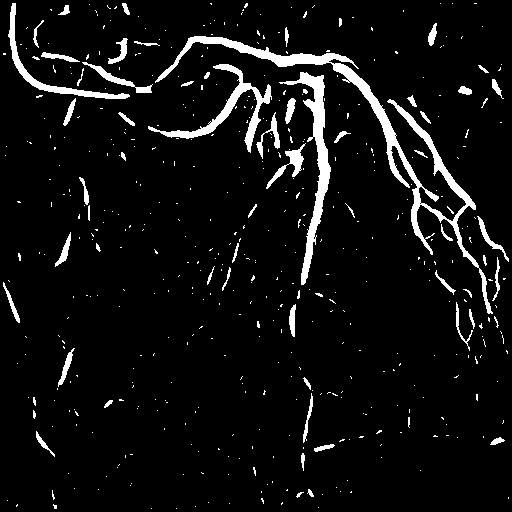} \\
Image & \multicolumn{1}{c:}{Ground truth} & U-Net (image) & \multicolumn{1}{c:}{U-Net (video)} & Hessian \\
& \multicolumn{1}{c:}{} & \multicolumn{2}{c:}{\textbf{Supervised}} \\
&
\includegraphics[width=0.2\textwidth]{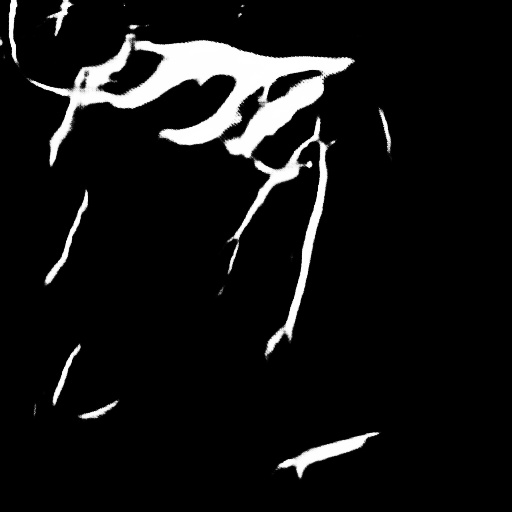} & 
\includegraphics[width=0.2\textwidth]{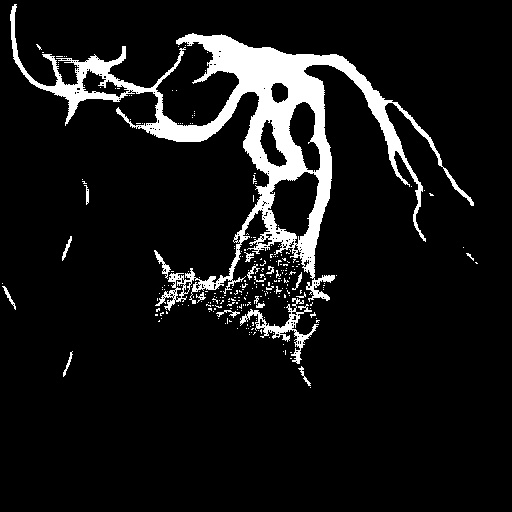} & 
\multicolumn{1}{c:}{\includegraphics[width=0.2\textwidth]{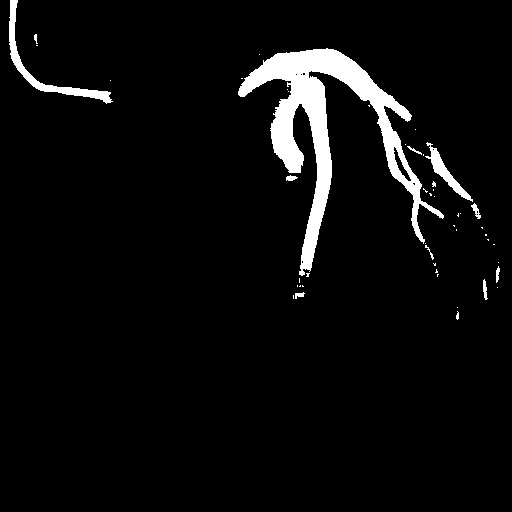}} & 
\includegraphics[width=0.2\textwidth]{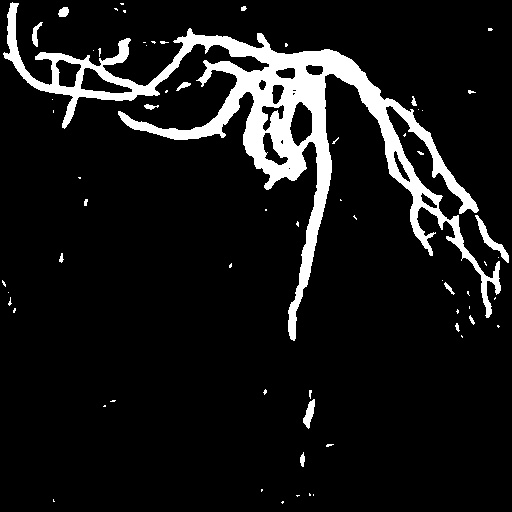} \\
& SSVS & DARL & \multicolumn{1}{c:}{FreeCOS} & \textbf{Ours} \\
& \multicolumn{3}{c:}{\textbf{Self-supervised}} & \textbf{Unsupervised} \\
\includegraphics[width=0.2\textwidth]{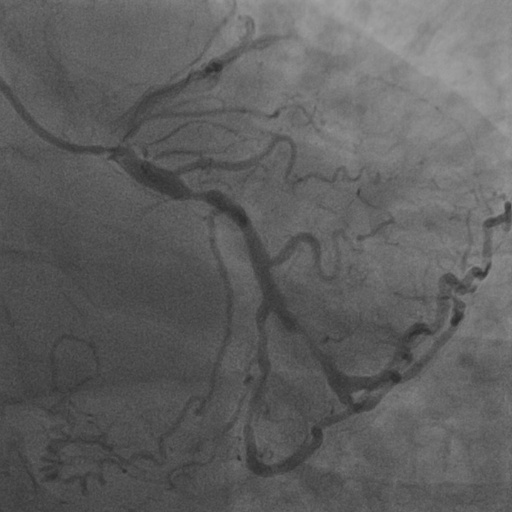} & 
\multicolumn{1}{c:}{\includegraphics[width=0.2\textwidth]{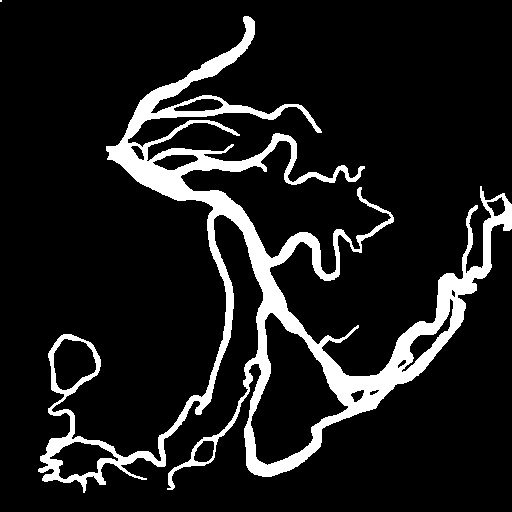}} & 
\includegraphics[width=0.2\textwidth]{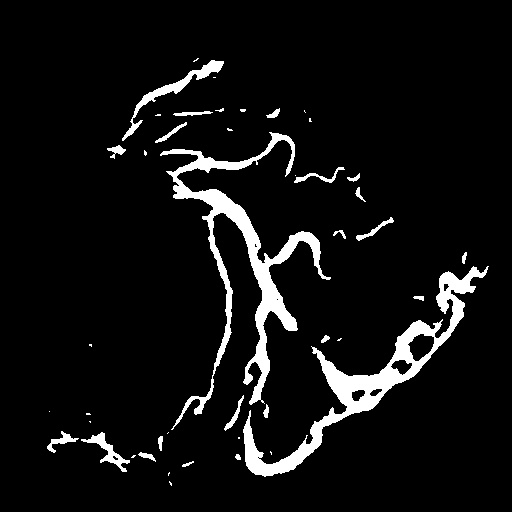} & 
\multicolumn{1}{c:}{\includegraphics[width=0.2\textwidth]{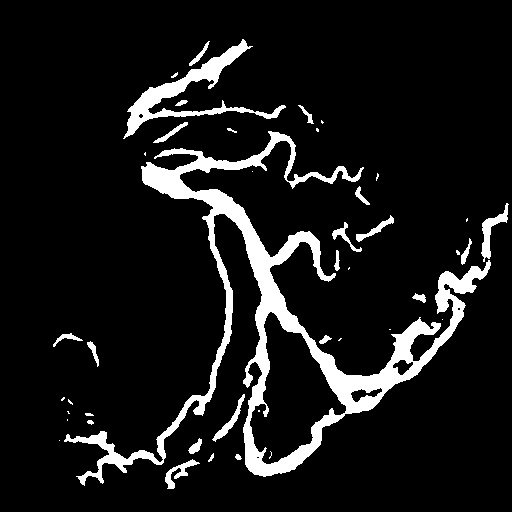}} & 
\includegraphics[width=0.2\textwidth]{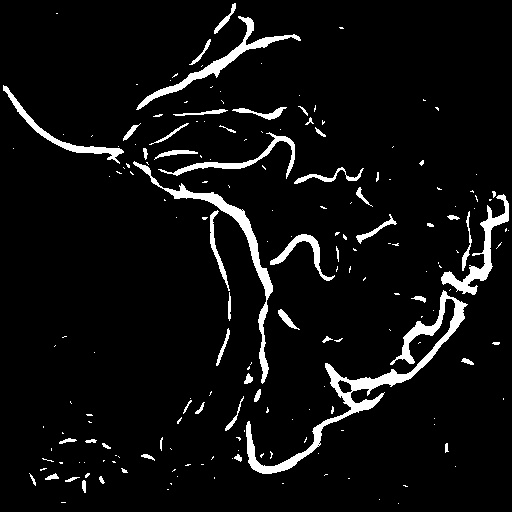} \\
Image & \multicolumn{1}{c:}{Ground truth} & U-Net (image) & \multicolumn{1}{c:}{U-Net (video)} & Hessian \\
& \multicolumn{1}{c:}{} & \multicolumn{2}{c:}{\textbf{Supervised}} \\
&
\includegraphics[width=0.2\textwidth]{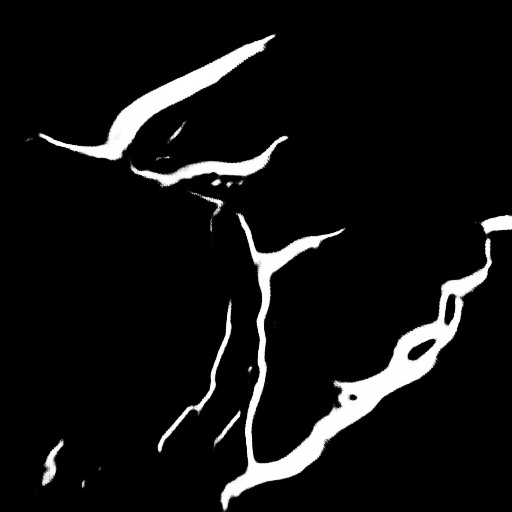} & 
\includegraphics[width=0.2\textwidth]{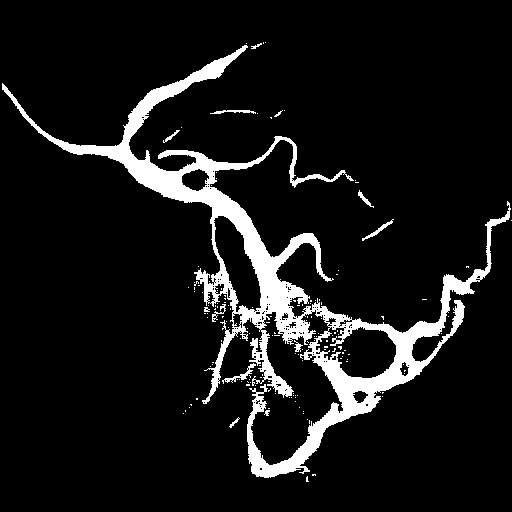} & 
\multicolumn{1}{c:}{\includegraphics[width=0.2\textwidth]{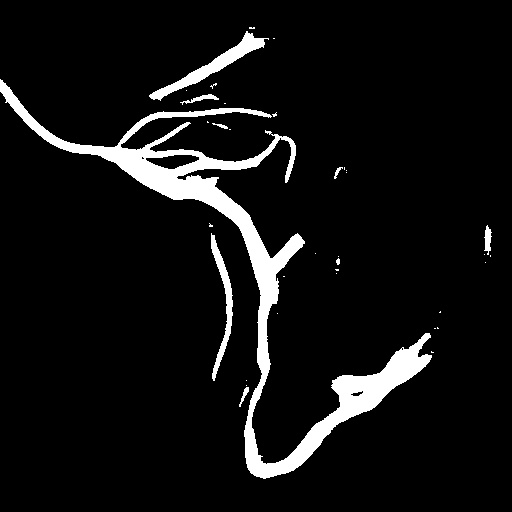}} & 
\includegraphics[width=0.2\textwidth]{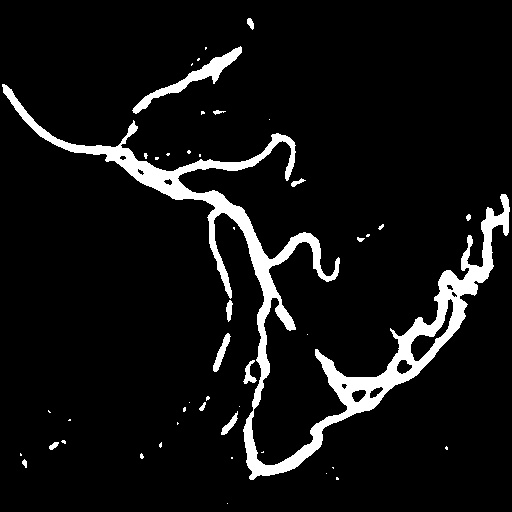} \\
& SSVS & DARL & \multicolumn{1}{c:}{FreeCOS} & \textbf{Ours} \\
& \multicolumn{3}{c:}{\textbf{Self-supervised}} & \textbf{Unsupervised} \\
\end{tabular}%
}
% \vspace{-5mm}
\caption{\textbf{Additional visualization results on the vessel segmentation.}}
% \vspace{-1mm}
\label{fig:qualitative_appendix_3}
\end{figure*}

\cref{fig:qualitative_appendix,fig:qualitative_appendix_2,fig:qualitative_appendix_3} demonstrate a comprehensive comparison where we consider supervised learning~\citep{isensee2018nnunet} as the upper bound for the vessel segmentation task, as well as all baseline methods mentioned in the main paper. For the supervised learning approach, both image-based and video-based inputs were considered. The image-based input utilized only the annotated image, while the video-based input involved using the annotated image along with two preceding and two subsequent frames, totaling five frames, as input.
The results show that although supervised learning theoretically offers the best performance, our method achieves close to those of supervised learning methods without ground truth. Additionally, we found that using five consecutive images as input for nn-UNet~\citep{isensee2018nnunet} was only slightly better than using a single image as input. In contrast, our method exhibits significant improvement compared to both the traditional Hessian-based filter and self-supervised methods, demonstrating that the robust performance of our approach is not solely attributed to the increase in input images.
We showcase some examples at the following anonymous URL: \url{https://colab.research.google.com/drive/1IYGiJECwAaoLPq7KGHQE_dvtrdHz9fUA?authuser=2&hl=zh-tw#scrollTo=n1ppvOhqbRkV}.
Additionally, we include 5 out of 111 sequences of our XACV dataset and the source code in the zip file.

\section{Temporal Coherency}
\begin{figure*}[h!]
\centering
\small
\setlength{\tabcolsep}{1pt}
\renewcommand{\arraystretch}{1}
\resizebox{1.0\textwidth}{!} 
{
\begin{tabular}{cc:cc:c:ccc:c}
\includegraphics[width=0.125\textwidth]{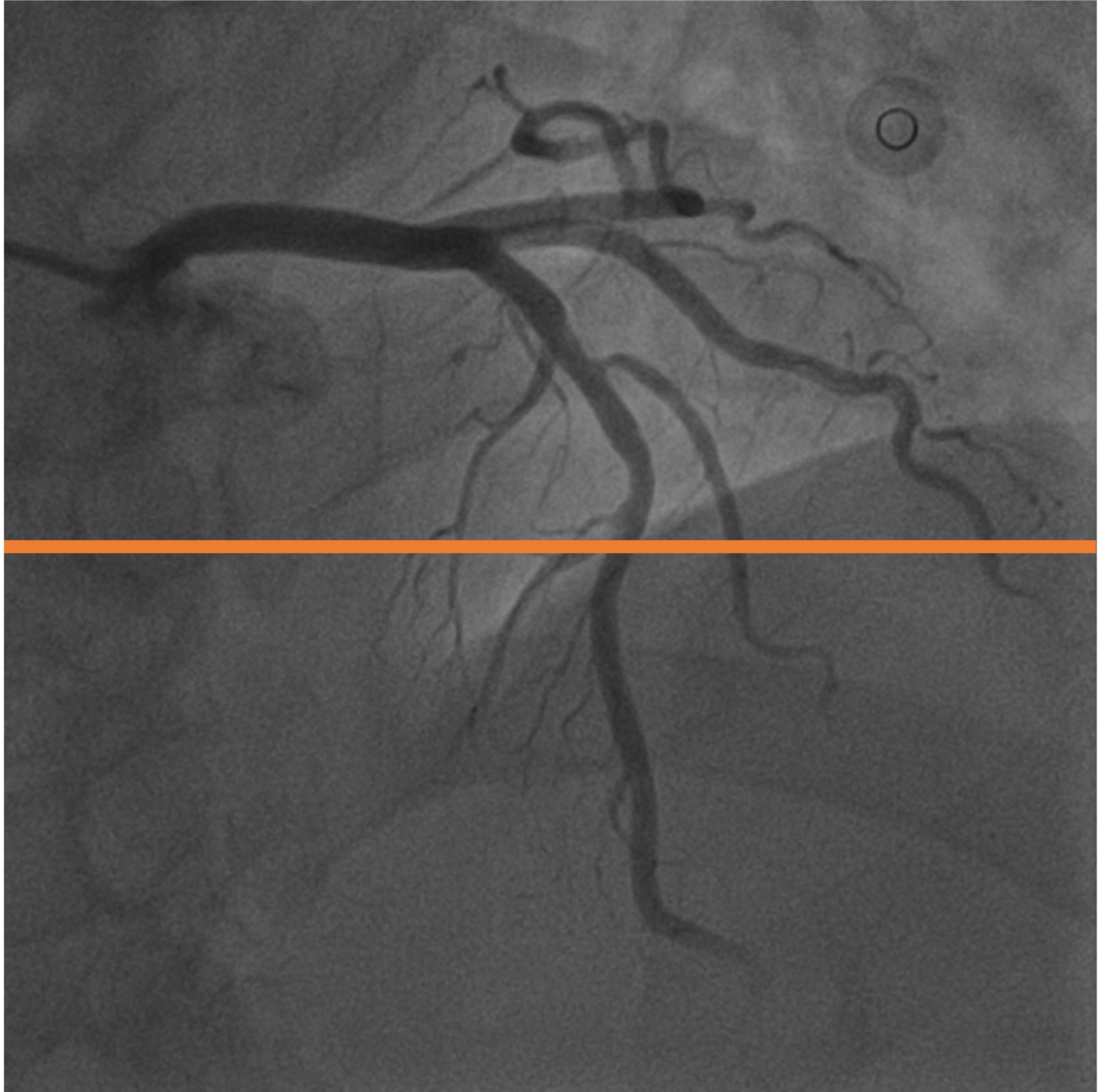} &
\includegraphics[width=0.125\textwidth]{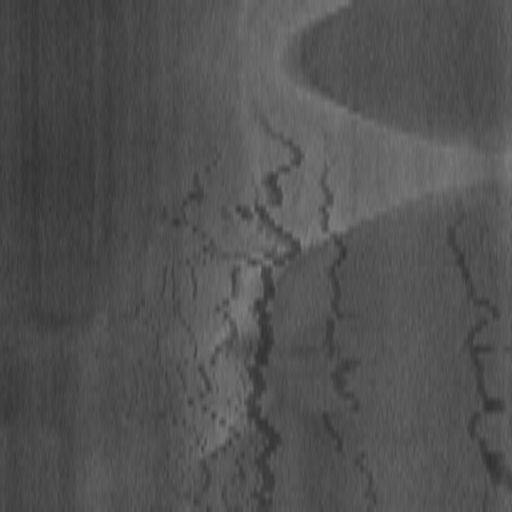} &
\includegraphics[width=0.125\textwidth]{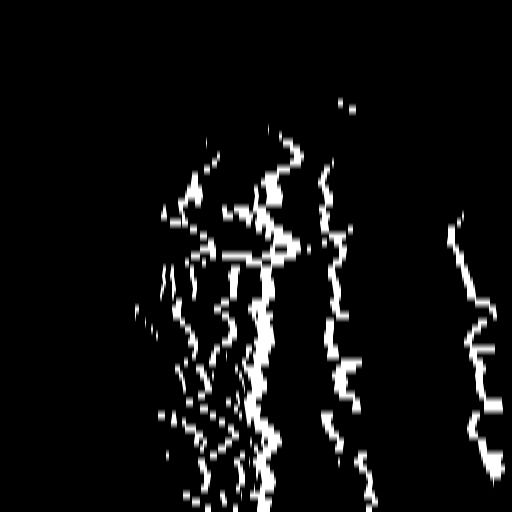} &
\includegraphics[width=0.125\textwidth]{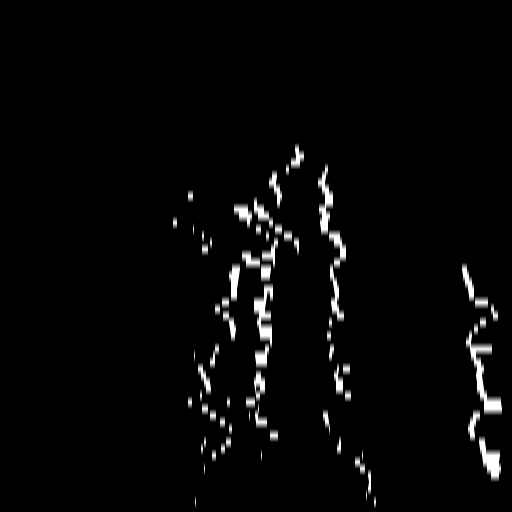} &
\includegraphics[width=0.125\textwidth]{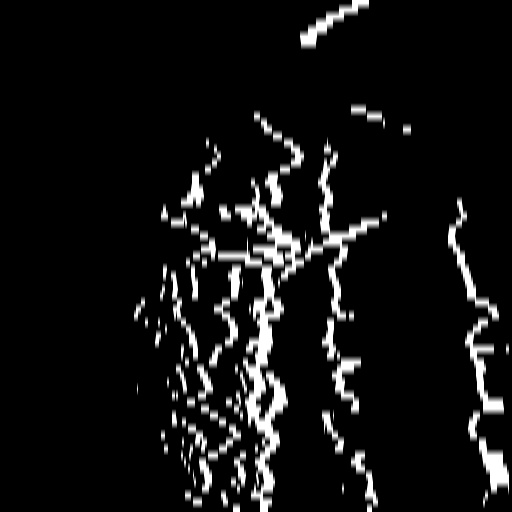} &
\includegraphics[width=0.125\textwidth]{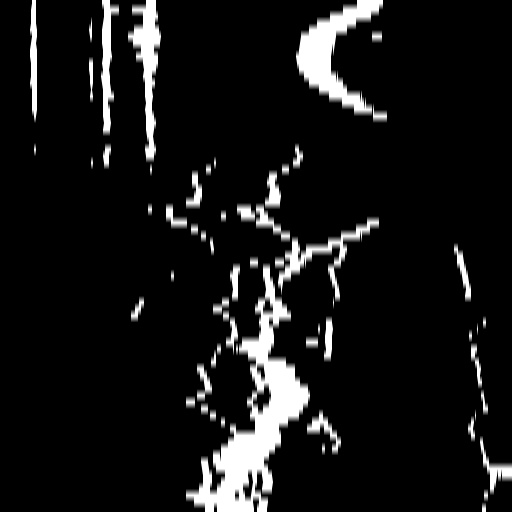} &
\includegraphics[width=0.125\textwidth]{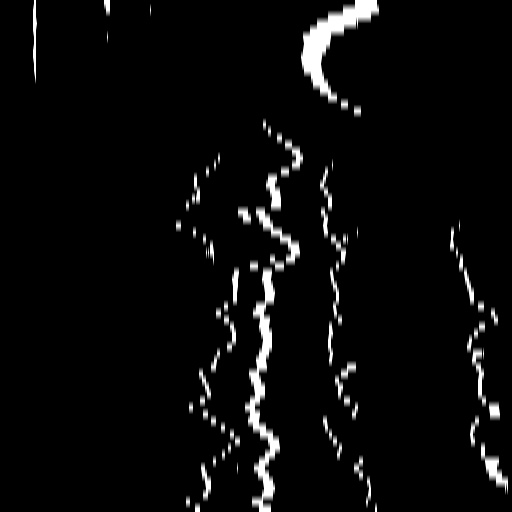} &
\includegraphics[width=0.125\textwidth]{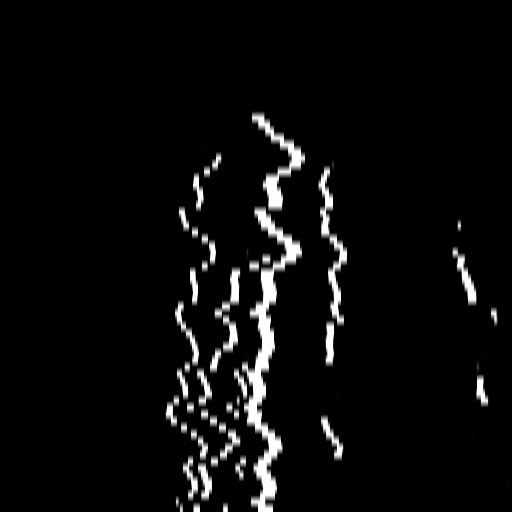} & 
\includegraphics[width=0.125\textwidth]{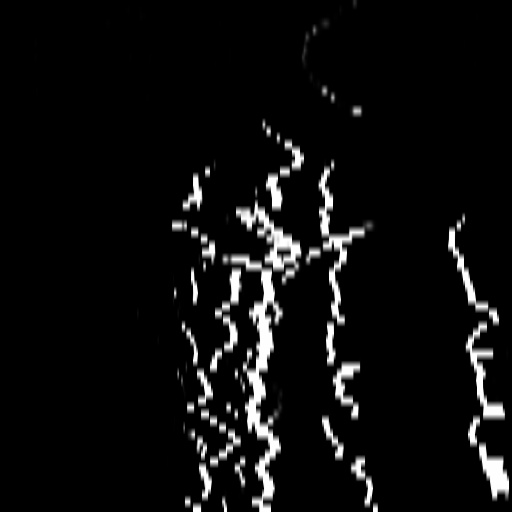}\\
\includegraphics[width=0.125\textwidth]{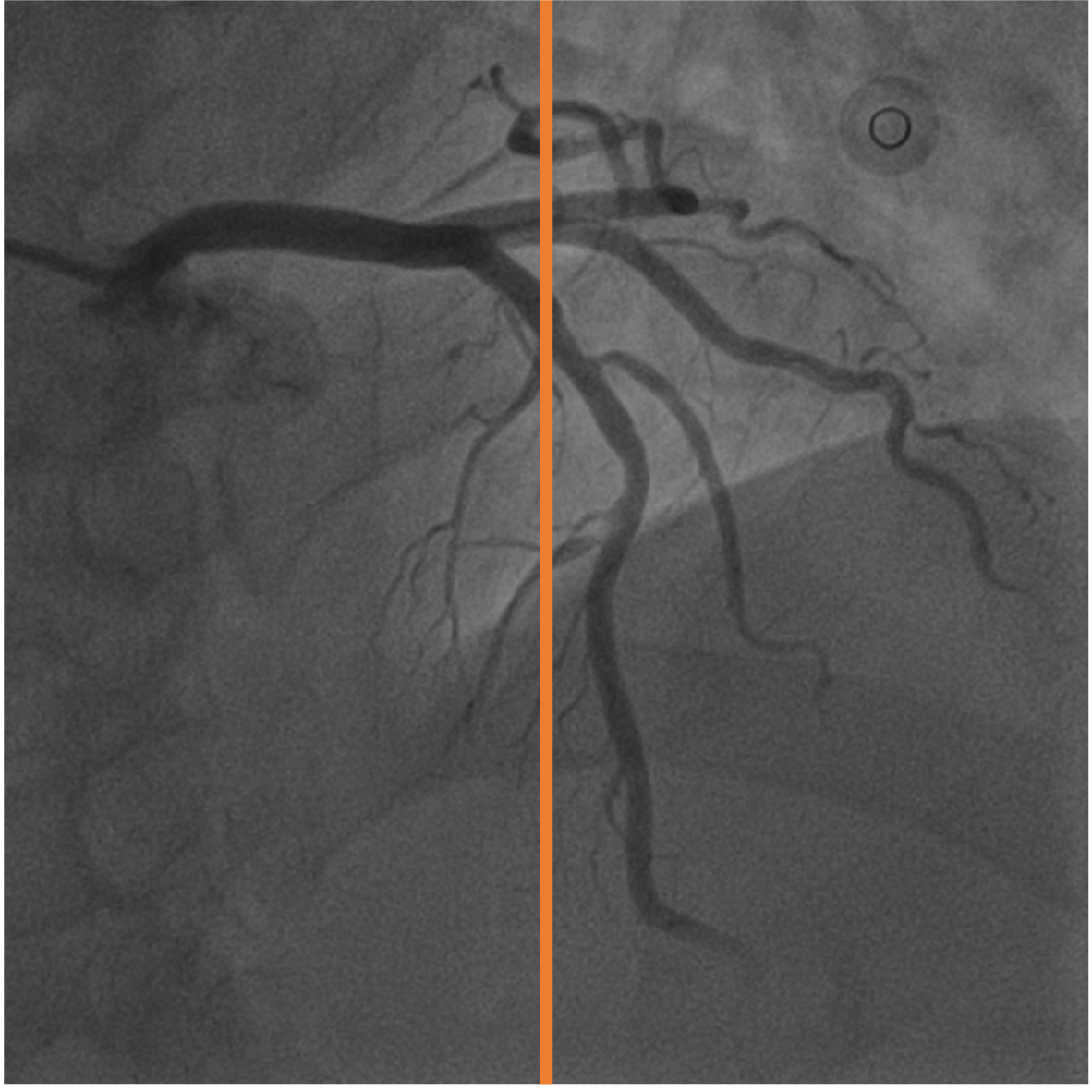} &
\includegraphics[width=0.125\textwidth]{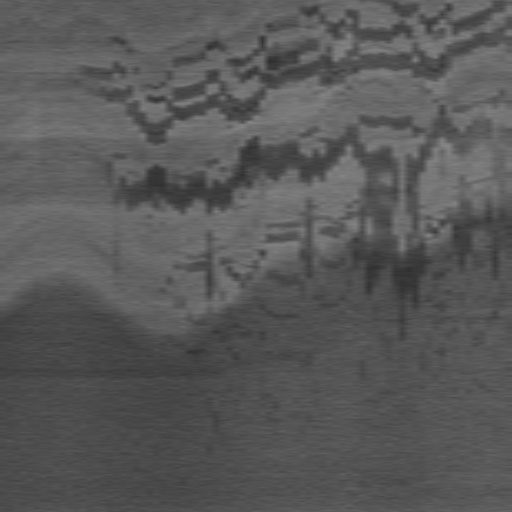} &
\includegraphics[width=0.125\textwidth]{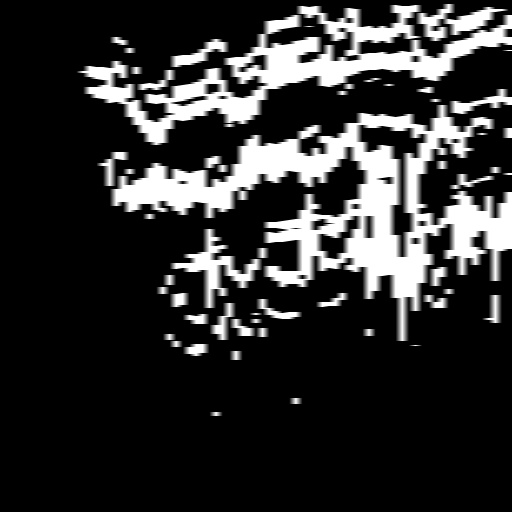} &
\includegraphics[width=0.125\textwidth]{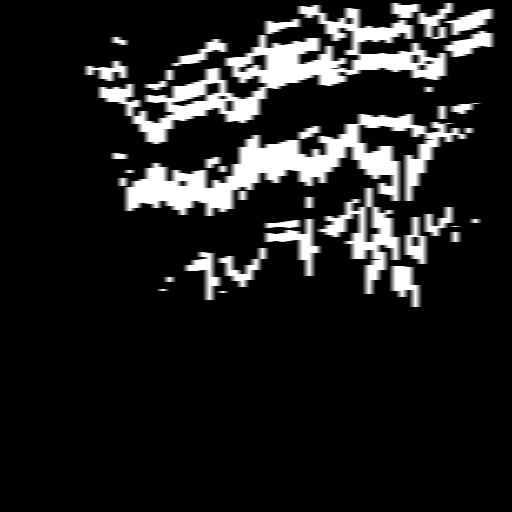} &
\includegraphics[width=0.125\textwidth]{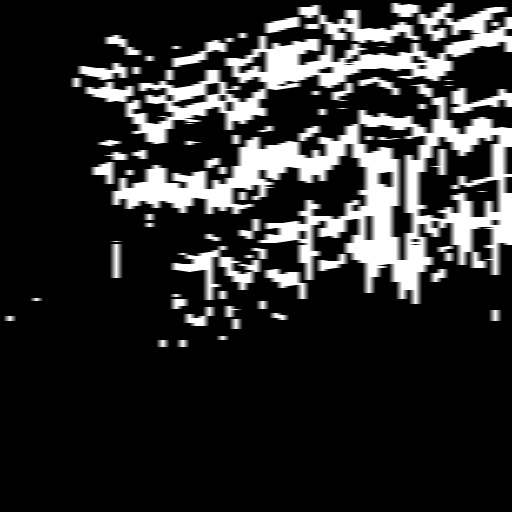} &
\includegraphics[width=0.125\textwidth]{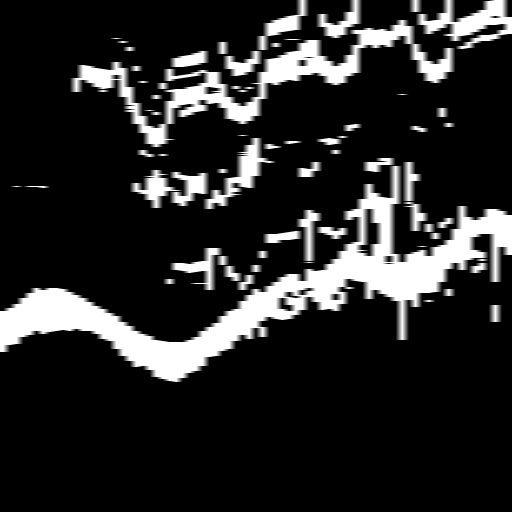} &
\includegraphics[width=0.125\textwidth]{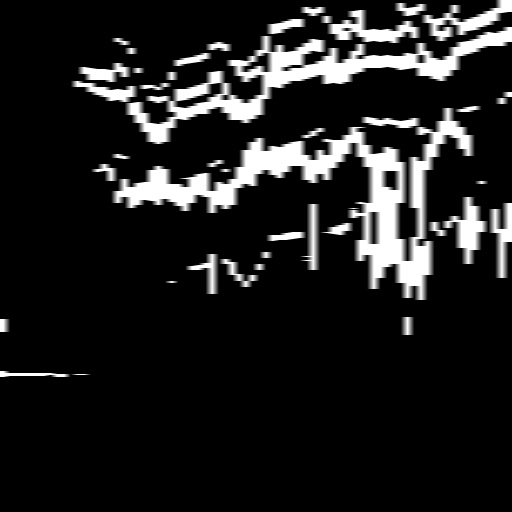} &
\includegraphics[width=0.125\textwidth]{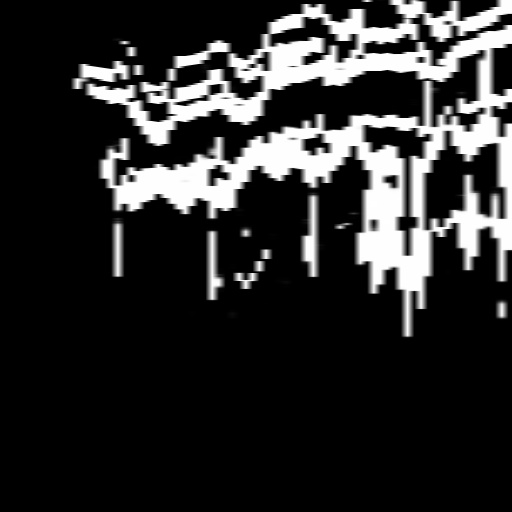} & 
\includegraphics[width=0.125\textwidth]{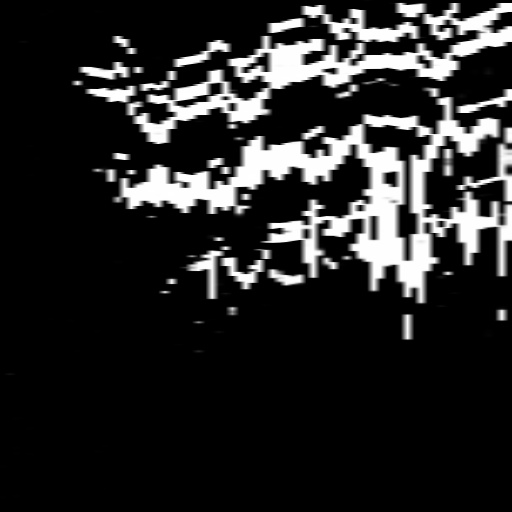}\\
\includegraphics[width=0.125\textwidth]{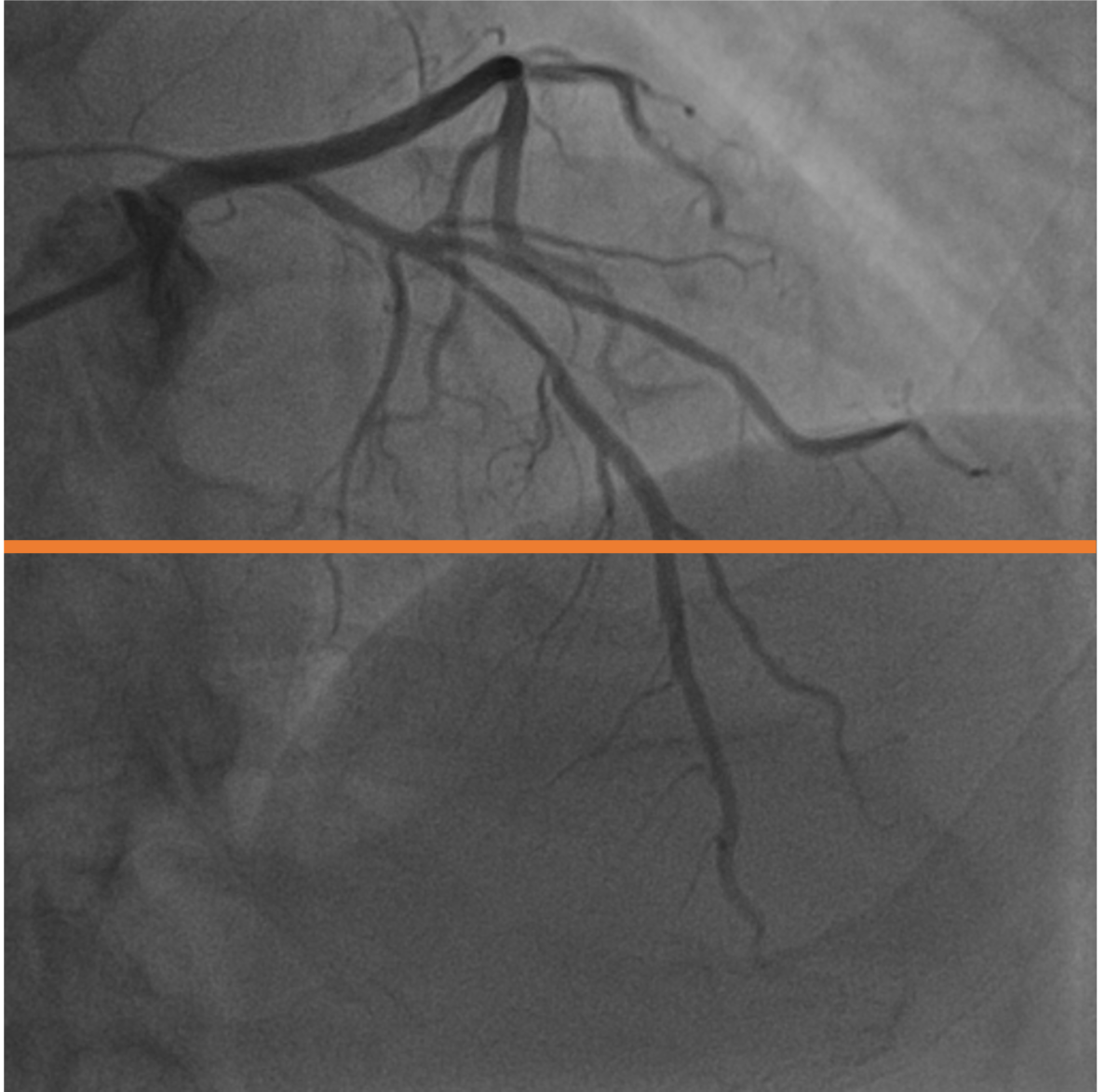} &
\includegraphics[width=0.125\textwidth]{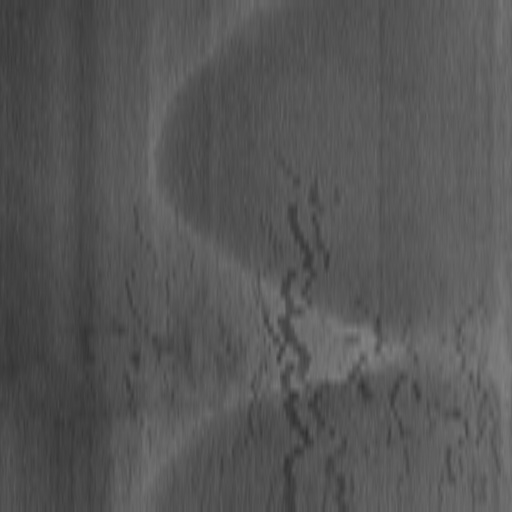} &
\includegraphics[width=0.125\textwidth]{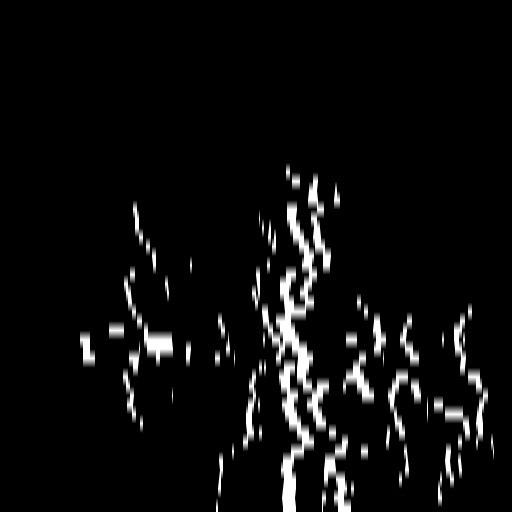} &
\includegraphics[width=0.125\textwidth]{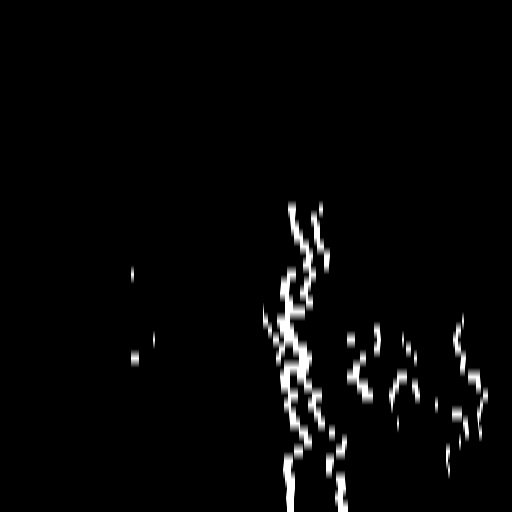} &
\includegraphics[width=0.125\textwidth]{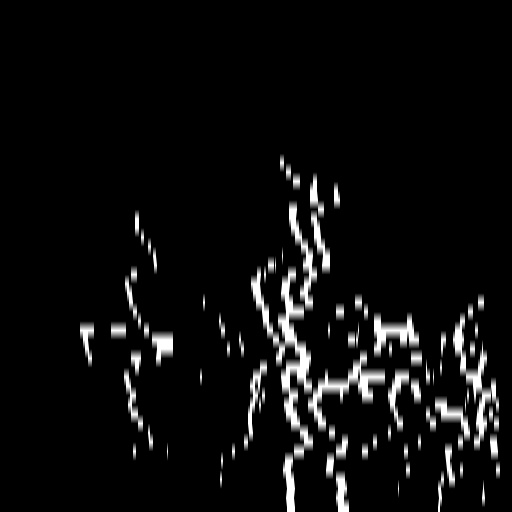} &
\includegraphics[width=0.125\textwidth]{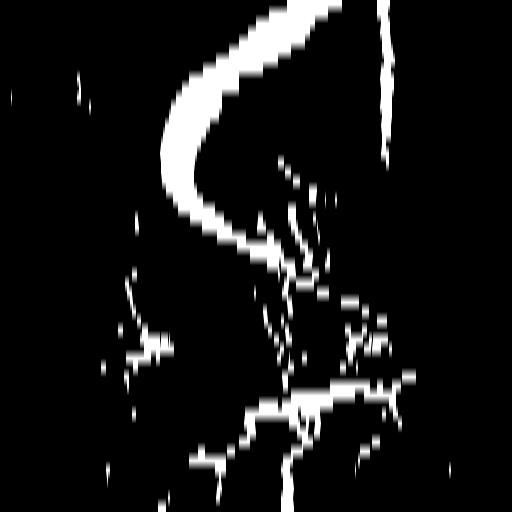} &
\includegraphics[width=0.125\textwidth]{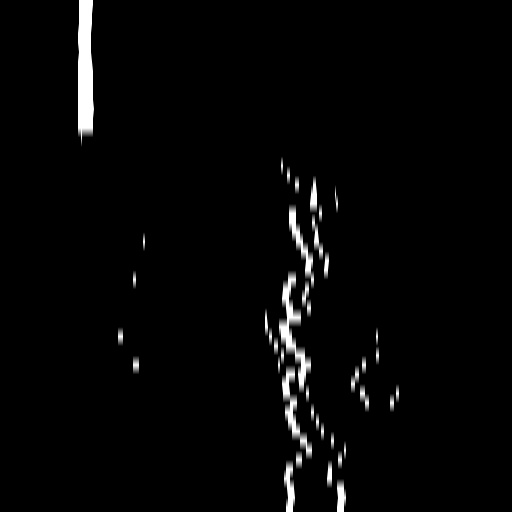} &
\includegraphics[width=0.125\textwidth]{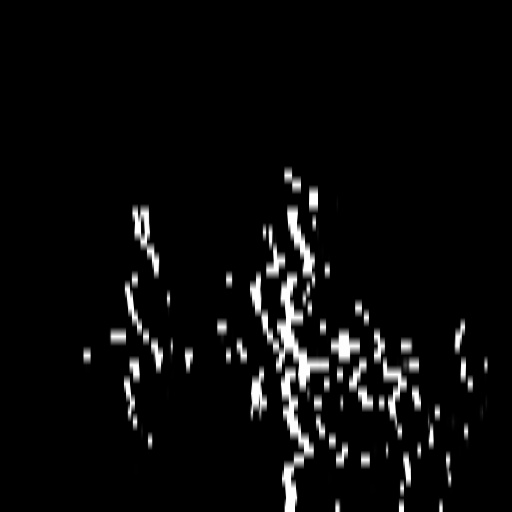} & 
\includegraphics[width=0.125\textwidth]{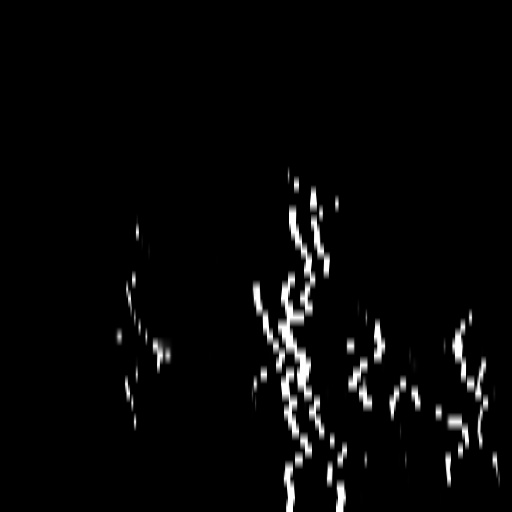}\\
\includegraphics[width=0.125\textwidth]{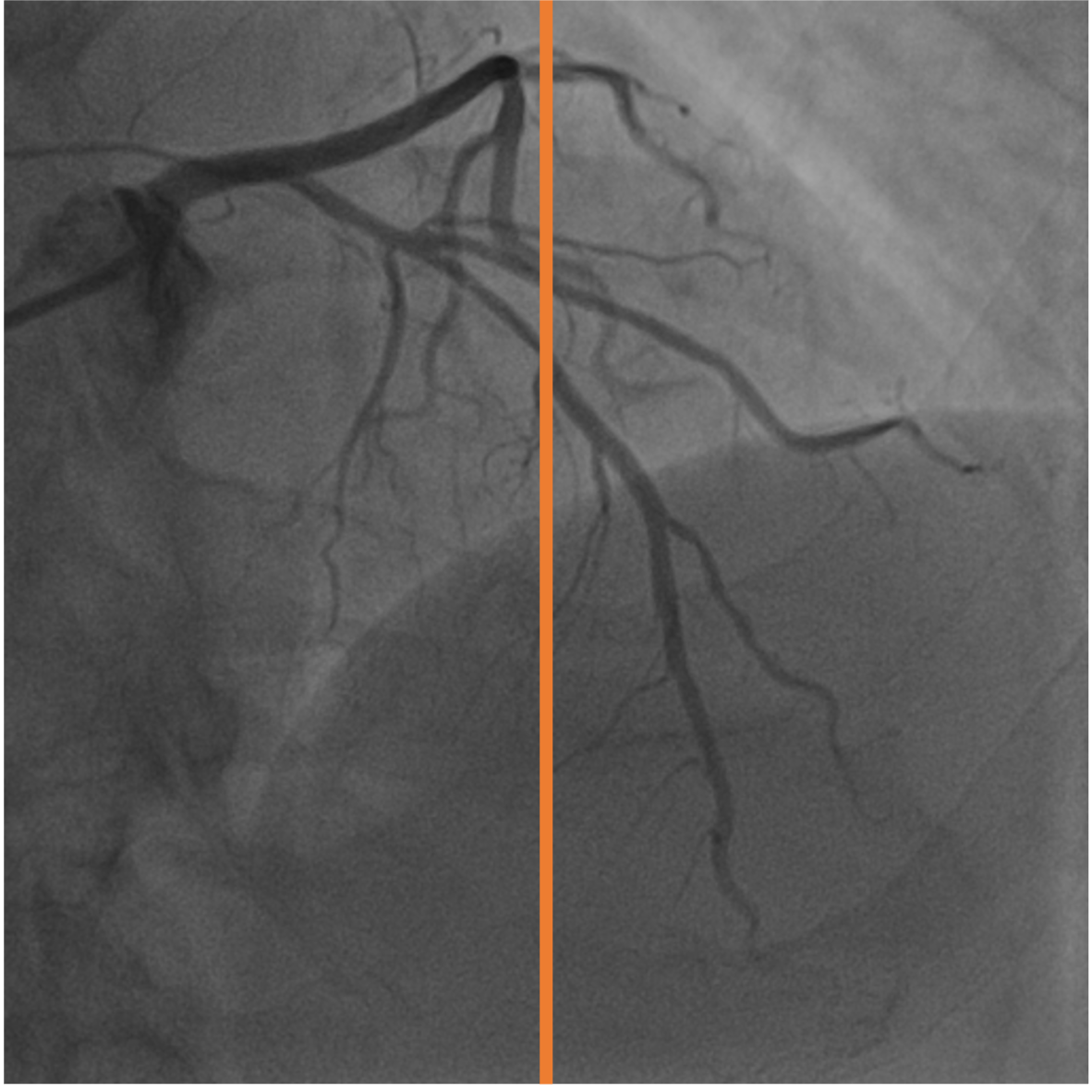} &
\includegraphics[width=0.125\textwidth]{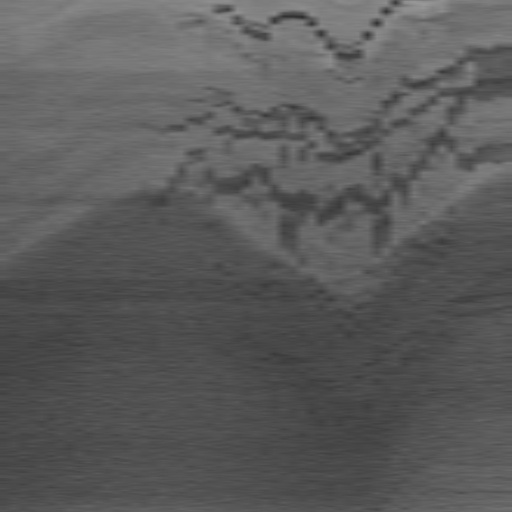} &
\includegraphics[width=0.125\textwidth]{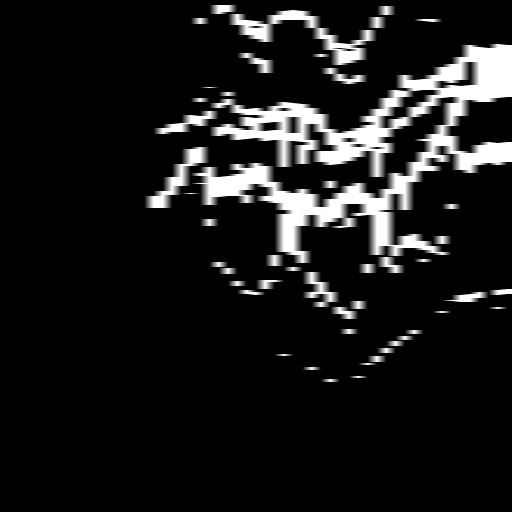} &
\includegraphics[width=0.125\textwidth]{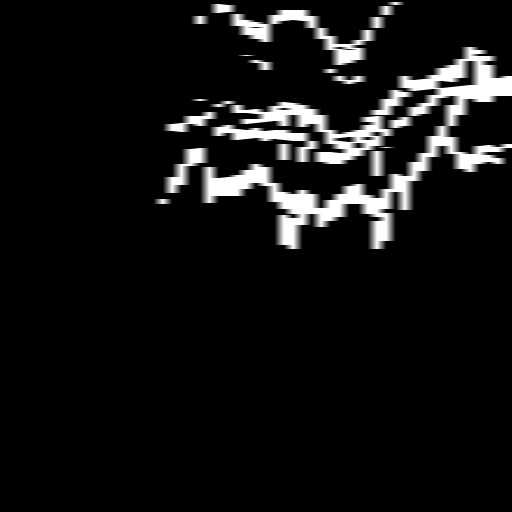} &
\includegraphics[width=0.125\textwidth]{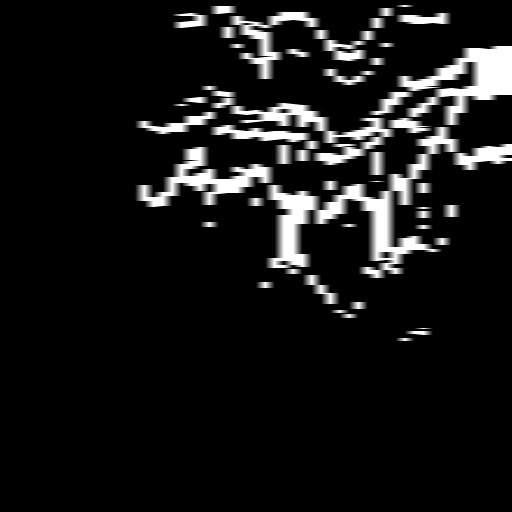} &
\includegraphics[width=0.125\textwidth]{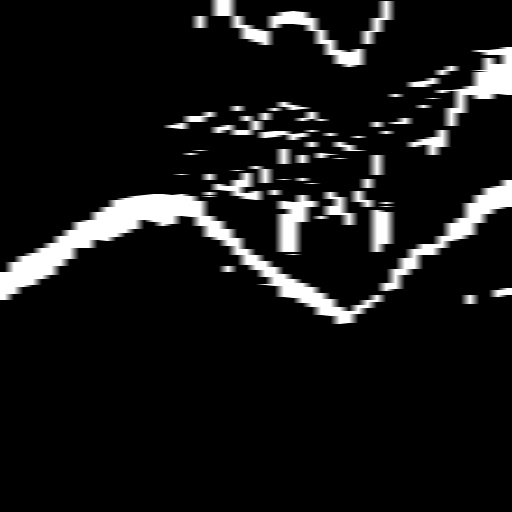} &
\includegraphics[width=0.125\textwidth]{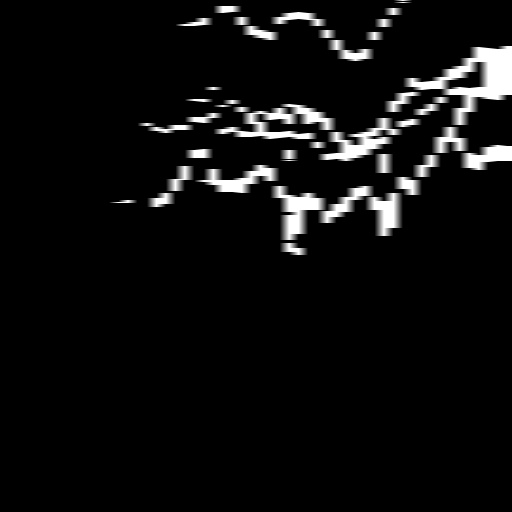} &
\includegraphics[width=0.125\textwidth]{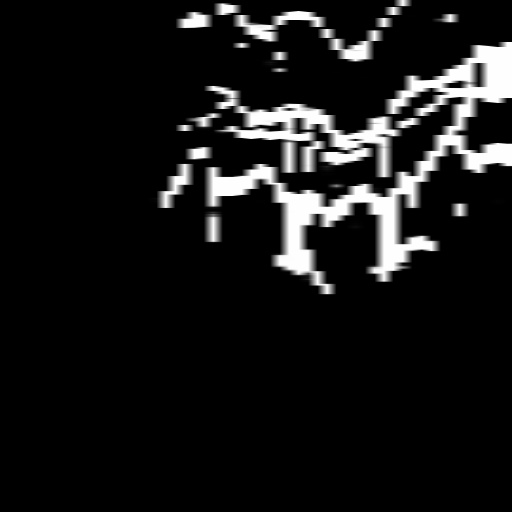} & 
\includegraphics[width=0.125\textwidth]{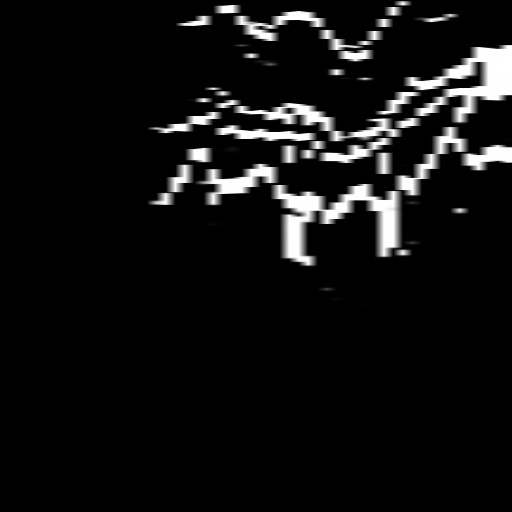}\\
\multicolumn{2}{c:}{Frame \& Temporal slice} & U-Net (image) & U-Net (video) & Hessian & SSVS & DARL & FreeCOS & \textbf{Ours} \\
& & \multicolumn{2}{c:}{\textbf{Supervised}} & \textbf{Traditional} & & \textbf{Self-supervised} & & \textbf{Unsupervised} \\
\end{tabular}%
}
\caption{\textbf{Temporal coherency.} }%~\yulunliu{Move this to the appendix?}}
% \vspace{-1mm}
\label{fig:temporal}
\end{figure*}

Our method takes an entire X-ray video as input, thus producing segmentation results with better temporal coherency.
Temporal coherency is essential for making medical diagnoses, especially when dealing with blood flow in vessels.
Therefore, we conduct visual comparisons between our method and other compared methods by slicing horizontally or vertically and stacking the segmentation results.
The results in \cref{fig:temporal} show our method strikes a better balance between segmentation accuracy and temporal coherency. 
While other baseline methods either produce false segmentation results or do not maintain consistent prediction along the temporal dimension.

\section{Impact of prior}
\begin{figure}[t]
\centering
\setlength{\tabcolsep}{1pt}
\renewcommand{\arraystretch}{1}
\resizebox{1.0\columnwidth}{!}{%
\begin{tabular}{cccc}
\includegraphics[width=0.25\columnwidth]{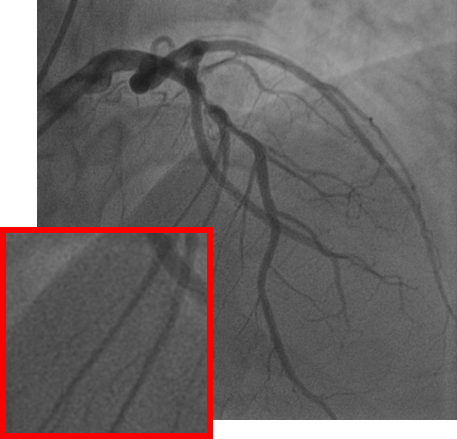}  &
\includegraphics[width=0.25\columnwidth]{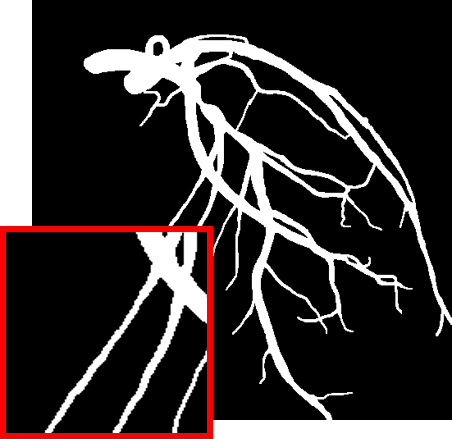}  &
\includegraphics[width=0.25\columnwidth]{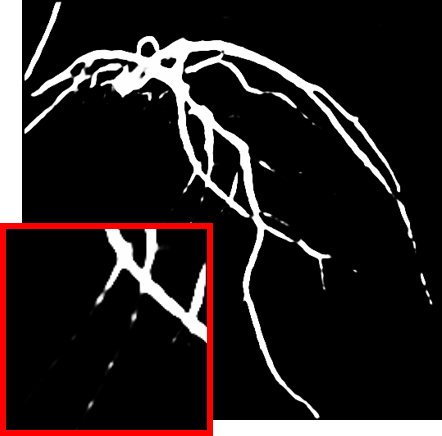}  &
\includegraphics[width=0.25\columnwidth]{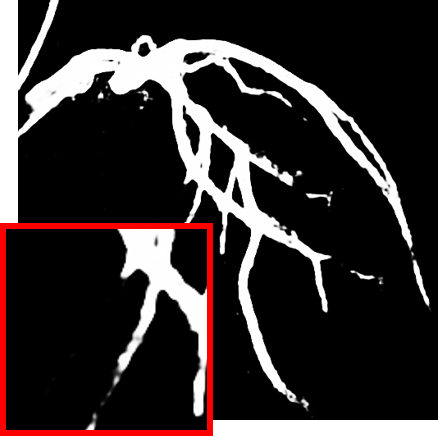} \\
Image & ground truth & Hessian prior mask & Freecos prior mask \\
\end{tabular}%
}
\vspace{-3mm}
\caption{\textbf{Impact of prior.}  We test the impact of the prior on our model. Replacing the original Hessian prior with the FreeCOS prediction results in a 2.5\% improvement in dice score. Red zoom-in patches show that the FreeCOS-based prior has better predictive capabilities.} 
\label{fig:prior}
\end{figure}

We add experiments demonstrating how the Hessian prior affects subsequent results, including ablation studies with different prior qualities. In our experiments, We replace the Hessian prior mask with a better mask (FreeCOS prediction) and observe a 2.5\% improvement in dice score.
We also provide visual results in \cref{fig:prior}.

\section{Model and training details}
We elaborate on the architectural details and training methodologies for all neural network components.

\subsection{Stage1: Layer Separation on bootstrapping}
This MLP (Multi-Layer Perceptron) model consists of these main components: 
\begin{itemize}
\item Input Layer: Input dimension is 3 (color channels).
\item Hidden Layer 1: Takes input of dimension 3 and outputs a dimension of 2. This layer has 256 neurons, with 4 hidden layers and an outermost linear layer.
\item Hidden Layer 2: Takes input of dimension 2 and outputs a dimension of 3. This layer also has 256 neurons, with 4 hidden layers and an outermost linear layer.
\item Output Layer: Takes input of dimension 3 and outputs a dimension of 4. This layer has 256 neurons, with 4 hidden layers and an outermost linear layer.
\end{itemize}

Important hyperparameters:
\begin{itemize}
\item $\lambda_\text{smooth}$: Controls the weight of the smoothness term in the bootstrapping loss. We set it to 0.001.
\item $\lambda_\text{limit}$: Regularizes the foreground MLP in the bootstrapping loss. We set it to 0.02.
\end{itemize}

\subsection{Stage 2: Vessel decomposition}
In stage 2, We employ different standard U-Nets with three down and three up layers to predict masks and foreground canonical images. Both models utilize CNNs with 3$\times$3 kernels, strides of 1, and padding of 1. We use batch norm and bilinear downsampling or upsampling after each layer in the U-Nets.

\paragraph{Training setting.}
We set the batch size to 16, including 512$\times$512 image resolution, and trained on a 4090 GPU. Training on 80-90 cardiac images takes approximately 20 minutes.

\end{document}